\documentclass[preprint,12pt,authoryear]{elsarticle}

\usepackage[utf8]{inputenc}
\usepackage{mathtools, amssymb}
\usepackage{enumitem}
\usepackage{xcolor}
\usepackage{kotex}
\usepackage{tabularray}

\usepackage{threeparttable}
\usepackage{booktabs}
\usepackage{bbm}
\usepackage{multirow}
\newtheorem{theorem*}{Theorem}

\newcommand{\BlackBox}{\rule{1.5ex}{1.5ex}}  
\newenvironment{proof}{\par\noindent{\bf Proof\ }}{\hfill\BlackBox\\[2mm]}
 
\newtheorem{theorem}{Theorem}
\newtheorem{lemma}[theorem]{Lemma} 
\newtheorem{proposition}[theorem]{Proposition} 
\newtheorem{remark}[theorem]{Remark}
\newtheorem{corollary}[theorem]{Corollary}

\newenvironment{skproof}{\par\noindent{\bf Sketch of Proof\ }}{\hfill\BlackBox\\[2mm]}
\newenvironment{nameproof}[1]{\par\noindent{\bf #1 \ }}{\hfill\BlackBox\\[2mm]}

\newtheorem{condition}{Condition}

\newcommand{\commaa}[1]{\expandafter\ifx\expandafter\relax\detokenize{#1}\relax
\else #1,\fi}

\newcommand{\CNN}[4]{{\Sigma}^{\commaa{#1}#2}_{\commaa{#3}#4}}
\newcommand{\oCNN}[4]{\overline{\CNN{#1}{#2}{#3}{#4}}}
\newcommand{\sCNN}[4]{#1 \left( \CNN{}{#2}{#3}{#4}\right) }
\newcommand{\ssCNN}[4]{#1 \left(\CNN{#1}{#2}{#3}{#4}\right)}
\newcommand{\osCNN}[4]{\overline{\sCNN{#1}{#2}{#3}{#4}}}
\newcommand{\ossCNN}[4]{\overline{\ssCNN{#1}{#2}{#3}{#4}}}

\journal{Neural Networks}

\begin{document}

\begin{frontmatter}



\title{Universal Approximation Property of Fully Convolutional Neural Networks with Zero Padding}


\author[label1]{Geonho Hwang}
\author[label2]{Myungjoo Kang\corref{cor1}}
\ead{mkang@snu.ac.kr}
\cortext[cor1]{Corresponding author}

\affiliation[label1]{organization={Center for AI and Natural Sciences, Korea Institute for Advanced Study},
            addressline={85, Hoegi-ro}, 
            city={Dongdaemun-gu},
            postcode={02455}, 
            state={Seoul},
            country={Republic of Korea}}

\affiliation[label2]{organization={Department of Mathematical Sciences, Seoul National University},
            addressline={1, Gwanak-ro}, 
            city={Gwanak-gu},
            postcode={08826}, 
            state={Seoul},
            country={Republic of Korea}}

\begin{abstract}
The Convolutional Neural Network (CNN) is one of the most prominent neural network architectures in deep learning. 
Despite its widespread adoption, our understanding of its universal approximation properties has been limited due to its intricate nature. 
CNNs inherently function as tensor-to-tensor mappings, preserving the spatial structure of input data.
However, limited research has explored the universal approximation properties of fully convolutional neural networks as arbitrary continuous tensor-to-tensor functions.
In this study, we demonstrate that CNNs, when utilizing zero padding, can approximate arbitrary continuous functions in cases where both the input and output values exhibit the same spatial shape.
Additionally, we determine the minimum depth of the neural network required for approximation and substantiate its optimality.
We also verify that deep, narrow CNNs possess the UAP as tensor-to-tensor functions.
The results encompass a wide range of activation functions, and our research covers CNNs of all dimensions.

\end{abstract}


\begin{highlights}
    \item We established that fully convolutional neural networks (FCNNs) with zero padding have the universal approximation property (UAP) in the continuous tensor-to-tensor function space.
    \item We demonstrated the relationship between the dimensions of the input data and the optimal depth of three-kernel FCNNs, which possess the UAP. 
    \item We proved that deep, narrow FCNNs with $c_x+c_y+2$ channels have the UAP, where $c_x$ and $c_y$ are the numbers of channels of the input and output data, respectively.
    This establishes a minimum upper bound that does not depend on the number of spatial components.
    \item We demonstrated that in the deep, narrow case, a polynomial activation function is as expressive as the specific non-polynomial continuous function. 
This result also applies to multilayer perceptrons (MLPs) and improves the existing result regarding the minimum width \citep{kidger2020universal}.
\end{highlights}


\end{frontmatter}


\section{Introduction}
\textit{Convolutional neural networks}(CNNs) \citep{cnnintro2015introduction, lenetlecun1998gradient} are one of the most widely used deep learning modules and have achieved tremendous success in various fields, including object detection \citep{zaidi2022surveyobjectdetection}, image classification \citep{elngar2021imageclassification}, and sound processing \citep{tan2021surveysound}. 
Starting with basic architectures such as LeNet5 \citep{lenetlecun1998gradient}, several well-known deep learning models, including VGGNet \citep{vggsimonyan2014very}, ResNet \citep{resnethe2016deep}, and ResNeXt \citep{resnextxie2017aggregated}, have been constructed based on CNNs. 

The \textit{Universal Approximation Property} (UAP) refers to the ability of a specific set of functions to approximate a sufficiently wide range of functions.
Understanding the UAP of CNNs is crucial because it enables us to comprehend their capabilities and limitations as function approximators.
However, despite the extensive range of applications of CNNs, their UAP remains not well understood, and much research focuses on scalar-valued functions.
For instance, \citet{cnn_zhou2020universality, he2022approximation_relu_cnn} investigated CNNs with a fully connected last layer and demonstrated the UAP of the network as a function from $\mathbb{R}^{d}$ to $\mathbb{R}$. 

However, CNNs often function as \textbf{tensor-to-tensor functions}, preserving the spatial structure of the input data. 
This is achieved through the use of \textit{fully convolutional neural networks} (FCNN), which consist solely of convolutional layers.
For instance, FCNNs can function as a mapping from $\mathbb{R}^{c\times d_1\times d_2}$ to $\mathbb{R}^{c\times d_1\times d_2}$.
Representative examples of such usage include object segmentation \citep{long2015fullyconvobjectsegment}, depth estimation \citep{bhoi2019monoculardepth}, or image processing such as deblurring \citep{zhang2022deepdeblurring}, inpainting \citep{suthar2014surveyinpainting}, and denoising \citep{fan2019briefdenoising}. 
Another typical role for CNNs is as feature extractors.
These extract information from the data and feed it to the latter part of the deep learning model.
Typically, the features extracted by the CNN have a spatial structure, such as $\mathbb{R}^{c\times W\times H}$. 
In cases where a single feature extractor is commonly utilized across multiple networks, as is typical in transfer learning using pre-trained models as feature extractors, CNNs must possess the UAP.

Some research \citep{invariantyarotsky2022universal, maron2019invariantuniversality} has explored the UAP of FCNNs with vector-valued output as translation invariant or equivariant functions.
These studies considered an FCNN as a function from $\mathbb{R}^d$ to $\mathbb{R}^d$.
However, the network's equivariance inevitably impedes the use of widely adopted padding methods, such as zero-padding. 
Additionally, the equivariance assumption restricts the UAP within the context of arbitrary continuous function spaces.

In this regard, we studied the UAP of CNNs with convolutional layers utilizing zero padding.
Unlike previous methods that only considered scalar output or translation invariant or equivariant functions, we addressed the UAP of the \textbf{fully convolutional neural networks} as \textbf{arbitrary continuous tensor-to-tensor functions}.
While zero-padding convolution has been widely adopted in CNNs, prior studies omitted its investigation due to its perceived impact on network equivariance. 
However, we demonstrated that zero padding plays a crucial role in achieving the UAP of CNNs as continuous functions. 
Specifically, the UAP arises because zero-padding disrupts the translation equivariance of CNNs.
Our examination of CNNs with zero padding involved exploring the minimum depth and width requirements for achieving the UAP. 
The contributions of this study are as follows:
\begin{itemize}
    \item We established that FCNNs have the UAP in the continuous tensor-to-tensor function space.
    \item We demonstrated the relation between the dimensions of the input data and the optimal depth of three-kernel CNNs, which possess the UAP. 
    \item We proved that deep, narrow FCNNs with $c_x+c_y+2$ channels have the UAP, where $c_x$ and $c_y$ are the numbers of channels of the input and output data, respectively. (Channels will be defined in Section \ref{sec:notations})
    This establishes a minimum upper bound that does not depend on the number of spatial components.
    \item We demonstrated that in the deep, narrow case, a polynomial activation function is as expressive as the specific non-polynomial continuous function. 
This result also applies to multilayer perceptrons (MLPs) and improves the existing result regarding the minimum width \citep{kidger2020universal}.
    
\end{itemize}
\subsection{Organization}
The remainder of this paper is organized as follows.
In Section 2, we provide a brief review of previous studies on the UAP of neural networks.
Section 3 defines the symbols and terms that will be used.
In Section 4, we address the UAP of CNNs in both the wide, shallow, and deep, narrow cases.
Section 5 extends our discussion to the UAP of multidimensional CNNs.
Section 6 concludes the study.

\section{Related Works}
In this section, we provide a brief overview of studies that have investigated the universal approximation theorem of neural networks.
Early studies examined whether two-layered MLPs have a UAP, which means whether they can approximate any continuous function.
\citet{Cybenko_1989} demonstrated that two-layered MLPs are universal when the activation function is sigmoidal.
Several other studies \citep{Hornik_1989, hornik1991approximation} attempted to clarify the conditions for the activation function.
Among them, \citet{leshno1993multilayer} proposed a simple and powerful result. They proved that the necessary and sufficient condition for two-layered MLPs to be universal is that the activation function is non-polynomial. 

The UAP of deep, narrow MLPs is another interesting area of research that has been widely investigated.
\citet{Lu_2017} addressed scenarios where the width of the MLP is limited.
\citet{hanin2017approximating} extended this result to cases where the output is a multidimensional vector. 
\citet{Johnson_2019, kidger2020universal} further extended the result to encompass more general activation functions, including non-affine polynomials, proposing the lower and upper bounds, respectively.
In contrast to previous research that focused on continuous function spaces, \citet{park2020minimum} investigated the universal approximation in $L_p$ spaces.

With the increasing variety of neural network types and the expanding range of function spaces of interest, there is a growing demand for adapting the universal approximation theorem to accommodate these diverse structures.
\citet{Schafer_2007} and \citet{Hanson_2020} studied the UAP of recurrent neural networks (RNNs) and demonstrated that RNNs could approximate arbitrary open dynamical systems.
\citet{Yun_2020} also demonstrated the UAP of the transformer network.

However, despite their widespread use, research on CNNs as continuous functions (non-equivariant) is relatively scarce.
\citet{cnn_zhou2020universality} investigated the UAP of CNNs when they function as mappings from vectors to scalar values.
This study introduced a fully connected layer in the last layer of the network to ensure that the network's output is a scalar.
Additionally, in order to leverage the homomorphism between the composition of convolutional layers and the multiplication of polynomials, the paper assumed that the spatial dimension of the data increases as it passes through the network.
Similarly, \citet{he2022approximation_relu_cnn} explored the UAP of residual networks with scalar-valued outputs, while \citet{zhou2020theory} delved into the UAP of CNNs with downsampling layers.
It is worth noting that these studies did not primarily focus on FCNNs; instead, they incorporated additional network structures, such as fully connected layers or downsampling layers. 
These structures collapse the spatial structure of the data into a single output spatial dimension.
In contrast, our research is centered on FCNNs that maintain the original shape of the input data throughout the output data.

Another branch of research focuses on the UAP of FCNNs in translation equivariant function spaces.
\citet{petersen2020equivalence} examined the UAP of periodic convolutional networks as translation equivariant functions. 
\citet{invariantyarotsky2022universal} considered more general group action equivariant or invariant functions.
However, it is important to note that translation equivariance fundamentally contradicts the UAP of continuous function space that maps $d$-dimensional input data to $d$-dimensional output data. 
Because translation equivariance is based on cyclic padding, it necessitates an analysis of padding methods that disrupt translation equivariance, such as zero padding.

\section{Notation and Definition} 
In this section, we introduce notations and define mathematical concepts that will be used in the remaining sections.

\subsection{Notations}\label{sec:notations}
$\mathbb{R}$ represents the set of real numbers, while $\mathbb{R}_{+}$ denotes the set of positive real numbers.
$\mathbb{Z}$ stands for the set of integers.
$\mathbb{N}$ represents the set of natural numbers, and $\mathbb{N}_0 = \mathbb{N}\cup \{0\}$. 
$[a,b]$ is the interval of integers $\{n\in\mathbb{Z}: a\leq n\leq b\}$.
$C(X,Y)$ denotes the set of continuous functions from $X$ to $Y$.
For any value $y\in Y$, we slightly abuse notation by identifying $y$ with the constant-valued function $\widetilde{y} \in C(X,Y)$, which is defined as $\widetilde{y}(x) = y$ for all $x\in X$.
$\boldsymbol{1}_n\in \mathbb{R}^n$ is the $n$-dimensional vector with all components equal to one.
$e^d_i\in \mathbb{R}^d$ represents the $i$-th standard basis vector of Euclidean space $\mathbb{R}^d$.
When the dimension $d$ is clear from the context, we may omit it and denote $e^d_i$ as $e_i$.
$(x_{i,j})_{1\leq i\leq d_1, 1\leq j\leq d_1}\in \mathbb{R}^{d_1\times d_2}$ or simply $(x_{i,j})_{i,j}$ denotes the $d_1\times d_2$ matrix with $x_{i,j}$ as an $i$-th row, $j$-th column component. 

For operations, we use the following notations:
The Hadamard product of two vectors, $x=(x_1, x_2, \dots, x_d), y=(y_1, y_2,\dots, y_d)\in \mathbb{R}^d$, is denoted as $ x\odot y = (x_1y_1, x_2y_2, \dots, x_d y_d)\in \mathbb{R}^d$. For the multiple Hadamard product of vectors $x^1,x^2, \dots, x^n\in \mathbb{R}^d$, we denote it as $\bigodot_{i=1}^n x^i:= x^1\odot x^2\odot \dots \odot x^n\in \mathbb{R}^d$.

We will typically work with values in the form of $\mathbb{R}^{c\times d}$, $\mathbb{R}^{c\times \mathbb{Z}}$, $\mathbb{R}^{c\times d_1\times d_2}$, or $\mathbb{R}^{c\times \mathbb{Z}\times \mathbb{Z}}$.
Here, the first axis, denoted as $c$ or its variant, will be referred to as the \textit{channel}.
When indexing the channel, we will use the superscripts.
For instance, for $x\in \mathbb{R}^{c\times d}$, we represent it as:
\begin{equation}\label{eq:channel_indexing}
    x = (x^1 ,x^2, \dots, x^c),
\end{equation}
where $x^i\in \mathbb{R}^d$ for $i \in [1,c]$.
The channel axis consistently takes precedence over the other axes.

Axes other than the channel axis will be referred to as \textit{spatial axes}, and the dimensions such as $d$ or $d_1\times d_2$ will be referred to as \textit{spatial dimensions}.
When indexing a spatial axis, we will use subscripts.
For example, we represent the components of $x^i\in \mathbb{R}^d$ in Eq (\ref{eq:channel_indexing}) as:
\begin{equation}
    x^i = (x^i_1, x^i_2,\dots, x^i_d),
\end{equation}
where $x^i_j\in\mathbb{R}$ for $i\in [1,c]$ and $j\in [1,d]$.

For the concatenation operation $\oplus$ along the channel, we define it as follows: 
For $x=(x^1, x^2, \dots, x^{c_1})\in \mathbb{R}^{c_1 \times d}$ and $y=(y^1, y^2, \dots, y^{c_2})\in \mathbb{R}^{c_2 \times d} $,
\begin{equation}
    x \oplus y := (x^1, x^2, \dots, x^{c_1},y^1, y^2, \dots, y^{c_2} )\in \mathbb{R}^{(c_1 + c_2)\times d}.
\end{equation}
The notations defined earlier are specifically for tensors; however, we will extend them to include operations between functions when the output of those functions is a tensor.
For instance, consider functions $A_i:\mathbb{R}^{c\times d}\rightarrow \mathbb{R}^{d}$. We will define $\left(\bigodot_i A_i\right):\mathbb{R}^{c\times d}\rightarrow \mathbb{R}^{d}$ as $\left(\bigodot_i A_i\right)(x) = \bigodot_i \left(A_i(x)\right)$.

\subsection{Definitions}
Here, we define the mathematical concepts that will be used in the following sections.
\begin{itemize}
    \item \textbf{Infinite-Length Convolution}: For $k\in \mathbb{N}$, let $w = (w_{-k}, w_{-k+1},\dots, w_k)\in \mathbb{R}^{2k+1}$.
    Then, an infinite-length convolution with the kernel $w$ is defined as a mapping $f_w:\mathbb{R}^{\mathbb{Z}} \rightarrow  \mathbb{R}^{\mathbb{Z}}$ defined as follows: For $x=(\dots, x_{-1},x_0, x_1, \dots)\in \mathbb{R}^{\mathbb{Z}}$,
    \begin{equation}
        \left(f_w (x)\right)_i := \sum_{j=-k}^{k} w_{j}x_{i-j},
    \end{equation}
    where $f_w(x)=(\dots, \left(f_w(x)\right)_{-1}, \left(f_w(x)\right)_{0}, \left(f_w(x)\right)_{1},\dots)\in \mathbb{R}^\mathbb{Z}$.
    A convolution with a kernel is said to have a kernel size $2k+1$.

    \item \textbf{Zero-Padding Convolution}: Let $\iota:\mathbb{R}^{d}\rightarrow \mathbb{R}^{\mathbb{Z}} $ be the natural inclusion map. Formally, for $x=(x_1, x_2, \dots, x_d)\in \mathbb{R}^d$,
    \begin{equation}
       \iota(x)_i =\iota_i(x) := \begin{cases}
            x_i \text{ \quad if } 1\leq i\leq d \\
            0  \text{ \quad\;otherwise } 
        \end{cases},
    \end{equation}
    where $\iota(x)=(\dots, \iota_{-1}(x), \iota_0(x), \iota_1(x),\dots)\in \mathbb{R}^\mathbb{Z}$.
    Let $p_d: \mathbb{R}^{\mathbb{Z}}\rightarrow \mathbb{R}^{d} $ be the projection map; that is, for $x = (\dots, x_{-1},x_0,x_1, \dots)\in \mathbb{R}^{\mathbb{Z}}$, $p_d(x)$ is defined as
    \begin{equation}
                p_d(x) := (x_1, x_2, \dots, x_d).
    \end{equation}
    Let $w\in \mathbb{R}^{2k+1}$ be a kernel.
    Then, zero-padding convolution with kernel $w$ is a function $g:\mathbb{R}^{d}\rightarrow \mathbb{R}^{d}$ defined as
    \begin{equation}
       g := p_d \circ f_w\circ \iota  ,
    \end{equation}
    where $f_w$ is the infinite-length convolution with kernel $w$. 
    We also define it as the operation $\circledast :\mathbb{R}^{2k+1}\times \mathbb{R}^{d}\rightarrow \mathbb{R}^{d}$:
    \begin{equation}
        w \circledast  x:=g(x).
    \end{equation}
    We can interpret the composition as constructing a temporary infinite-length sequence by filling zeros in the remaining components, conducting the infinite-length convolution with the kernel, and cutting off the unnecessary components. 

Zero-padding convolution with kernel $w=(w_{-k},\dots, w_k)$ is a linear transformation and can be expressed as matrix multiplication; $w \circledast x = Tx$ is satisfied for the following matrix $T\in \mathbb{R}^{d\times d} $ if we consider $x$ as a column vector:
\begin{equation}
T:= 
  \begin{bmatrix}
  w_0             &w_{-1}     & \dots    & w_{-k}    &      &        &       &         &              &      \\
  w_{1}          &w_0     & \dots    & w_{-k+1}&w_{-k}   &        &       &         &              &      \\
     \vdots       &        &\ddots     &\ddots  &\ddots&        &       &         &              &      \\
    w_{k}        &w_{k-1}& \dots    & w_0    & w_{-1}&\dots  &w_{-k+1}& w_{-k}     &              &      \\ 
                  & w_{k} & \dots    & w_{1} & w_{2}&\dots  &w_{-k+2}& w_{-k+1} & w_{-k}          &      \\
                  &        &           &\ddots  &\ddots&\ddots  &\ddots &\ddots   &\ddots        &      \\   
                  &        &           &        &      &        &w_{k} & w_{k-1}&      \dots  &w_0    
     
\end{bmatrix}.
\end{equation}
We define the set of Toeplitz matrices as 
\begin{align}
    \mathcal{T}^d_s:= \left\{ (t_{i,j})_{i,j}\in \mathbb{R}^{d\times d}\left| k\in [-s,s], w_k\in\mathbb{R}, t_{i,j} =\begin{cases}
        w_{i-j}  &\text{  if } |i-j|\leq s
        \\ 0 &\text{  otherwise}
    \end{cases} \right.\right\}.  
\end{align}
Additionally, we define $U^d_s$, or simply $U_s$ if the dimension is clear from the context, as 
\begin{equation}
    U^d_s := (u_{i,j})_{_{1\leq i,j\leq d}},
\end{equation}
where
\begin{equation} \label{eq:u_matrix}
u_{i,j} :=
    \begin{cases}
        1 \text{ \quad if } i-j = s\\
        0 \text{ \quad otherwise } 
    \end{cases}.
\end{equation}
By definition, $U_0$ is the identity matrix, and  $U_{s}$ and $U_{-s}$ have a transpose relationship with each other; $U_{s}^{T} = U_{-s}$. The set $\{ U_{-s}, U_{-s+1}, \dots U_s \}$ is the basis of the set of Toeplitz matrices $\mathcal{T}^d_s$.
Evidently, $(U_{1})^s = U_s$, and $(U_{-1})^s = U_{-s}$ for $t\geq 0 $. 

Furthermore, zero-padding convolution with kernel $w=(w_{-s}, w_{-s+1},\dots, w_s)$ can be represented as
\begin{equation}\label{eq:conv_to_matrix}
    w\circledast x = \sum_{i=-s}^s w_{i}U_{i}x.
\end{equation}
Here, we interpret $x$ as a column vector. 
Henceforth, if we multiply the translation matrix $U_s$ by a spatial vector $x\in\mathbb{R}^d$, we will treat $x$ as a column vector.

We also define $E^d_{n,m}\in \mathbb{R}^{d\times d}$, or simply $E_{n,m}$, as $E^d_{n,m} := (e_{i,j})_{1\leq i,j\leq d}$, where
\begin{equation}
    e_{i,j} = \begin{cases}
    1 &\text{ \quad if } i=n \text{ and } j=m\\
    0 &\text{ \quad otherwise }
\end{cases}.
\end{equation}
For both $U^d_s$ and $E^d_{n,m}$, we can interpret matrices as linear transformations from $\mathbb{R}^d$ to $\mathbb{R}^d$.

To deal with the composition of convolutions, we define $S^d_{N}$ as follows:
\begin{equation}
    S^d_{N} := \left\{\sum_{i=1}^{n}\prod_{j=1}^N T_{i,j}\middle| T_{i,j}\in  \mathcal{T}^d_1, n\in \mathbb{N} \right\}.
\end{equation}
$S^d_{N}$ is a vector space of matrix representations that linear three-kernel, $N$-layered CNNs can express.
$S^d_{N}$ will be used in Section \ref{sec:cnn_width}.

    \item \textbf{Zero-Padding Convolutional Layer:} A convolutional layer with $c_1$ input channels and $c_2$ output channels is a function $f:\mathbb{R}^{c_1\times d}\rightarrow \mathbb{R}^{c_2\times d} $.
    For each $1 \leq i\leq c_2$ and $1\leq j\leq c_1$, there exist zero-padding convolutions with kernel $w_{i,j}\in \mathbb{R}^{2k+1}$ and bias $\delta_i\in\mathbb{R}$, such that for $x = (x^1, x^2, \dots, x^{c_1})\in \mathbb{R}^{c_1 \times d}$,
    \begin{equation}
        \left(f(x)\right)^i=f^{i}(x) := \sum_{j=1}^{c_1} w_{i,j} \circledast x^j + \delta_i \boldsymbol{1}_d,
    \end{equation}
    where $f(x) = (f^1(x),f^2(x),\dots, f^{c_2}(x))$.
    We define the set of convolutional layers with $c_1$ input channels, $c_2$ output channels, and a kernel size $2k+1$ as $\mathcal{L}^{d, 2k+1}_{c_1, c_2} \subset C( \mathbb{R}^{c_1\times d}
        , \mathbb{R}^{c_2\times d} )
    $, or $\mathcal{L}^{2k+1}_{c_1, c_2}$ if dimension $d$ is evident from the context:
    \begin{equation}\label{eq:definition_convolutional_layer}
        \mathcal{L}^{d, 2k+1}_{c_1, c_2}:=
        \left\{ 
        f\left|
        {w_{i,j}\in \mathbb{R}^{2k+1},
        \delta_i\in\mathbb{R},}
        f^{i}(x)=\sum_{j=1}^{c_1} w_{i,j} \circledast x^j + \delta_i \boldsymbol{1}_d \right.\right\}
    \end{equation}
    
    We extend the operation $\circledast$ to the multiplication between vector-valued matrices and vectors.
    Let $M_{n,m}\left(\mathbb{R}^d\right)$ be the $n\times m$ matrix whose components are $d$-dimensional vectors in $\mathbb{R}^d$.
    Then, for $A=(a_{i,j})_{1\leq i\leq n,1\leq j\leq m}\in M_{n,m}(\mathbb{R}^{2k+1})$ and $B=(b_{j,k})_{1\leq j\leq m, 1\leq k\leq l}\in M_{m,l}(\mathbb{R}^{d})$, we denote matrix multiplication $\circledast$ between $A$ and $B$ as
    \begin{equation}
        C := A\circledast B,
    \end{equation}
    where $C=(c_{i,k})_{1\leq i\leq n,1\leq k\leq l}\in M_{n,l}(\mathbb{R}^{d})$, and $c_{i,k}$ is calculated as
    \begin{equation}
        c_{i,k} := \sum_{j=1}^m a_{i,j}\circledast b_{j,k}.
    \end{equation}
    The zero-padding convolutional layer can be interpreted as a matrix multiplication between the weight matrix $W=(w_{i,j})_{1\leq i\leq c_2, 1\leq j\leq c_1}\in M_{c_2, c_1}(\mathbb{R}^d)$ and the input vector $X = (x^j)_{1\leq j\leq c_1}\in M_{c_1, 1}(\mathbb{R}^{d})$ with bias summation added.
    \begin{equation}\label{eq:matrix_multiplication}
        \begin{bmatrix}
        f^1\\f^2\\ \vdots\\f^{c_2}
        \end{bmatrix}
        =  
        \begin{bmatrix}
        w_{1,1} & w_{1,2}&\dots & w_{1,c_1}
        \\w_{2,1} & w_{2,2}&\dots & w_{2,c_1}
        \\\vdots& \vdots&\ddots&\vdots
        \\w_{c_2,1} & w_{c_2,2}&\dots & w_{c_2,c_1}
        \end{bmatrix}
        \circledast
        \begin{bmatrix}
        x^1\\x^2\\ \vdots\\x^{c_1}
        \end{bmatrix}
         + \begin{bmatrix}
        \delta_1 \boldsymbol{1}_d\\ \delta_2\boldsymbol{1}_d \\\vdots\\\delta_{c_2}\boldsymbol{1}_d
        \end{bmatrix}
.    \end{equation}
  \item \textbf{Activation Function:} An activation function $\sigma$ is a scalar function $\sigma:\mathbb{R}\rightarrow \mathbb{R}$. 
  We component-wisely extend the function to multivariate versions such as $\sigma_{d}: \mathbb{R}^{d}\rightarrow \mathbb{R}^{d}$ and $\sigma_{c,d}: \mathbb{R}^{c\times d}\rightarrow \mathbb{R}^{c\times d}$.
  We will slightly abuse notation to ensure that $\sigma$ means $\sigma$, $\sigma_{d}$, or $ \sigma_{c, d}$, depending on the context.
  
We also define a modified version of the activation function that selectively applies an activation function to each channel.
  For $I\subset [1,c]$, we define $\widetilde{\sigma}_{I}:\mathbb{R}^{c\times d}\rightarrow \mathbb{R}^{c\times d}$ as follows:
If $x = (x^1,x^2,\dots, x^c)$ and $x^i\in \mathbb{R}^d$,
\begin{equation}
    \widetilde{\sigma}^i_{I}(x) = \begin{cases}
        \sigma(x^i) \text{ \quad if } i\in I
        \\ x^i \text{ \;\quad \quad  otherwise } 
    \end{cases}
    ,
\end{equation}
where $  \widetilde{\sigma}_{I} = (\widetilde{\sigma}^1_{I}, \widetilde{\sigma}^2_{I},\dots, \widetilde{\sigma}^c_{I})$.

  \item \textbf{CNN:} 
  An $N$-layered CNN with channel sizes $\mathbf{c} = (c_0, c_1, \dots, c_N)$ and a kernel size $2k+1$ is a mapping $f:\mathbb{R}^{c_0\times d}\rightarrow\mathbb{R}^{c_N\times d} $ that is constructed with the following $N$ convolutional layers and an activation function. 
For convolutional layers $C_{i}\in \mathcal{L}^{d,2k+1}_{c_{i-1},c_i}$ and activation function $\sigma$, $f$ is defined as follows:
 \begin{equation}
     f := C_{N} \circ \sigma \circ C_{N-1}\circ \dots \circ \sigma \circ C_1.
 \end{equation}
We define the set of CNNs with channel sizes $\mathbf{c} = (c_0, c_1, \dots, c_N)$ and a kernel size $2k+1$ as 
\begin{equation}
     \CNN{\sigma}{d,2k+1}{N}{\mathbf{c}} :=\left.\left\{C_{N} \circ \sigma \circ C_{N-1}\circ \dots \circ \sigma \circ C_1:\mathbb{R}^{c_0\times d}\rightarrow\mathbb{R}^{c_N\times d}\right| C_{i}\in \mathcal{L}^{2k+1}_{c_{i-1},c_i} \right\}.
\end{equation}
We also define $\CNN{\sigma}{d, 2k+1}{N}{c,c'}$ as the set of all $N$-layered CNNs with $c$ input channels, $c'$ output channels, and a kernel size $2k+1$:
\begin{equation}
    \CNN{\sigma}{d, 2k+1}{N}{c,c'} := \bigcup_{\substack{c_1, c_2, \dots c_{N-1}\in \mathbb{N}\\ 
    \mathbf{c} = (c,c_1,c_2,\dots,c_{N-1}, c')}}  
    \CNN{\sigma}{d, 2k+1}{N}{\mathbf{c}}.
\end{equation}
If the activation function is clear from the context, we omit it and denote the set as $\CNN{}{d, 2k+1}{N}{c,c'}$.
The same omit method applies to the size of the input and kernel: $\CNN{}{}{N}{c,c'}$, $\CNN{}{d,2k+1}{N}{c,c'}$, and $\CNN{}{\sigma}{N}{c,c'}$ means $\CNN{\sigma}{d,2k+1}{N}{c,c'}$.

Additionally, we define $\ssCNN{\sigma}{d,2k+1}{N}{c,c'}\subset C(\mathbb{R}^{c\times d},\mathbb{R}^{c'\times d})$ as
 \begin{equation}
     \ssCNN{\sigma}{d,2k+1}{N}{c,c'} := \left\{\left. \sum^{n}_{i=1}a_i (\sigma \circ f_i) \; \right|\; f_i\in \CNN{\sigma}{d,2k+1}{N}{c,c'}, a_i\in \mathbb{R},n\in \mathbb{N}_0\right\}
 \end{equation}
 Similarly, $\sCNN{\sigma}{}{N}{c,c'}$ means $\ssCNN{\sigma}{d,2k+1}{N}{c,c'}$.

 Furthermore, we define the set of deep, narrow CNNs $\Delta^{\sigma, d, 2k+1}_{c,c',l}$, or simply $\Delta_{c,c',l}$, as
\begin{equation}
    \Delta^{\sigma, d, 2k+1}_{c,c',l}
    := \bigcup_{\substack{N\in\mathbb{N} \\
    c_1, c_2, \dots c_{N-1}\in [1, l]\\ 
    \mathbf{c} = (c,c_1,c_2,\dots,c_{N-1}, c')}}  
    \CNN{\sigma}{d, 2k+1}{N}{\mathbf{c}}.
\end{equation}
The sets $\CNN{\sigma}{d, 2k+1}{N}{c,c'}$, $\ssCNN{\sigma}{d,2k+1}{N}{c,c'}$, and $\Delta^{\sigma, d, 2k+1}_{c,c',l}$, are subsets of $C(\mathbb{R}^{c\times d},\mathbb{R}^{c'\times d})$, representing the set of continuous functions defined over the entire domain $\mathbb{R}^{c\times d}$.
However, we will slightly abuse the notation to allow these sets to denote functions restricted to a smaller domain, provided the domain is clear from the context.
For instance, $\CNN{\sigma}{d, 2k+1}{N}{c,c'} $ can be a subset of $C(K, \mathbb{R}^{c'\times d})$ for $K\subset \mathbb{R}^{c\times d}$.

\end{itemize}

\section{Main Theorem}
\subsection{Problem Formulation}
The discussion in this paper focuses on the UAP of CNNs concerning whether a continuous function from $\mathbb{R}^{c\times d}$ to $\mathbb{R}^{c'\times d}$ can be uniformly approximated by CNNs.
We consider the uniform norm $\|\cdot\|_{\infty,K}$ in $C(K, \mathbb{R}^{c'\times d})$ for each non-empty compact subset $K$ of $\mathbb{R}^{c\times d}$ in the following manner:
\begin{equation}
    ||f-g||_{\infty, K} = \sup_{x\in K}||f(x)-g(x)||_{\infty}.
\end{equation}
We will denote the closure of a subset $A$ of $C(K, Y)$ with respect to the uniform norm $\|\cdot\|_{\infty,K}$ as $\overline{A}$.

In Section \ref{sec:cnn_depth}, we present a theorem demonstrating the optimal depth required for shallow, wide CNNs to approximate entire continuous functions:
\begin{theorem}[The UAP of Shallow, Wide CNNs]\label{thm:main_theorem}
 For a spatial dimension $d\in \mathbb{N}$, channel sizes $c, c'\in \mathbb{N}$, and a non-polynomial continuous activation function $\sigma$, the minimum depth $N_d$ required for three-kernel CNNs to achieve the UAP in the continuous function space from any arbitrary compact set $K\subset \mathbb{R}^{c\times d}$ to $\mathbb{R}^{c'\times d}$ is given by:
 \begin{equation}
N_d=     \begin{cases}
    2 & \text{      if\quad} d=1,2
    \\ 3 &\text{   if\quad }  d=3
    \\ d-1 &\text{  if\quad }  d\geq 4
\end{cases}.
 \end{equation}
In other words, the relation
 \begin{equation}
      \oCNN{\sigma}{d,3}{N_d}{c,c'} = C(K,\mathbb{R}^{c'\times d}),
 \end{equation}
holds for an arbitrary compact set $K\subset \mathbb{R}^{c\times d}$, and there exists a compact set $K\subset \mathbb{R}^{c\times d}$ such that the following relation holds:
\begin{equation}
    \oCNN{\sigma}{d,3}{N_d-1}{c,c'} \nsupseteq C(K,\mathbb{R}^{c'\times d}).
\end{equation}
\end{theorem}

In Section \ref{sec:cnn_width}, we will demonstrate that CNNs with bounded width have the UAP in $C(K, \mathbb{R}^{c'\times d})$ concerning the uniform norm:
\begin{theorem}[The UAP of Deep, Narrow CNNs]\label{theorem:narrow}
For a spatial dimension $d$, channel sizes $c_1, c_2\in \mathbb{N}$, and a compact set $K\subset \mathbb{R}^{c_1\times d}$, a kernel size $3\leq 2k+1\in \mathbb{N}$, and a non-affine continuous activation function $\sigma$ satisfying Condition \ref{condition:activation}, the following relation holds:
\begin{equation}
    \overline{\Delta^{\sigma, d, 2k+1}_{c_1,c_2,c_1 + c_2+ 2}} = C(K, \mathbb{R}^{c_2\times d}).
\end{equation}
\end{theorem}
 Condition \ref{condition:activation} will be defined in Subsection \ref{sec:cnn_width}.

\subsection{Lemma}\label{sec:lemma}
Before delving into the main theorems, we will initially establish the lemma to be used for the subsequent proofs.
\begin{lemma}\label{spacelemma}
The following statements hold:
\begin{enumerate}
    \item $\CNN{}{}{N}{c,c'}$ is closed under concatenation: for $f_1\in \CNN{}{}{N}{c,c'}$ and $ f_2\in \CNN{}{}{N}{c,c''}$, $f_1\oplus f_2\in \CNN{}{}{N}{c,c'+c''}$.
    \item $\CNN{}{}{N}{c,c'}$ and $\sCNN{\sigma}{}{N}{c,c'}$ are vector spaces. 
    \item For a $C^{\infty}$ activation function $\sigma$, $\oCNN{\sigma}{d,2k+1}{N}{c,c'} $ is closed under partial differentiation with respect to a parameter; for a $C^{\infty}$ function $(x,\theta)\mapsto f(x, \theta)$ and $f_{\theta}(x):= f(x, \theta)$, if $f_{\theta}(x) \in {\oCNN{}{}{N}{c,c'}}$ for any $\theta \in \mathbb{R}$, then $\frac{\partial f_{\theta}(x)}{\partial \theta}\in \oCNN{}{}{N}{c,c'}$.
    Furthermore, $\osCNN{\sigma}{}{N}{c,c'}$ is closed under partial differentiation in the same sense.
    \item For $f\in \osCNN{\sigma}{}{N}{c,c'}$ and a convolutional layer $C\in \mathcal{L}^3_{c', c''}$, $C \circ f\in  \oCNN{}{}{N+1}{c,c''}$.
\end{enumerate}
\end{lemma}
    The proof of Lemma \ref{spacelemma} can be found in \ref{appendix:proof_spacelemma}. 
This lemma delineates the operations applicable for constructing the space of CNNs, encompassing concatenation, linear combination, partial differentiation, and function composition.

\subsection{Minimum Depth for the Universal Approximation Property of CNNs}\label{sec:cnn_depth}
In this subsection, we illustrate the minimum depth required for three-kernel CNNs to have the UAP. 
Unlike MLPs, which achieve the UAP with only two-layered networks \citep{leshno1993multilayer}, CNNs require a significantly deeper minimum depth.
This deeper depth requirement stems from the receptive field limitation imposed by convolution using a finite kernel.
With a convolutional layer having a kernel size of three, each output is influenced by neighboring spatial components on the left and right.
Hence, in a CNN constructed by stacking these $N$ layers of convolutional layers, the input can involve values from the left and right $N$ components.
Consequently, for a function with $d$-dimensional input and output, a minimum of $d-1$ layers is necessary for the first output component to consider the last input component. 
Therefore, to achieve the UAP using CNNs with a kernel size of three, a minimum of $d-1$ layers is necessary.
This naturally raises the question of whether a depth of $d-1$ suffices for the UAP.
The following proposition demonstrates that a depth of $d-1$ is insufficient for $d=3$.

\begin{proposition}\label{prop:d3_not_universal}
For channel sizes $c,c'\in \mathbb{N}$, an open set $U\subset \mathbb{R}^{c\times 3}$ containing the origin, and a compact domain $K\subset \mathbb{R}^{c\times 3}$ containing $U$, three-kernel, two-layered CNNs do not have the UAP as a function from $K$ to $\mathbb{R}^{c'\times 3}$.

In other words, the following relation holds:
\begin{equation}
    \oCNN{\sigma}{3,3}{2}{c,c'}\nsupseteq C(K, \mathbb{R}^{c'\times 3}).
\end{equation}
\end{proposition}
\begin{skproof}
The detailed proof of Proposition \ref{prop:d3_not_universal} can be found in \ref{appendix:proof_d3_not_universal}. 
The key idea lies in utilizing the translation equivariance of the CNN.
A CNN constructed using infinite-length convolution satisfies translation equivariance.
In CNNs using zero padding, however, translation equivariance is not entirely preserved and the small number of layers in CNNs imposes a similar restriction on the output.
Considering a CNN $f:\mathbb{R}^{c\times 3}\rightarrow \mathbb{R}^{1\times 3}$ represented as $f=(f_1, f_2,f_3)$, we compare $f_1(z)$ and $f_2(w)$, where $w$ is the translation of $z$.
    The function $f_1(z)-f_2(w)$ imposes constraints preventing $f$ from having the UAP.
\end{skproof}

As demonstrated in the proof, the persistence of translation equivariance in a shallow CNN hampers its ability to achieve the UAP.
Nevertheless, this constraint can be overcome by employing a deeper CNN.
Here, we present the primary proposition of this section, indicating that a depth of $d-1$ is sufficient for achieving the UAP when $d\geq 4$.

\begin{proposition}\label{lemma:main_theorem}
  For a spatial dimension $d \geq 4$, channel sizes $c,c'\in \mathbb{N}$, a non-polynomial continuous activation function $\sigma$, and a compact domain $K\subset \mathbb{R}^{c\times d}$, $(d-1)$-layered three-kernel CNNs have the UAP in the continuous function space from $K$ to $\mathbb{R}^{c'\times d}$. 
  
  In other words, the following relation holds:
  \begin{equation}
      \oCNN{\sigma}{d, 3}{d-1}{c,c'} = C(K, \mathbb{R}^{c'\times d}).
  \end{equation}
\end{proposition}
Before we proceed to prove Proposition \ref{lemma:main_theorem}, we will establish several lemmas.
\begin{lemma}\label{lemma:product}
 For $i\in [1, n]$, $l\in \mathbb{N}$, a non-polynomial $C^{\infty}$ activation function $\sigma$, and $A_i\in  \oCNN{}{}{l}{c,1}$, the following relation holds:
    \begin{equation}
         \bigodot_{i=1}^n A_i\in \osCNN{\sigma}{}{l}{c,1}.
    \end{equation}
\end{lemma}
\begin{proof}
   Given that $\oCNN{}{}{l}{c,1}$ forms a vector space according to Lemma \ref{spacelemma}, a linear combination of its elements remains within $\oCNN{}{}{l}{c,1}$.
    Thus, for $i\in [1,n]$, $a_i\in \mathbb{R}$, and $\delta \boldsymbol{1}_d\in \oCNN{}{}{l}{c,1}$,
    \begin{equation}
        f:=\sum_{i=1}^n a_iA_i + \delta \boldsymbol{1}_d \in \oCNN{}{}{l}{c,1}.
    \end{equation}
    By the definition of $\osCNN{\sigma}{}{l}{c,1}$,
    \begin{equation}
        \sigma(f)=\sigma\left(\sum_{i=1}^n a_iA_i + \delta \boldsymbol{1}_d \right)\in  \osCNN{\sigma}{}{l}{c,1}.
    \end{equation}
    According to Lemma \ref{spacelemma}, $\osCNN{\sigma}{}{l}{c,1}$ is closed under partial differentiation with respect to parameters. 
    Therefore,
    \begin{equation}
        \left(\prod_{i=1}^n\frac{\partial}{\partial a_i}\right)\left[\sigma\left(\sum_{i=1}^n a_iA_i +\delta \boldsymbol{1}_d \right)\right]\in  \osCNN{\sigma}{}{l}{c,1}.
    \end{equation}
The partial differentiation results in the Hadamard product:
\begin{equation}
     \left(\prod_{i=1}^n\frac{\partial}{\partial a_i}\right)\left[\sigma\left(\sum_{i=1}^n a_iA_i +\delta \boldsymbol{1}_d \right)\right] = \bigodot_{i=1}^n A_i \odot \sigma^{(n)}(f) \in  \osCNN{\sigma}{}{l}{c,1}.
\end{equation}
As $\sigma$ is a non-polynomial function, there exists $\delta_0$ such that $\sigma^{(n)}(\delta_0)\neq 0$.
When substituting all coefficients $a_i$ with zero and $\delta$ with $\delta_0$, we obtain:
\begin{equation}
    \left.\bigodot_{i=1}^n A_i \odot \sigma^{(n)}(f)\right|_{ \substack{a_1=\cdots=a_n=0,\\ \delta = \delta_0}} = \bigodot_{i=1}^n A_i \odot\sigma^{(n)} (\delta_0 \boldsymbol{1}_d)= \sigma^{(n)}(\delta_0)\bigodot_{i=1}^n A_i \in  \osCNN{\sigma}{}{l}{c,1}.
\end{equation}
As $\osCNN{\sigma}{}{l}{c,1}$ constitutes a vector space, $\bigodot_{i=1}^n A_i  \in \osCNN{\sigma}{}{l}{c,1}$.
This concludes the proof.
\end{proof}
This lemma will be used multiple times in the proof.
Furthermore, the lemma leads to the following corollary.

\begin{corollary}\label{Corollary:subset}
 For channel sizes $c,c'\in \mathbb{N}$, $l,l_0\in \mathbb{N}$ and a non-polynomial $C^{\infty}$ activation function $\sigma$, the following relations hold.
\begin{itemize}
    \item $\oCNN{}{}{l}{c,1}\subset \osCNN{\sigma}{}{l}{c,1}\subset \oCNN{}{}{l+1}{c,1}$.
    \item For $g\in \oCNN{}{}{l_0}{c, c'}$, $i\in [1,c']$, and $j\in [-l,l]$, define 
$f$ as $f(x):=U_j g^i(x)$. 
 Here, $g^i(x)$ is interpreted as a column vector.
Then, $ f\in \oCNN{}{}{l + l_0}{c,1}$. 
   
\end{itemize}
\end{corollary}
\begin{proof}
  The relation $\oCNN{}{}{l}{c,1}\subset \osCNN{\sigma}{}{l}{c,1}$ is a specific case of Lemma \ref{lemma:product} with $n=1$. 
  The relation $\osCNN{\sigma}{}{l}{c,1}\subset \oCNN{}{}{l+1}{c,1}$ holds because for an arbitrary $f\in\osCNN{\sigma}{}{l}{c,1}$ and the identity function $Id\in \mathcal{L}^3_{1,1}$, $f = Id\circ f \in \oCNN{}{}{l+1}{c,1}$ as shown by Lemma \ref{spacelemma}.
  
  For the second argument, we use mathematical induction with respect to $l$.
  For $l=1$, considering the definition of the convolutional layer (Eq (\ref{eq:definition_convolutional_layer})) and the matrix representation of zero-padding convolution (Eq (\ref{eq:conv_to_matrix})), we obtain $x\mapsto U_1x^i, U_0x^i, U_{-1}x^i\in \mathcal{L}^3_{c',1}$.
  Therefore, by Lemma \ref{spacelemma}, $U_j g^i\in \oCNN{}{}{l_0+1}{c,1}$ for $g\in \oCNN{}{}{l_0}{c,1}$ and $j=-1,0,1$, validating the induction hypothesis for $l=1$.
  
  Now, assuming the induction hypothesis holds for $l=l'$, i.e., $U_j g\in \oCNN{}{}{l_0+l'}{c,1}$ for $j\in [-l',l']$, we only need to prove that $U_{(l'+1)}g, U_{(-l'-1)}g \in \oCNN{}{}{l_0+l'+1}{c,1}$ because $\oCNN{}{}{l_0+l'}{c,1}\subset \oCNN{}{}{l_0+l'+1}{c,1}$.
  Considering that $\oCNN{}{}{l_0+l'}{c,1}\subset \osCNN{\sigma}{}{l_0+l'}{c,1}$, $U_{l'}g$ and $U_{-l'}g$ are elements of $\osCNN{\sigma}{}{l_0+l'}{c,1}$.
  Furthermore, the mappings $y \mapsto U_1y$ and $y \mapsto U_{-1}y$ belongs to $\mathcal{L}^{3}_{1,1}$, where $y\in \mathbb{R}^{d}$.
  According to Lemma \ref{spacelemma}, $x\mapsto U_1U_{l'}g^i(x) = U_{(l'+1)}g^i(x)$ and $x\mapsto U_{-1}U_{-l'}g^i(x) = U_{(-l'-1)}g^i(x)$ are elements of $\oCNN{}{}{l_0+l'+1}{c,1}$.
  Thus, the induction hypothesis is satisfied for $l=l'+1$, thereby concluding the proof.
\end{proof}

\begin{corollary}\label{lemma:onehot}
 For a spatial dimension $d\in \mathbb{N}$, $n\in \mathbb{N}$, $ i\in [1, n-1]\cup [d-n+2, d]$, and a non-polynomial $C^{\infty}$ activation function $\sigma$, $e^d_i\in \oCNN{\sigma}{d,3}{n}{c,1}$.
\end{corollary}

\begin{proof}
The constant function $\boldsymbol{1}_d \in \osCNN{\sigma}{}{1}{c,1}$. 
For the convolutional layer $B\in\mathcal{L}^3_{1,1}$ with kernel $b = (b_{-1},b_0,b_1)$ and bias $\delta$, it follows that
    \begin{equation}
     B\circ \boldsymbol{1}_d    \in \oCNN{}{}{2}{c,1}.
    \end{equation}
    In more detail:
    \begin{equation}
         B\circ \boldsymbol{1}_d = \delta \boldsymbol{1}_d +(b_{-1} + b_{0}, b_{-1}+ b_{0}+b_1, \dots, b_{-1}+ b_{0}+b_1,b_{0} +  b_{1}).
    \end{equation}
     Substituting $\delta$ with $-(b_{-1} + b_0 +b_1)$ leads to:
     \begin{equation}
         B\circ \boldsymbol{1}_d = (- b_{1},0,\dots,0,- b_{-1})\in\oCNN{}{}{2}{c,1},
     \end{equation}
for arbitrary $b_{-1}, b_1\in \mathbb{R}$.
Thus, $e_1, e_d \in \oCNN{}{}{2}{c,1}$.
By Corollary \ref{Corollary:subset}, $U_{j}e_1, U_{j}e_d \in \oCNN{}{}{n}{c,1}$ for $ j\in [-n+2, n-2]$.
Because $U_{j}e_1 = e_{j+1}$ for $j\geq 0$, and $U_{j}e_d = e_{d+j}$ for $j\leq 0$, $e_i\in  \oCNN{}{}{n}{c,1} $ for $i\in [1, n-1]\cup [d-n+2, d]$. This completes the proof.
\end{proof}

Using the above lemmas, we briefly sketch the proof of Proposition \ref{lemma:main_theorem} 
\begin{nameproof}{Proof Sketch of Proposition \ref{lemma:main_theorem}}
    The detailed proof of Proposition \ref{lemma:main_theorem} can be found in \ref{appendix:proof_main_theorem}. 
    Our objective is to construct arbitrary monomials of input components in arbitrary output spatial positions.
    Because $\oCNN{}{}{N}{c,c'}$ is a vector space, it implies that all polynomials are in $\oCNN{}{}{N}{c,c'}$, enabling us to approximate arbitrary continuous functions via the Stone--Weierstrass theorem \citep{de1959stone}.
    To construct the monomial, we leverage Lemma \ref{lemma:product}. 
    After an input traverses $N$ layers, it can be translated across up to $n$ spatial components (Corollary \ref{Corollary:subset}).
    Lemma \ref{lemma:product} allows us to combine these translated inputs, forming any product.
    For instance, given an input $x = (x_1, x_2, \dots ,x_d)\in \mathbb{R}^d$, the translated vectors $(0, x_1, x_2, \dots ,x_{d-1})$ and $(x_2, x_3,\dots ,x_d, 0)$ belong to $\oCNN{\sigma}{d,3}{3}{1,1}$.
    Therefore, applying Lemma \ref{lemma:product}, $(0, x_1x_2x_3, x_2x_3x_4, \dots, x_{d-2}x_{d-1}x_d, 0)\in \ossCNN{\sigma}{d,3}{3}{1,1}$.
    This method constructs arbitrary monomials in arbitrary spatial positions. 
    However, as demonstrated in the example, these monomials exhibit high correlation: $x_1x_2x_3$ in the second component and $x_2x_3x_4$ in the third component.
    To eliminate this correlation, we introduce the standard basis vector $e_i$ into the product.
    When $e_2\in \CNN{\sigma}{d,3}{3}{1,1}$, $e_2\odot (0, x_1x_2x_3, x_2x_3x_4, \dots, x_{d-2}x_{d-1}x_d, 0) = (0, x_1x_2x_3, 0,\dots, 0)\in \ossCNN{\sigma}{d,3}{3}{1,1}$.
    This approach narrows down the challenge to determining the values of $i$ for which the standard basis $e_i$ and the translated vectors are in $\oCNN{\sigma}{d,3}{l}{c,1}$, which is illustrated in Corollary \ref{lemma:onehot}.
    Through straightforward calculations, we confirm that functions from a sufficiently broad range belong to the set.

     However, it's important to note that Lemma \ref{lemma:product} only applies to $C^{\infty}$ activation functions.
    To extend the proof to encompass arbitrary continuous activation functions, we rely on Lemma \ref{lemma:activation_convolution}.
    This lemma suggests that CNNs with activation function $\sigma$ are universal if those with activation function $\varphi * \sigma$ are universal for a smooth function $\varphi$. 
    Given that the convolution of a function with a smooth function is smooth, CNNs with $\varphi * \sigma$ are universal, thus confirming the universality of CNNs for arbitrary non-polynomial continuous functions.
\end{nameproof}

\begin{lemma}\label{lemma:activation_convolution}
For a continuous activation function $\sigma$ and a compactly supported $C^{\infty}$ function $\varphi$, the convolution operation $*$ is defined as follows:
  \begin{equation}
   (\sigma * \varphi)(x) := \int^\infty_{-\infty} \sigma(x-y)\varphi(y)dy.
  \end{equation}
Then, the following relation holds for $d,N,c,c'\in \mathbb{N}$, and $k\in \mathbb{N}_0$:
\begin{equation}
\oCNN{\sigma*\varphi}{d,2k+1}{N}{c,c'}
    \subset \oCNN{\sigma}{d,2k+1}{N}{c,c'},.
\end{equation}
\end{lemma}

\begin{skproof}
    The detailed proof of Proposition \ref{lemma:activation_convolution} can be found in \ref{appendix:proof_activation_convolution}. 
    Because the convolution operator $\sigma *\varphi$ is a linear integral and $\sigma$ is continuous, we can approximate it using a linear combination of $\sigma$ with a sufficiently small error.
    Consequently, we can iteratively approximate each layer using a linear combination involving the composition of $\sigma$ and the previous layer.
\end{skproof}

\begin{remark}
Translation equivariance is often cited as the foundation of the advantages of FCNN models.
Infinite-length convolution is translation equivariant.
However, this property conflicts with the UAP for general tensor-to-tensor functions due to the relation between output and input vectors.
As demonstrated in the proof process, padding plays a crucial role.
The asymmetry originating from the boundary gradually propagates toward the center, ultimately facilitating the achievement of the UAP in the tensor-to-tensor case.
\end{remark}

To ensure completeness across all cases, we present the following lemma for $d=2,3$.
\begin{lemma}\label{lemma: d23}
  For a spatial dimension $d \in \{2,3\}$, channel sizes $c,c'\in \mathbb{N}$, a non-polynomial continuous activation function $\sigma$, and a compact domain $K\subset \mathbb{R}^{c\times d}$, $d$-layered three-kernel CNNs have the UAP in the continuous function space from $K$ to $\mathbb{R}^{c'\times d}$.
  
  In other words, the following relation holds:
  \begin{equation}
      \oCNN{\sigma}{d,3}{d}{c,c'} = C(K, \mathbb{R}^{c'\times d}).
  \end{equation}
\end{lemma}
The detailed proof of Lemma \ref{lemma: d23} can be found in \ref{appendix:proof_d23}.

Combining Proposition \ref{prop:d3_not_universal}, Proposition \ref{lemma:main_theorem}, and Lemma \ref{lemma: d23} yields Theorem \ref{thm:main_theorem}.
\begin{remark}
   Although we specifically focused on three-kernel CNNs, the core concept of the proof can be readily extended to kernels of any size.
       As long as the constant function represented by networks propagates sufficiently fast from the edge to the center, we can prove the UAP of CNNs using the same techniques.
\end{remark}

\subsection{Minimum Width for the Universal Approximation Property of CNNs}\label{sec:cnn_width}
In this section, we aim to prove the UAP of deep, narrow CNNs, as presented in Theorem \ref{theorem:narrow}.
The proof unfolds in the following steps:
Firstly, construct an arbitrary linear transformation using only convolutional layers as outlined in Lemma \ref{lemma:linear}.
Secondly, employ Lemma \ref{lemma:one_activation} to approximate arbitrary continuous functions by composing the linear summation with an activation layer.
Finally, construct the deep, narrow CNN capable of approximating the aforementioned network.


In this section, we will consider activation functions that satisfy the following condition:
\begin{condition}\label{condition:activation}
There exists an $\alpha\in \mathbb{R}$ and $\epsilon\in \mathbb{R}_+$ such that the activation function $\sigma$ is $C^1$ in the interval $(\alpha - \epsilon, \alpha + \epsilon)$, and $\sigma'(\alpha)\neq 0$.
\end{condition}
The following lemma asserts that CNNs with partial activation functions can be approximated by those sharing the same structure but with full activation functions.
\begin{lemma}\label{lemma:activation_remove}
For an activation function $\sigma$ satisfying Condition \ref{condition:activation}, convolutional layers $C_1\in \mathcal{L}^{2k+1}_{c_1,c_2}$, $C_2\in \mathcal{L}^{2k+1}_{c_2,c_3}$, an index set $I\subset [1, c_2]$, a compact set $K\subset \mathbb{R}^{c_1\times d}$, and a positive number $ \epsilon\in \mathbb{R}_+$, there exist convolutional layers $C'_1\in \mathcal{L}^{2k+1}_{c_1,c_2}$ and $C'_2\in \mathcal{L}^{2k+1}_{c_2,c_3}$ that satisfy the following equation: 
    \begin{equation}
        ||C_2 \circ \widetilde{\sigma}_I \circ C_1 - C'_2\circ\sigma  \circ C'_1||_{\infty, K} < \epsilon.
    \end{equation}
\end{lemma}
\begin{skproof}
The detailed proof of Lemma \ref{lemma:activation_remove} can be found in \ref{appendix:proof_activation_remove}. 
The key idea involves linearly approximating the function using Taylor expansion near $\alpha$.
This approximation yields $\sigma(x)\approx \sigma(\alpha) + (x-\alpha)\sigma'(\alpha)$, allowing for the proper selection of weights to recover a linear activation.
\end{skproof}

\begin{lemma}\label{lem:remove_activation_multi}
 For an activation function $\sigma$ satisfying Condition \ref{condition:activation}, a natural number $N\geq 2$, channel sizes $(c_0,c_1,\dots,c_N)$, $i\in [1,N]$, indices $I_i\subset [1, c_i]$, a compact set $K\subset \mathbb{R}^{c_0\times d}$, and convolutional layers $C_i\in \mathcal{L}^{2k+1}_{c_{i-1}, c_i}$, the CNN $f$ is defined as
\begin{equation}
    f:=C_N\circ \widetilde{\sigma}_{I_{N-1}} \circ  C_{N-1}\circ  \dots \circ \widetilde{\sigma}_{I_1} \circ C_1 .
\end{equation}
Then, for a positive number $ \epsilon\in \mathbb{R}_+$, there exists $g\in \CNN{\sigma}{d,2k+1}{N}{(c_0,c_1,\dots,c_N)}$ such that 
\begin{equation}
    \|f-g\|_{\infty, K}<\epsilon.
\end{equation}

%
\end{lemma}
\begin{skproof}
The detailed proof of Lemma \ref{lem:remove_activation_multi} can be found in \ref{appendix:proof_remove_activation_multi}. 
We employ mathematical induction and repeatedly apply Lemma \ref{lemma:activation_remove}.
\end{skproof}
Lemma \ref{lem:remove_activation_multi} implies that we can freely replace a part of the activation function with the identity.
Consequently, multiple convolutions can be composed without being affected by the activation function.
We will demonstrate that an arbitrary linear combination of the input can be generated through a composition of convolutions.

\begin{lemma}\label{lemma:linear}
For $d\in \mathbb{N}$ and an arbitrary matrix $L\in \mathbb{R}^{d\times d}$, it holds that $L\in S^d_{d}$. 
\end{lemma}
\begin{proof}
In this proof, it is convenient to interpret the matrix multiplication with $U_s$ in the following way: Suppose  $A$ is a matrix or a column vector. Then, $U_{s}A$ and $U_{-s}A$ move $A$ downward by $s$ rows and upward by $s$ rows, respectively. Similarly, $AU_{s}$ and $AU_{-s}$ move $A$ to the left by $s$ columns and right by $s$ columns, respectively.

We aim to prove that for arbitrary $1\leq n,m \leq d$, $E_{n,m}\in S^d_{d}$.
    Firstly, we know that $ U_0 - U_1 U_{-1} = E_{1,1} $. 
    Consequently, $ E_{n,m}=U_{1}^{n-1}E_{1,1}U_{-1}^{m-1} =U_{1}^{n-1} (U_0 - U_1 U_{-1})U_{-1}^{m-1} =U_{1}^{n-1}U_{-1}^{m-1} - U_{1}^{n}U_{-1}^{m}$. 
Hence, if $n+m\leq d$, then $E_{n,m}\in S^d_{d}$.
 Similarly, $U_0-U_{-1}U_1 = E(d,d)$. 
 Additionally, $E_{n,m} = U_{-1}^{d-n}E_{d,d}U_{1}^{d-m} = U_{-1}^{d-n}(U_0-U_{-1}U_1 )U_{1}^{d-m} = U_{-1}^{d-n}U_{1}^{d-m} - U_{-1}^{d-n+1}U_{1}^{d-m+1}$.
 Thus, if $(d-n+1) + (d-m+1)  \leq d$, then $E_{n,m}\in S^d_{d}$.
Alternatively, if $n+m\geq d+2$, then $ E_{n,m} \in S^d_{d} $.

The remaining task is to prove the case when $n+m = d+1$.
 Divide the case into two subcases. Firstly, consider the case where $n \geq m$. 
We can observe that $(U_1)^{n-m} = \sum_{i= - m+1}^{d-n} E_{n+i,m+i} $.
It is known that $ E_{n+i,m+i}\in S^d_{d}$ for all $ i<0$ (Because $(n+i)+(m+i) = d+1 + 2i\leq d $) and $i>0$ (Because $(n+i)+(m+i) = d+1 + 2i\geq d+2 $).
Because $(U_1)^{n-m} \in S^d_{d}$, $E_{n,m} = (U_1)^{n-m}- \sum_{i\neq  0}E_{n+i,m+i} \in S^d_{d}$. 
Similarly, if $ n<m$, then $(U_{-1})^{m-n} = \sum_{i= -n+1}^{d-m} E_{n+i,m+i} $, and thus $E_{n,m} = (U_{-1})^{m-n}- \sum_{i\neq  0}E_{n+i,m+i} \in S^d_{d}$, completing the proof.


\end{proof}

Lemma \ref{lemma:linear} implies that by compositing and adding convolutions, we can achieve an arbitrary linear transformation of input.

The following lemma demonstrates that using linear transformations and just one activation function layer, we can approximate an arbitrary continuous tensor-to-tensor function.
\begin{lemma}\label{lemma:one_activation}
For a spatial dimension $d$ and a channel size $c$, let the set of functions $T\subset C({\mathbb{R}^{c\times d},\mathbb{R}^{d}})$ be defined as follows:
For $x=(x^1, x^2, \dots, x^c)\in \mathbb{R}^{c\times d}$, where $x^i\in \mathbb{R}^d$ for $i\in [1,c]$, a compact domain $K\in \mathbb{R}^{c\times d}$, and a non-polynomial continuous activation function $\sigma$ satisfying Condition \ref{condition:activation},
\begin{equation}
    T^{\sigma}_d:=\left\{\left. \sum_{j=1}^n a_j\sigma\left(\sum_{i=1}^{c} L_{j,i} x^i + \boldsymbol{\delta}_{j} \right) \right| n\in \mathbb{N}_0, L_{j,i}\in \mathbb{R}^{d\times d}, \boldsymbol{\delta}_{j}\in \mathbb{R}^d, a_j\in \mathbb{R} \right\}.
\end{equation}
Then, $\overline{T^{\sigma}_d} = C(K, \mathbb{R}^{1\times d})$.
\end{lemma}
\begin{proof}
Firstly, it sufficies to prove the case where the activation function $\sigma$ is $C^{\infty}$. 
As demonstrated in the proof of Proposition \ref{lemma:main_theorem}, we can select a compact-supported $C^{\infty}$ function denoted as $\varphi$ such that $\varphi * \sigma$ is $C^{\infty}$. 
Therefore, the lemma holds for $\varphi * \sigma$ if the lemma is true for $C^{\infty}$ activation function.
 Then, following the proof methodology of Lemma \ref{lemma:activation_convolution}, $\varphi * \sigma$ can be approximated by a linear combination of $\sigma$, and $\overline{T^{\varphi * \sigma}_d} \subset \overline{T^{\sigma}_d}$.
Hence, our focus narrows down to proving the case where the activation function $\sigma$ is $C^{\infty}$. 
 
Now, let $x^i = (x^i_1, x^i_2,\dots, x^i_d)\in \mathbb{R}^d$.
We define an arbitrary monomial of variables $x^i_j$ as:
\begin{equation}
    M = \prod_{i=1}^c \prod_{j=1}^d (x^i_j)^{\alpha_{i,j}},
\end{equation}
where degrees $\alpha_{i,j}\in \mathbb{N}_0$.
Our aim is to demonstrate that for $k\in [1,d]$,
 \begin{equation}
     Me_k = (0,0,\dots,0,M,0,\dots,0)\in \overline{T^{\sigma}_d}.
 \end{equation}
Then, the lemma is proven by the Stone--Weierstrass theorem \citep{de1959stone}.
Just as in the proof techniques for Lemma \ref{spacelemma}, it can be verified that $\overline{T^{\sigma}_d}$ is a vector space and is closed under partial differentiation with respect to the parameters. 
For $\boldsymbol{\delta} = (\delta_1, \delta_2, \dots, \delta_d)$ and $b_{i,j}\in \mathbb{R}$, define $f:\mathbb{R}^{c\times d}\rightarrow\mathbb{R}^{d}$ as follows:
\begin{equation}
    f(x):=\sum_{i=1}^{c}\sum_{j=1}^d b_{i,j} E_{k,j} x^i +\boldsymbol{\delta }.
\end{equation}
Then, according to the definition of $T^{\sigma}_d$, we have:
\begin{equation}
    \sigma\circ f(x)=\sigma\left(\sum_{i=1}^{c}\sum_{j=1}^d b_{i,j} E_{k,j} x^i +\boldsymbol{\delta } \right) \in \overline{T^{\sigma}_d}.
\end{equation}
The partial differentiation of $f$ with respect to $b_{i,j}$ results in:
\begin{equation}
    \frac{\partial}{\partial b_{i,j}}f = E_{k,j}x^i = e_k x^i_j.
\end{equation}
Subsequently, the partial differentiation with respect to all parameters $b_{i,j}$ gives the following equation:
 \begin{equation}
      \left( 
       \prod_{i=1}^c \prod_{j=1}^d \left(\frac{\partial}{\partial b_{i,j}} \right)^{\alpha_{i,j}} \right)\sigma(f) 
      =  \left(\prod_{i=1}^c \prod_{j=1}^d (x^i_j)^{\alpha_{i,j}} \right) e_k \odot\sigma^{(n)}(f),
\end{equation}
where $n = \sum_{i=1}^c\sum_{j=1}^d \alpha_{i,j} $.
Then, selecting $\delta_k$ such that $\sigma^{(n)}(\delta_k)\neq 0$ and $b_{i,j}=0$, we obtain
\begin{equation}
    Me_k\in \overline{T^{\sigma}_d}.
\end{equation}
Therefore, all polynomials belong to $\overline{T^{\sigma}_d}$, and by the Stone--Weierstrass theorem, $\overline{T^{\sigma}_d} = C(K, \mathbb{R}^{d})$.
\end{proof}

Now, we proceed to demonstrate the UAP of the deep, narrow CNN through the following proof.
\begin{nameproof}{Proof of Theorem \ref{theorem:narrow}}
Given that $ \overline{\Delta^{\sigma, d, 3}_{c_x,c_y,c_x + c_y+ 2}}\subset  \overline{\Delta^{\sigma, d, 2k+1}_{c_x,c_y,c_x + c_y+ 2}}$ for $2k+1\geq 3$, we only need to consider $2k+1=3$.
Assume $\sigma$ is non-polynomial.
To start, consider a function $f$ with $c$ input channels and one output channel:
\begin{equation}
    f:\mathbb{R}^{c \times d} \rightarrow \mathbb{R}^{1\times d}.
\end{equation}
We denote the input as $x$ and each channel of input as $x=(x^1, x^2, \dots, x^c)$. 
By Lemma \ref{lemma:one_activation}, there exists a function $g:\mathbb{R}^{c\times d}\rightarrow\mathbb{R}^{1\times d}$ defined as follows:
\begin{equation}
    g(x):=\sum_{j=1}^n a_j\sigma\left(\sum_{i=1}^{c} L_{j,i} x^i + \boldsymbol{\delta}_{j}\right),
\end{equation}
which can approximate $f$ with an error smaller than $\epsilon$.
Now, construct a deep, narrow CNN with a channel size of $c+3$, which approximates $g$.
By Lemma \ref{lemma:linear}, for an arbitrary $L_{j,i}\in \mathbb{R}^{d\times d}$, there exists $C^{k,l}_{i,j} \in  \mathcal{T}^d_1$ such that
\begin{equation}
    L_{j,i} = \sum_{l=1}^{m_{i,j}}\prod_{k=1}^d C^{k,l}_{i,j}.
\end{equation}
Moreover, there exists $\widetilde{C}^{k,l}_{j} \in  \mathcal{T}^d_1$ such that
\begin{equation}
     \boldsymbol{\delta}_j = \sum_{l=1}^{\widetilde{m}_{j}}\prod_{k=1}^d \widetilde{C}^{k,l}_{j} \boldsymbol{1}_d.
\end{equation}
Then, $g$ becomes
\begin{equation}
g(x)=    \sum_{j=1}^n a_j\sigma\left(\sum_{i=1}^{c} L_{j,i}x^i +\boldsymbol{\delta}_j \right) = \sum_{j=1}^n a_j\sigma\left(\sum_{i=1}^{c} \sum_{l=1}^{m_{i,j}}\prod_{k=1}^d C^{k,l}_{i,j} x^i + \sum_{l=1}^{\widetilde{m}_j}\prod_{k=1}^d \widetilde{C}^{k,l}_{j} \boldsymbol{1}_d\right).
\end{equation}
Define a deep, narrow CNN with a channel size of $c+3$ to calculate the above equation.
Using Lemma \ref{lemma:activation_remove}, if we can approximate the function with a CNN using a partial activation function, we can approximate it with the original CNN of the same size. 
Therefore, we can preserve $c$ channels from the input and process only the $(c+1)$-th, $(c+2)$-th, and $(c+3)$-th channels. 
We obtain the desired output based on the following process of function compositions.
\begin{enumerate}[label*=\arabic*.]
\item Repeat 2 and 3 for $j = 1,2,\dots, n$:
    \item Calculate $\sigma\left(\sum_{i=1}^{c} \sum_{l=1}^{m_{i,j}}\prod_{k=1}^d C^{k,l}_{i,j} x^i + \boldsymbol{\delta}_j \right) $ in the $(c+2)$-th channel, excluding the use of the $(c+3)$-th channel.
    \begin{enumerate}[label*=\arabic*.]
    \item Repeat the following for $i = 1,2,\dots, c$ and $l=1,2,\dots, m_{i,j}$ to get $\sum_{i=1}^{c} \sum_{l=1}^{m_{i,j}}\prod_{k=1}^d C^{k,l}_{i,j} x^i $ in the $(c+2)$-th channel.
    \begin{enumerate}[label*=\arabic*.]
        \item Calculate $\prod_{k=1}^d C^{k,l}_{i,j}x^i$ in the $(c+1)$-th channel, not using the $(c+2)$-th and $(c+3)$-th channels.
        \begin{enumerate}[label*=\arabic*.]
        \item Copy $x^i$ from the $i$-th channel to the $(c+1)$-th channel.
        \item Conduct convolution with kernel $C^{k,l}_{i,j}$ and the bias $0$ on the $(c+1)$-th channel for $k=1,2,\dots, d$.
        \end{enumerate}
    \item Add $\prod_{k=1}^d C^{k,l}_{i,j}x^i$ to the $(c+2)$-th channel and set the $(c+1)$-th channel to $0$. 
    \end{enumerate}
    \item Repeat the following for $l=1,2,\dots, \widetilde{m}_{j}$ to add $\boldsymbol{\delta}_j = \sum_{l=1}^{\widetilde{m}_{j}}\prod_{k=1}^d \widetilde{C}^{k,l}_{j} \boldsymbol{1}_d$ to the $(c+2)$-th channel.
    \begin{enumerate}[label*=\arabic*.]
        \item Conduct the convolution with kernel $(0,0,0)$ and the bias $1$ on the $(c+1)$-th channel and obtain $\boldsymbol{1}_d$ on the $(c+1)$-th channel.
        \item Conduct the convolution with kernel $\widetilde{C}^{k,l}_{j}$ and the bias $0$ on the $(c+1)$-th channel for $k=1,2,\dots, d$ and obtain $\prod_{k=1}^d \widetilde{C}^{k,l}_{j} \boldsymbol{1}_d$ in the $(c+1)$-th channel.
        \item Add $\prod_{k=1}^d \widetilde{C}^{k,l}_{j} \boldsymbol{1}_d$ to the $(c+2)$-th channel and set the $(c+1)$-th channel to $0$.
    \end{enumerate}
    \item Apply the activation function on the $(c+2)$-th channel and obtain $\sigma\left(\sum_{i=1}^{c} \sum_{l=1}^{m_{i,j}}\prod_{k=1}^d C^{k,l}_{i,j} x^i + \boldsymbol{\delta}_j \right)$ in the $(c+2)$-th channel.
\end{enumerate}
    
\item Add  $a_j\sigma\left(\sum_{i=1}^{c} \sum_{l=1}^{m_{i,j}}\prod_{k=1}^d C^{k,l}_{i,j} x^i + \boldsymbol{\delta}_j \right) $ to the $(c+3)$-th channel and set the $(c+2)$-th channel to $0$.
    \item Get $ \sum_{j=1}^n a_j\sigma\left(\sum_{i=1}^{c} \sum_{l=1}^{m_{i,j}}\prod_{k=1}^d C^{k,l}_{i,j} x^i + \boldsymbol{\delta}_j \right)$ in the $(c+3)$-th channel.
    \item Set the final convolutional layer with one output channel, which takes the value from the $(c+3)$-th channel.
\end{enumerate}
In this process, the $(c+1)$-th channel calculates the product $\prod_{k=1}^d C^{k,l}_{i,j}x^i$.
The $(c+2)$-th channel accumulates the summation $\sum_{i=1}^{c} \sum_{l=1}^{m_{i,j}}\prod_{k=1}^d C^{k,l}_{i,j} x^i$ from the $(c+1)$-th channel.
The $(c+3)$-th channel accumulates the final summation $\sum_{j=1}^n a_j\sigma\left(\sum_{i=1}^{c} \sum_{l=1}^{m_{i,j}}\prod_{k=1}^d C^{k,l}_{i,j} x^i + \boldsymbol{\delta}_j \right)$ after applying the activation function to the $(c+2)$-th channel. 
For the general case with an output channel size of $c_y$, repeating the above process while preserving the processed output components and using $c_x + c_y+2$ channels is sufficient to generate $c_y$ output vectors.

For a polynomial $\sigma$, the theorem is proved by Lemma \ref{lemma:polynomial}, and it completes the proof.
\end{nameproof}

\begin{lemma}\label{lemma:polynomial}
For a non-affine polynomial $p$, there exists a non-polynomial analytic function $\sigma$ such that the following relation holds for $d, c,c',w\in \mathbb{N}$, and $k\in \mathbb{N}_0$:
\begin{equation}
\overline{\Delta^{\sigma, d,2k+1}_{c,c',w}}
  \subset \overline{\Delta^{p, d,2k+1}_{c,c',w}}
\end{equation}
\end{lemma}
The proof of Lemma \ref{lemma:polynomial} can be found in \ref{appendix:proof_polynomial}.

The outcomes from the aforementioned lemma are not confined to CNNs; they can also help diminish the minimum width necessary for MLPs to have the UAP.
We define the set of $N$-layered MLPs with dimensions $\mathbf{l} = (l_0, l_1, \dots, l_N)\in \mathbb{N}^{N+1}$ and an activation function $\sigma$ as follows:
\begin{multline}
       \mathcal{M}^{N}_{\mathbf{l}} :=\{C_{N} \circ \sigma \circ C_{N-1}\circ \dots \circ \sigma \circ C_1:\mathbb{R}^{l_0}\rightarrow\mathbb{R}^{l_N}| 
       \\ 
     A_i\in \mathbb{R}^{l_{i}\times l_{i-1}}, b_i\in \mathbb{R}^{l_i},   C_i(x):= A_ix+b_i
     \}.
\end{multline}
And, define the set of deep, narrow MLPs $\mathcal{DM}^{\sigma}_{l,l', w}$ as follows:

\begin{equation}
    \mathcal{DM}^{\sigma}_{l,l', w}
    := \bigcup_{\substack{N\in\mathbb{N} \\
    l_1, l_2, \dots l_{N-1}\in [1, w]\\ 
    \mathbf{ls} = (l,l_1,l_2,\dots,l_{N-1}, l')}} 
    \mathcal{M}^{N}_{\mathbf{l}}.
\end{equation}
Then, the following corollary holds:
\begin{corollary}\label{corollary:polynomial_mlp}
For natural numbers $n,m\in \mathbb{N}$, a compact set $K\subset\mathbb{R}^n$, and a non-affine polynomial $p$, the following relation holds:
\begin{equation}
\overline{\mathcal{DM}^{p}_{n,m, n+m+1}} = C(K, \mathbb{R}^m).
\end{equation}
\end{corollary}
\begin{proof}
A CNN with a kernel size of one and $d=1$ essentially becomes an MLP:
\begin{equation}
    \Delta^{p, 1,1}_{c,c',w} = \mathcal{DM}^{p}_{c,c', w}.
\end{equation}
Thus, there exists a non-polynomial analytic activation function $\sigma$ such that the set of MLPs utilizing the activation function $\sigma$ is encompassed within those employing a polynomial activation function of the same channel size (width):
\begin{equation}
   \overline{ \mathcal{DM}^{\sigma}_{c,c', w} }= \overline{\Delta^{\sigma, 1,1}_{c,c',w}}
  \subset \overline{\Delta^{p, 1,1}_{c,c',w}} = \overline{\mathcal{DM}^{p}_{c,c', w}}.
\end{equation}
   According to Proposition 4.9 in \cite{kidger2020universal}, MLPs equipped with a non-polynomial activation function satisfying Condition \ref{condition:activation} and a width of $n+m+1$ have the UAP in the continuous function space:
   \begin{equation}
       C(K, \mathbb{R}^m) =\overline{  \mathcal{DM}^{\sigma}_{n,m, n+m+1} } \subset \overline{\mathcal{DM}^{p}_{n,m, n+m+1}} \subset { C(K, \mathbb{R}^m)}.
   \end{equation}
   This concludes the proof.
\end{proof}

\begin{remark}
It is worth noting that a width of $c_x+c_y+2$ is relatively small, especially in the context of common CNN usage.
For instance, in object segmentation tasks, which often involve using RGB (3 channels) as input and one output channel, the network requires a total of six channels ($6 = 3+1+2$) to achieve the UAP.
Considering that the minimum width required for MLPs with the same input and output shape to exhibit the UAP increases in proportion to the number of pixels in an image, this observation could partially explain the superior performance of CNNs compared to MLPs.
\end{remark}

\section{Universal Approximation Property of Multidimensional CNNs}\label{sec:multidimensional}
In this section, we extend our theory of the UAP to multidimensional CNNs.
The notations and proofs mirror those of the one-dimensional case.
However, considering the prevalent use of two-dimensional CNNs, we include the proof for completeness.

We denote multidimensional spatial dimensions or indices using bold symbols, such as $\boldsymbol{d} = (d_1, d_2, \dots, d_D)\in\mathbb{N}^D$.
Notation $\mathbb{R}^{\boldsymbol{d}}$ represents $\mathbb{R}^{d_1\times d_2\times \dots\times d_D}$, and  $\mathbb{R}^{c\times \boldsymbol{d}}$ represents $\mathbb{R}^{c\times d_1\times d_2\times \dots\times d_D}$.
We utilize multi-indexing as follows:
For an index $\boldsymbol{i}=(i_1, i_2, \dots ,i_D)$ and $x\in \mathbb{R}^{\boldsymbol{d}}$,
\begin{equation}
    x_{\boldsymbol{i}} := x(i_1, i_2, \dots, i_D),
\end{equation}
where $x$ is interpreted as a function from $[1, d_1]\times [1,d_2]\times \dots\times [1,d_D]$ to $\mathbb{R}$.
When operating on indices, an operation refers to a componentwise operation unless stated otherwise.
For instance, given $\boldsymbol{i} = (i_1,\dots, i_D)$ and $\boldsymbol{j} = (j_1,\dots, j_D)$, $\boldsymbol{i}+\boldsymbol{j} = (i_1+j_1,\dots, i_D+j_D)$. 
Moreover, $\boldsymbol{i}\leq \boldsymbol{j}$ indicates $i_t\leq j_t$ for all $t\in [1,D]$.
When a scalar value operates with an index vector, the scalar value is broadcasted to match the dimension.
For example, $\boldsymbol{i}+1$ implies $\boldsymbol{i}+\boldsymbol{1}_D =(i_1 + 1, i_2+1, \dots, i_D+1)$, $\boldsymbol{i}\leq 1$ means $i_k\leq 1$ for all $k\in [1,D]$, and $3\boldsymbol{1}_D$ denotes $(3,3,\dots, 3)$.

$\boldsymbol{1}_{\boldsymbol{d}}$ represents the tensor in $\mathbb{R}^{\boldsymbol{d}}$ with all components set to one.
$e^{\boldsymbol{d}}_{\boldsymbol{i}}$ or $e_{\boldsymbol{i}}$ denotes the tensor in $\mathbb{R}^{\boldsymbol{d}}$ where $\left(e^{\boldsymbol{d}}_{\boldsymbol{i}}\right)_{\boldsymbol{i}}$ equals one, and all other components are zero.
Similar to the one-dimensional case, $\odot$ is the Hadamard product (componentwise product), while $\oplus$ denotes concatenation along the channels. 
It is also used for concatenating indices: 
    For $\boldsymbol{i} = (i_1, \dots, i_n)$ and $\boldsymbol{j} = (j_1, \dots, j_m)$,
    $\boldsymbol{i}\oplus \boldsymbol{j} = (i_1, \dots, i_n, j_1, \dots, j_m)$.
  Let $\otimes$ denote a tensor (outer) product: for $x\in \mathbb{R}^{\boldsymbol{d}_1}$ and $y\in \mathbb{R}^{\boldsymbol{d}_2}$, an element of $x \otimes y\in \mathbb{R}^{\boldsymbol{d}_1 \oplus \boldsymbol{d}_2}$ can be identified as:
    \begin{equation}
        (x \otimes y)_{\boldsymbol{i} \oplus \boldsymbol{j}} := x_{\boldsymbol{i}}y_{\boldsymbol{j}}. 
    \end{equation}
   A tensor product of multiple tensors $x_1, x_2,\dots, x_n$ is denoted as 
    \begin{equation}
        \bigotimes_{i=1}^n x_i:= x_1\otimes x_2\otimes \dots \otimes x_n.
    \end{equation}

    Tensor products of linear transformations are defined as follows:
    For $i\in [1,n]$, vector spaces $V_i$, and linear transformations $T_i:V_i\rightarrow V_i$, the tensor product of linear transformations $\bigotimes_{i=1}^n T_i:\bigotimes_{i=1}^nV_i\rightarrow \bigotimes_{i=1}^nV_i$ is identified as: 
     \begin{equation}
         \left(\bigotimes_{i=1}^n T_i\right)\left(\bigotimes_{i=1}^n v_i\right):=\bigotimes_{i=1}^n\left(T_i(v_i)\right),
     \end{equation}
     for $v_i\in V_i$.
     Any other function values are determined by linear combinations.

\begin{table}[t]
    \centering
    \begin{threeparttable}
        \begin{tabular}{l|cc}
            \toprule
             & One-dimensional Case  & Multidimensional Case 
            \\
            \midrule
             UAP of Wide CNNs & Theorem \ref{thm:main_theorem} & Theorem \ref{thm:multidimensional_main_theorem} \\
            Inclusion Relation & Corollary \ref{Corollary:subset} & Corollary \ref{corollary:multidimensional_subset}\\
            Constant Functions in CNNs & Corollary \ref{lemma:onehot} & Lemma \ref{lemma:multidimensional_onehot}\\
            $d\geq 4$: UAP & Proposition \ref{lemma:main_theorem} & Proposition \ref{lemma:multidimensional_main_theorem} \\
            $d=3$: Counterexample for UAP & Proposition \ref{prop:d3_not_universal}& Proposition \ref{lemma:multidimesional_d3_not_universal}  \\
            $d=2, 3$: UAP & Lemma \ref{lemma: d23}& Lemma \ref{lemma:multidimensional_d23}  \\
            \midrule

            Spanning Linear Transformation & Lemma \ref{lemma:linear} & Lemma \ref{lemma:multidimensional_subspace}\\
           UAP of Deep, Narrow CNNs & Theorem \ref{theorem:narrow} & Theorem \ref{theorem:multidimensional_narrow}\\
            \bottomrule
        \end{tabular}
        \caption{ Correspondence of theorems, propositions, and lemmas between one-dimensional and multidimensional convolution neural network cases. $d$ means $\operatorname{max}_{i}d_i$ in the multidimensional case.}
        \label{table:correspondence}
    \end{threeparttable}
\end{table}

\begin{itemize}
    \item \textbf{Infinite-Length Convolution}: 
     For a natural number $D\in \mathbb{N}$, $\boldsymbol{k} = (k_1, k_2,\dots,k_D)\in \mathbb{N}^D$, consider $w\in\mathbb{R}^{2\boldsymbol{k}+1}$ as a $D$-dimensional kernel with size $2\boldsymbol{k}+1$.
    Then, $D$-dimensional convolution $f:\mathbb{R}^{\mathbb{Z}^D}\rightarrow\mathbb{R}^{\mathbb{Z}^D}$ with kernel $w$ is defined as:
    \begin{equation}
     f(x)_{\boldsymbol{i}} = \sum_{|\boldsymbol{j}|\leq \boldsymbol{k}} {w}_{\boldsymbol{j}}x_{\boldsymbol{i} - \boldsymbol{j}},
\end{equation}
where $|\cdot|$ denotes the componentwise absolute operation. 
\item \textbf{Zero-Padding Convolution}: Similar to the one-dimensional case, we define $\iota$ as the natural inclusion mapping from $\mathbb{R}^{\boldsymbol{d}}$ to $\mathbb{R}^{\mathbb{Z}^D}$ and $p$ as the natural projection mapping from $\mathbb{R}^{\mathbb{Z}^D}$ to $\mathbb{R}^{\boldsymbol{d}}$.
For an infinite-length convolution $g$ with kernel $w$, the zero-padding convolution $f$ with kernel $w$ is defined as:
\begin{equation}
    f:= p\circ g \circ \iota .
\end{equation}
We also denote $f$ as $w \circledast x$.

For $\boldsymbol{d}\in \mathbb{N}^D$ and $\boldsymbol{i}\in \mathbb{Z}^D$, we define $U^{\boldsymbol{d}}_{\boldsymbol{i}}$, the linear transformation from $\mathbb{R}^{\boldsymbol{d}}$ to $\mathbb{R}^{\boldsymbol{d}}$, as follows:
\begin{equation}
    \left(U^{\boldsymbol{d}}_{\boldsymbol{i}}(x)\right)_{\boldsymbol{l}} 
    = \begin{cases}
        x_{\boldsymbol{l} - \boldsymbol{i}} &\text{ if } \boldsymbol{1}_D \leq \boldsymbol{l} - \boldsymbol{i} \leq \boldsymbol{d} 
        \\ 0 &\text{ otherwise } 
    \end{cases}
\end{equation}
where $\boldsymbol{l} = (l_1, l_2, \dots, l_D)\in\prod_{k=1}^D[1,d_k]$.
$U^{\boldsymbol{d}}_{\boldsymbol{i}}$ can be interpreted as a translation operation shifting by $\boldsymbol{i}$.
Then, zero-padding convolution with kernel $w\in\mathbb{R}^{2\boldsymbol{k}+1}$ can be represented as:
\begin{equation}
    w\circledast x = \sum_{|\boldsymbol{i}|\leq \boldsymbol{k}} w_{\boldsymbol{i}}U^{\boldsymbol{d}}_{\boldsymbol{i}}(x).
\end{equation}
Furthermore, the equation
\begin{equation}
    U^{\boldsymbol{d}} _{\boldsymbol{i}}U^{\boldsymbol{d}} _{\boldsymbol{j}} =  U^{\boldsymbol{d}} _{\boldsymbol{i} + \boldsymbol{j}},
\end{equation}
holds if $\boldsymbol{i}\odot \boldsymbol{j}\geq 0$, where $\boldsymbol{i}$ and $\boldsymbol{j}$ have the same signs (including zero) for all corresponding components.

 \item \textbf{Zero-Padding Convolutional Layer:} 
  Define the set of multidimensional convolutional layers $\mathcal{L}^{\boldsymbol{d},2\boldsymbol{k}+1}_{c_1, c_2} \subset C({\mathbb{R}^{c_1\times \boldsymbol{d}}, \mathbb{R}^{c_2\times \boldsymbol{d}}})$ with $c_1$ input channels, $c_2$ output channels, and kernel size $2\boldsymbol{k}+1$ as 
\begin{equation}\label{eq:multidimensional_definition_convolutional_layer}
        \mathcal{L}^{\boldsymbol{d},2\boldsymbol{k}+1}_{c_1, c_2}:=\left\{ 
        f \left|
        w_{i,j}\in \mathbb{R}^{2\boldsymbol{k}+1},\delta_i\in\mathbb{R} ,f^{i}(x)=\sum_{j=1}^{c_1} w_{i,j} \circledast x^j + \delta_i \boldsymbol{1}_{\boldsymbol{d}} \right.\right\}.
        \end{equation}
Define $S^{\boldsymbol{d}}_{n}$ as:
\begin{equation}
    S^{\boldsymbol{d}}_{n} :=  \left\{\left.\sum_{i=1}^{N}\prod_{k=1}^n U^{\boldsymbol{d}}_{\boldsymbol{j}_{i,k}}\right| \boldsymbol{j}_{i,k}\in \mathbb{Z}^D, \left|\boldsymbol{j}_{i,k}\right|\leq 1, N\in \mathbb{N} \right\},
\end{equation}
where $\prod$ indicates the compositions of functions.
Furthermore, for $ \boldsymbol{n}, \boldsymbol{m}\in \mathbb{N}^D$, define the linear transformation $E^{\boldsymbol{d}}_{\boldsymbol{n},\boldsymbol{m}}:\mathbb{R}^{\boldsymbol{d}}\rightarrow \mathbb{R}^{\boldsymbol{d}}$ as:
\begin{equation}
E^{\boldsymbol{d}}_{\boldsymbol{n},\boldsymbol{m}}\left(e^{\boldsymbol{d}}_{\boldsymbol{i}} \right) =\begin{cases}
         e^{\boldsymbol{d}}_{\boldsymbol{n}} &\text{ if } \boldsymbol{i}= \boldsymbol{m}
        \\ 0 &\text{ otherwise}
    \end{cases}.
\end{equation}

\item \textbf{CNN:} 
For channel sizes $\boldsymbol{c} = (c_0, c_1, \dots, c_N)$, define the set of $N$-layered multidimensional CNNs with a kernel size $2\boldsymbol{k}+1$ as:
\begin{equation}
     \CNN{\sigma}{\boldsymbol{d}, 2\boldsymbol{k}+1}{N}{\mathbf{c}} :=\left.\left\{C_{N} \circ \sigma \circ C_{N-1}\circ \dots \circ \sigma \circ C_1:\mathbb{R}^{c_0\times \boldsymbol{d}}\rightarrow\mathbb{R}^{c_N\times \boldsymbol{d}}\right| C_{i}\in \mathcal{L}^{\boldsymbol{d}, 2\boldsymbol{k}+1}_{c_{i-1},c_i} \right\}.
\end{equation}
Then, define $\CNN{\sigma}{\boldsymbol{d}, 2\boldsymbol{k}+1}{N}{c,c'}$ as the set of all $N$-layered CNNs with $c$ input channels, $c'$ output channels, and kernel size $2\boldsymbol{k}+1$:
\begin{equation}
    \CNN{\sigma}{\boldsymbol{d}, 2\boldsymbol{k}+1}{N}{c,c'} := \bigcup_{\substack{c_1, c_2, \dots c_{N-1}\in \mathbb{N}\\ 
    \mathbf{c} = (c,c_1,c_2,\dots,c_{N-1}, c')}}  
    \CNN{\sigma}{\boldsymbol{d}, 2\boldsymbol{k}+1}{N}{\mathbf{c}}
\end{equation}

Define $\sCNN{\sigma}{\boldsymbol{d}, 2\boldsymbol{k}+1}{N}{c,c'}$ as:
 \begin{equation}
     \sCNN{\sigma}{\boldsymbol{d}, 2\boldsymbol{k}+1}{N}{c,c'} := \left\{\left. \sum^{n}_{i=1}a_i (\sigma \circ f_i):\mathbb{R}^{c\times \boldsymbol{d}}\rightarrow\mathbb{R}^{c'\times \boldsymbol{d}}\; \right|\; f_i\in \CNN{\sigma}{\boldsymbol{d}, 2\boldsymbol{k}+1}{N}{c,c'}, a_i\in \mathbb{R},n\in \mathbb{N}_0\right\}.
 \end{equation}
 Additionally, we define the set of deep, narrow CNNs $\Delta^{\sigma, \boldsymbol{d}, 2\boldsymbol{k}+1}_{c,c',l}$, or simply $\Delta_{c,c',l}$, as:
\begin{equation}
    \Delta^{\sigma, \boldsymbol{d}, 2\boldsymbol{k}+1}_{c,c',l}
    := \bigcup_{\substack{N\in\mathbb{N} \\
    c_1, c_2, \dots c_{N-1}\in [1, l]\\ 
    \mathbf{c} = (c,c_1,c_2,\dots,c_{N-1}, c')}}  
    \CNN{\sigma}{\boldsymbol{d}, 2\boldsymbol{k}+1}{N}{\mathbf{c}}.
\end{equation}
\end{itemize}

\subsection{Minimum Depth for the Universal Approximation Property of Multidimensional CNNs}
In this subsection, we present a multidimensional version of Theorem \ref{thm:main_theorem}, which closely resembles the original.
\begin{theorem}\label{thm:multidimensional_main_theorem}
For a natural number $D\in \mathbb{N}$, a spatial dimension $\boldsymbol{d}=(d_1,\dots, d_D)\in \mathbb{N}^D$, channel sizes $c,c'\in\mathbb{N}$, and a non-polynomial continuous activation function $\sigma$, the minimum depth $N_{\boldsymbol{d}}$ required for three-kernel, $D$-dimensional CNNs to have the UAP in the continuous function space from any arbitrary compact set $K\subset \mathbb{R}^{c\times \boldsymbol{d}}$ to $\mathbb{R}^{c'\times \boldsymbol{d}}$ is given by:
 \begin{equation}
N_{\boldsymbol{d}}=     \begin{cases}
    2 &\text{  if \quad} \operatorname{max}_{i}d_i=1,2
    \\ 3 &\text{  if \quad}  \operatorname{max}_{i}d_i=3
    \\ \operatorname{max}_{i}d_i-1 & \text{  if \quad}  \operatorname{max}_{i}d_i\geq 4
\end{cases}.
 \end{equation}
 In other words, the relation 
 \begin{equation}
     \oCNN{\sigma}{\boldsymbol{d}, 3\boldsymbol{1}_D}{N_{\boldsymbol{d}}}{c,c'} = C(K,\mathbb{R}^{c'\times \boldsymbol{d}})
 \end{equation}
 holds for an arbitrary compact set $K\subset \mathbb{R}^{c\times \boldsymbol{d}}$, and there exists a compact set $K\subset \mathbb{R}^{c\times \boldsymbol{d}}$ such that the following relation holds:
 \begin{equation}
     \oCNN{\sigma}{\boldsymbol{d}, 3\boldsymbol{1}_D}{N_{{\boldsymbol{d}}}-1}{c,c'} \nsupseteq C(K,\mathbb{R}^{c'\times \boldsymbol{d}}).
 \end{equation}
\end{theorem}

To prove Theorem \ref{thm:multidimensional_main_theorem}, we can check the fulfillment of Lemmas \ref{spacelemma}, \ref{lemma:product}, and \ref{lemma:activation_convolution} in the multidimensional scenario by substituting $\CNN{\sigma}{d, 2k+1}{N}{c,c'}$ with $\CNN{\sigma}{\boldsymbol{d},2\boldsymbol{k}+1}{N}{c,c'}$.
The subsequent task involves adapting spatial dimension-related proofs. 
Here, we provide the multidimensional versions and their proofs of Proposition \ref{prop:d3_not_universal}, Corollary \ref{Corollary:subset}, Corollary \ref{lemma:onehot}, Proposition \ref{lemma:main_theorem}, and Lemma \ref{lemma: d23}.
The correspondence between one-dimensional and multidimensional versions is outlined in Table \ref{table:correspondence}.
\begin{proposition} \label{lemma:multidimesional_d3_not_universal}
For a natural number $D\in \mathbb{N}$, channel sizes $c,c'\in \mathbb{N}$, a spatial dimension $\boldsymbol{d} = (d_1, d_2,\dots, d_D)\in \mathbb{N}^D$ satisfying $\operatorname{max}_{i\in [1,D]}d_i=3$, an open set $U\subset \mathbb{R}^{c\times \boldsymbol{d}}$ containing the origin, and a compact domain $K\subset \mathbb{R}^{c\times \boldsymbol{d}}$ containing $U$, three-kernel, two-layered, $D$-dimensional CNNs do not have the UAP as a function from $K$ to $\mathbb{R}^{c'\times \boldsymbol{d}}$.

In other words, the following relation holds:
\begin{equation}
    \oCNN{}{\boldsymbol{d},3\boldsymbol{1}_D}{2}{c,c'}\nsupseteq C(K, \mathbb{R}^{c'\times \boldsymbol{d}}).
\end{equation}
\end{proposition}
The proof of Proposition \ref{lemma:multidimesional_d3_not_universal} can be found in \ref{appendix:proof_multidimesional_d3_not_universal}.

\begin{corollary}\label{corollary:multidimensional_subset}
   For channel sizes $c,c'\in \mathbb{N}$, $l,l_0\in \mathbb{N}$ and a non-polynomial $C^{\infty}$ activation function $\sigma$, the following relations hold:
\begin{itemize}
    \item $\oCNN{}{}{l}{c,1}\subset \osCNN{\sigma}{}{l}{c,1}\subset \oCNN{}{}{l+1}{c,1}$.
    \item For $g\in \oCNN{}{}{l_0}{c, c'}$, $\boldsymbol{j}$ satisfying $|\boldsymbol{j}| \leq l\boldsymbol{1}_D$, $i\in [1,c']$, and $j\in [-l,l]$, define 
$f$ as $f:=U^{\boldsymbol{d}}_{\boldsymbol{j}} \circ g^i$. 
Then, $f\in \oCNN{}{}{l + l_0}{c,1}$.
\end{itemize}
\end{corollary}
The proof of Corollary \ref{corollary:multidimensional_subset} can be found in \ref{appendix:proof_multidimensional_subset}.

\begin{lemma}\label{lemma:multidimensional_onehot}
Consider a natural number $D\in \mathbb{N}$, channel sizes $c,c'\in \mathbb{N}$, and a spatial dimension $\boldsymbol{d} = (d_1, d_2,\dots, d_D)\in \mathbb{N}^D$.
 For $i\in [1,c]$ and $n\in \mathbb{N}$, let $V^1_{n,i}, V^2_{n,i}$, and $V^3_{n,i}$ represent sets of vectors in $\mathbb{R}^{d_i}$ defined as follows:
    \begin{equation}
        V^1_{n,i}:=\left\{ e^{d_i}_j \middle| j\in [1, n-1]\cup [d_i-n+2, d_i]\right\},
    \end{equation}
    \begin{equation}
        V^2_{n,i}:=\left\{ \sum^{d_i-n+1}_{i=n} e^{d_i}_i \right\},
    \end{equation}    
    and
    \begin{equation}
        V^3_{n,i}:=V^1_{n,i} \cup V^2_{n,i}.
    \end{equation}
    Then, for a non-polynomial $C^{\infty}$ activation function $\sigma$, the following relation holds for any vectors $v_i\in V^3_{n,i}$:
    \begin{equation}
        \bigotimes_{i=1}^D v_i\in \oCNN{\sigma}{\boldsymbol{d}, 3\boldsymbol{1}_D}{n}{c,1}.
    \end{equation}
\end{lemma}
The proof of Lemma \ref{lemma:multidimensional_onehot} can be found in \ref{appendix:proof_multidimensional_onehot}.

\begin{proposition}\label{lemma:multidimensional_main_theorem}
For a spatial dimension $\boldsymbol{d}=(d_1, \dots, d_D)$ satisfying
$d:=\operatorname{max}_i d_i \geq 4$, channel sizes $c,c\in\mathbb{N}$, a non-polynomial continuous activation function $\sigma$, and a compact domain $ K\in \mathbb{R^{c\times \boldsymbol{d}}}$, $(d-1)$-layered, three-kernel, $D$-dimensional CNNs have the UAP in the continuous function space from $K$ to $\mathbb{R}^{c'\times \boldsymbol{d}}$.

In other words, the following relation holds:
\begin{equation}
     \oCNN{\sigma}{\boldsymbol{d},3\boldsymbol{1}_D}{d-1}{c,c'} = C(K, \mathbb{R}^{c'\times\boldsymbol{d}}).
\end{equation} 
\end{proposition}
The proof of Proposition \ref{lemma:multidimensional_main_theorem} can be found in \ref{appendix:multidimensional_main_theorem}.

\begin{lemma}\label{lemma:multidimensional_d23}
  For a spatial dimension $\boldsymbol{d}$ satisfying $d:= \operatorname{max}_i {d_i} \in \{2,3\}$, channel sizes $c,c'\in\mathbb{N}$, a non-polynomial continuous activation function $\sigma$, and a compact domain $K\subset\mathbb{R}^{c\times \boldsymbol{d}}$, $d$-layered, three-kernel, $D$-dimensional CNNs have the UAP in the continuous function space from $K$ to $\mathbb{R}^{c'\times \boldsymbol{d}}$.

  In other words, the following relation holds:
  \begin{equation}
      \oCNN{\sigma}{\boldsymbol{d}, 3\boldsymbol{1}_D}{d}{c,c'} = C(K, \mathbb{R}^{c'\times \boldsymbol{d}}).
  \end{equation}
\end{lemma}
Proof of Lemma \ref{lemma:multidimensional_d23} can be found in \ref{appendix:proof_multidimensional_d23}.

\subsection{Minimum Width for the Universal Approximation Property of Multidimensional CNNs}
In this subsection, we present a multidimensional version of Theorem \ref{theorem:narrow}.
Lemmas \ref{lemma:activation_remove} and \ref{lem:remove_activation_multi} hold true for the multidimensional case.
Here, we provide the multidimensional version of Lemma \ref{lemma:linear}. The correspondence between the one-dimensional and multidimensional versions is outlined in Table \ref{table:correspondence}.

\begin{lemma}\label{lemma:multidimensional_subspace}
For a natural number $D\in\mathbb{N}$, a spatial dimension$\boldsymbol{d}=(d_1,\dots, d_D)\in \mathbb{N}^D$ where $d:=\operatorname{max}_{k\in [1,D]} d_k$, and an arbitrary linear transformation $L: \mathbb{R}^{\boldsymbol{d}}\rightarrow \mathbb{R}^{\boldsymbol{d}}$, it holds that $L\in S^{\boldsymbol{d}}_{n}$. 
\end{lemma}
The proof of Lemma \ref{lemma:multidimensional_subspace} can be found in \ref{appendix:proof_multidimensional_subspace}.
 
When replacing one-dimensional notations with their multidimensional counterparts, we derive the following theorem.
\begin{theorem}\label{theorem:multidimensional_narrow}
      For a natural number $D\in\mathbb{N}$, a spatial dimension $\boldsymbol{d}=(d_1,\dots, d_D)\in \mathbb{N}^D$, channel sizes $c_1,c_2\in \mathbb{N}$, a compact set $K\subset \mathbb{R}^{c_1\times \boldsymbol{d}}$, a kernel size $2\boldsymbol{k}+1\in \mathbb{N}^D$ satisfying $\operatorname{max}_i 2k_i+1\geq 3$, and a non-affine continuous activation function $\sigma$ that satisies
Condition \ref{condition:activation}, the following relation holds:
\begin{equation}
    \overline{\Delta^{\sigma, \boldsymbol{d}, 2\boldsymbol{k}+1}_{c_1,c_2,c_1 + c_2+ 2}} = C(K, \mathbb{R}^{c_2\times \boldsymbol{d}}).
\end{equation}

      
\end{theorem}

\section{Conclusion}
In this study, we investigate the UAP of CNNs under two specific configurations: limited depth with unlimited width, and limited width with unlimited depth.
Although our focus centers on investigating the UAP of three-kernel convolutions, we believe that this concept readily extends to other kernel sizes.
Additionally, convolution methods involving striding and dilation, along with the amalgamation of convolutional layers and pooling, emerge as intriguing research avenues in understanding the UAP of CNNs.
While our current research lacks a quantitative error estimate, the exploration of CNNs' approximation capabilities as tensor-to-tensor functions holds significant importance, especially in analyzing complex networks that integrate diverse architectural structures.
From this perspective, we anticipate that our findings, showcasing the UAP of CNNs as tensor-to-tensor functions, will establish a robust foundation for future investigations.

\section*{Acknowldegement}
This work was supported by a KIAS Individual Grant [AP092801] via the Center for AI and Natural Sciences at Korea Institute for Advanced Study. Myungjoo Kang was supported by the NRF grant [2021R1A2C3010887], the ICT R\&D program of MSIT/IITP[1711117093, 2021-0-00077, Artificial Intelligence Graduate School Program(SNU)]

\section*{Declaration of Competing Interest}
The authors declare that they have no known competing financial interests or personal relationships that could have appeared to influence the work reported in this paper.

\section*{Declaration of generative AI in scientific writing}
During the preparation of this work, the authors used ChatGPT in order to improve readability and language. After using this service, the authors reviewed and edited the content as needed and take full responsibility for the content of the publication.

\appendix
\newpage
\section{Notation Table}
\begin{center}
    \begin{table}[h]
        \begin{tabular}{l l}
            \textbf{Symbol} & \textbf{Description} \\
            \toprule
            $c$& Channel size\\
            $d$, $\boldsymbol{d}$& Spatial dimension\\
            $2k+1$, $2\boldsymbol{k}+1$& Kernel size\\
            $\boldsymbol{1}_d$, $\boldsymbol{1}_{\boldsymbol{d}}$& 1-vector in $\mathbb{R}^d$, 1-tensor in $\mathbb{R}^{\boldsymbol{d}}$  \\ 
            $e^d_{i}$, $e^{\boldsymbol{d}}_{\boldsymbol{i}}$& The $i$-th one-hot vector in $\mathbb{R}^d$, the $\boldsymbol{i}$-th one-hot tensor in $\mathbb{R}^{\boldsymbol{d}}$\\ 
            $\odot$&Hadamard product operator\\ 
            $\oplus$& Concatenation operator \\
            $M_{n,m}(\mathbb{R}^d)$& $n\times m$ Matrix consisting of $\mathbb{R}^d$ \\
            $\sigma$ & Activation function \\
              $\circledast$& Zero-Padding Convolution Operator\\
            $\mathcal{T}^d_s$& Set of linear convolutions (matrices)\\
            $U^d_{t}$& Translation function (matrix) by $t$ components\\
            $E^d_{n,m}$& Linear transformation(matrix) sending $e^d_m$ to $e^d_n$ and others to zero\\ 
            $S^d_N$& Set of compositions of $N$ linear convolutions\\ 
            $\mathcal{L}^{d,2k+1}_{c_1, c_2}$& Set of convolutional layers with $c_1$ input channels, $c_2$ output channels,\\
            & a spatial dimension $d$, and a kernel size $2k+1$\\
            $\CNN{\sigma}{d,2k+1}{N}{\boldsymbol{c}}$&  Set of CNNs with channel sizes $\boldsymbol{c}$\\
            $\CNN{\sigma}{d,2k+1}{N}{c_1, c_2}$& Set of $N$-layered CNNs with $c_1$ input channels and $c_2$ output channels\\
            $\sCNN{\sigma}{d,2k+1}{N}{c_1, c_2}$& Linear span of activation processed CNNs \\
            $\Delta^{\sigma,d,2k+1}_{c,c',l}$& Set of deep CNNs with $c$ input channels, $c'$ output channels,\\
            & and a maximum of $l$ intermediate channels\\
            $\otimes$&Tensor product operator\\
            \bottomrule
        \end{tabular}
    \end{table}
\end{center}
\section{Proofs in Section \ref{sec:lemma} and Section \ref{sec:cnn_depth}}
\subsection{Proof of Lemma \ref{spacelemma}}\label{appendix:proof_spacelemma}
\begin{proof}
\begin{enumerate}
    \item  For $f_1\in \CNN{}{}{N}{c,c'}$, there exists channel sizes $(c,c_1,c_2,\dots,c_{N-1},c')$ such that $f_1\in \CNN{}{}{N}{(c,c_1,c_2,\dots,c_{N-1},c')}$. 
    Represent $f_1$ as:
    \begin{equation}
        f_1 :=C_{N} \circ \sigma \circ C_{N-1}\circ \dots \circ \sigma \circ C_1,
    \end{equation}
    where $C_{i}\in \mathcal{L}^{d,2k+1}_{c_{i-1},c_i}$. Here, $c_0=c$, and $C_N=c'$.
    Similarly, let $f_2\in \CNN{}{}{N}{(c,c'_1,c'_2,\dots,c'_{N-1},c'')}$ be represented as
    \begin{equation}
        f_2 := C'_{N} \circ \sigma \circ C'_{N-1}\circ \dots \circ \sigma \circ C'_1,
    \end{equation}
    where $C'_{i}\in \mathcal{L}^{d,2k+1}_{c'_{i-1},c'_i}$. Here, $c'_0 = c$, and $C'_N=c''$. 
    Following Equation (\ref{eq:matrix_multiplication}), express $C_i$ as:
            \begin{equation}
        C_i(x) =  W_i\circledast x + \boldsymbol{\delta}_i,
        \end{equation}
    where $W_i\in M_{c_i, c_{i-1}}(\mathbb{R}^{2k+1})$ is the matrix of kernels, and $\boldsymbol{\delta}_i\in M_{c_i,1}(\mathbb{R}^d)$ is the vector of biases. 
    Similarly, represent $C'_i$ as
            \begin{equation}
        C'_i(x) =  W'_i \circledast x + \boldsymbol{\delta'}_i,
        \end{equation}    
        where $W'_i\in M_{c'_i, c'_{i-1}}(\mathbb{R}^{2k+1})$ is the matrix of kernels, and $\boldsymbol{\delta}'_i\in M_{c'_i,1}(\mathbb{R}^d)$ is the vector of biases.
    Then, define the concatenation $C''_i\in \mathcal{L}^{d,2k+1}_{c_{i-1}+c'_{i-1},c_i+c'_i}$ for $i = 2,3,\dots, N$ as:
    \begin{eqnarray}
        C''_i(x\oplus y) : = 
        \begin{bmatrix}
        W_i & 
        \\  &W'_i
        \end{bmatrix}\circledast\begin{bmatrix}
        x\\y
        \end{bmatrix}+\begin{bmatrix}
        \boldsymbol{\delta}_i\\ \boldsymbol{\delta'}_i
        \end{bmatrix}=C_i(x)\oplus C'_i(y) .
    \end{eqnarray}
 Define $C''_1\in \mathcal{L}^{d,2k+1}_{c,c_1+c'_1}$ as:
    \begin{equation}
        C''_1(x):=\begin{bmatrix}
         W_1\\W'_1
        \end{bmatrix}\circledast x +\begin{bmatrix}
        \boldsymbol{\delta}_1\\ \boldsymbol{\delta'}_1 \end{bmatrix} = C_1(x)\oplus C'_1(x).
    \end{equation}
      Then, construct $f\in \CNN{}{}{N}{(c,(c_1 + c'_1),(c_2 + c'_2),\dots,(c_{N-1}+ c'_{N-1}),(c'+ c''))}$ as:
      \begin{equation}
           f :=C''_{N} \circ \sigma \circ C''_{N-1}\circ \dots \circ \sigma \circ C''_1.
      \end{equation}
      Finally, it can be shown that:
     \begin{align}
         f(x)&= C''_{N} \circ \sigma \circ C''_{N-1}\circ \dots \circ \sigma \circ C''_1(x)
         \\ &= C''_{N} \circ \sigma \circ C''_{N-1}\circ \dots \circ \left(\sigma \circ C_1(x)\oplus \sigma \circ C'_1(x)\right)
          \\&= (C_{N} \circ \sigma \circ C_{N-1}\circ \dots \circ \sigma \circ C_1(x))\oplus  (C'_{N} \circ \sigma \circ C'_{N-1}\circ \dots \circ \sigma \circ C'_1(x))
         \\ &= f_1(x) \oplus f_2(x).
     \end{align}
     This completes the proof.
     
    \item For arbitrary $f_1, f_2\in \CNN{}{}{N}{c,c'}$, express $f_1$ and $f_2$ as:
    $ f_1 :=C_{N} \circ \sigma \circ C_{N-1}\circ \dots \circ \sigma \circ C_1$ and $f_2 := C'_{N} \circ \sigma \circ C'_{N-1}\circ \dots \circ \sigma \circ C'_1$.
    To explore scalar multiplication, substituting $C_N$ with $\alpha C_N $ in $f_1$ for $\alpha\in \mathbb{R}$ yields $\alpha f_1$.
   For the axiom that $\CNN{}{}{N}{c,c'}$ is closed under addition, construct $g $ by concatenating of $g_1:= C_{N-1}\circ \dots \circ \sigma \circ C_1$ and $g_2:=C'_{N-1}\circ \dots \circ \sigma \circ C'_1$; thus, $g=g_1 \oplus g_2$.
Given
\begin{equation}
        C_N(x) =  W \circledast x + \boldsymbol{\delta},
\end{equation}
and 
\begin{equation}
        C'_N(x) =  W' \circledast x + \boldsymbol{\delta'},
\end{equation}
  we can construct a convolutional layer with weights $\begin{bmatrix}
         W&W'
        \end{bmatrix}$ and bias $\boldsymbol{\delta}+ \boldsymbol{\delta'}$.
        This satisfies:
   \begin{equation}
       C''_N(x\oplus y) = \begin{bmatrix}
         W&W'
        \end{bmatrix}\circledast \begin{bmatrix}
        x\\y
        \end{bmatrix} + (\boldsymbol{\delta}+ \boldsymbol{\delta'})= C_N(x) + C'_N(y).
   \end{equation}
   Hence,
   \begin{equation}
       C''_N\circ \sigma\circ g(x) = C''_N \circ \sigma \circ (g_1\oplus g_2)(x) = C_N\circ\sigma\circ g_1(x) + C'_N\circ\sigma\circ g_2(x) = f_1(x)+f_2(x).
   \end{equation}
   This confirms that $f_1+f_2\in \CNN{}{}{N}{c,c'}$ establishing $\CNN{}{}{N}{c,c'}$ as a vector space.
   
    Regarding $\sigma(\CNN{}{}{N}{c,c'})$, it is evident from the definition of $\sigma(\CNN{}{}{N}{c,c'})$.

    \item Given $\CNN{}{}{N}{c,c'}$ as a vector space, $\frac{f_{\theta + \epsilon}(x) - f_{\theta}(x)}{\epsilon} \in\CNN{}{}{N}{c,c'}$. 
    And because 
    \begin{equation}
        \left\|\frac{f_{\theta + \epsilon}(x) - f_{\theta}(x)}{\epsilon} - \frac{\partial}{\partial \theta}f_{\theta}(x)\right\|_{\infty,K} = \frac{\epsilon}{2} \left\|\frac{\partial^2}{\partial^2\theta} f_{\theta + t}(x)\right\|_{\infty,K},
    \end{equation}
    for $|t|<|\epsilon|$, it uniformly converges to zero within the compact domain $x\in K$. Therefore, $\frac{\partial}{\partial \theta}f_{\theta}(x)
    \in  \oCNN{}{}{N}{c,c'} $. 
    A similar rationale applies to $\osCNN{\sigma}{}{N}{c,c'}$.
    \item If $f\in \osCNN{\sigma}{}{N}{c,c'}$, there exist $f_i\in \sigma(\CNN{}{}{N}{c,c'})$, such that $f_i\xrightarrow[]{i\rightarrow\infty} f$. 
    Then, 
    \begin{equation}
        f_i = \sum^{n_i}_{j=1}a_{i,j}(\sigma \circ f_{i,j}) ,
    \end{equation}
    where $f_{i,j}\in \CNN{}{}{N}{c,c'}$ and $a_{i,j}\in \mathbb{R}$.
    Decompose $C$ into $C = L + \boldsymbol{\delta}$ where $L$ is the linear transformation and $\boldsymbol{\delta}$ is the bias:
    \begin{equation}
        C\circ f_i = (L + \boldsymbol{\delta})\circ \sum^{n_i}_{j=1}a_{i,j}(\sigma \circ f_{i,j}) =   \boldsymbol{\delta} + \sum^{n_i}_{j=1} a_{i,j}L\circ\sigma \circ f_{i,j} \in 
        \CNN{}{}{N+1}{c, c''},
    \end{equation}
    because $\CNN{}{}{N+1}{c, c''}$ is a vector space.
    If $\{f_i\}_{i\in\mathbb{N}}$ uniformly converge to $f$, then $\{C\circ f_i\}_{i\in\mathbb{N}}$ uniformly converge to $C\circ f$ for the $C^{\infty}$ continuous function $C$ in the compact domain, completing the proof.

\end{enumerate}
\end{proof}
\subsection{Proof of Proposition \ref{prop:d3_not_universal}}\label{appendix:proof_d3_not_universal}
\begin{proof}
It suffices to prove the case for $c'=1$ because $\oCNN{\sigma}{d,3}{2}{c,1}$ is the projection of $\oCNN{\sigma}{d,3}{2}{c,c'}$ to a single channel. 
Therefore, $\oCNN{}{}{2}{c,c'}\nsupseteq C(K, \mathbb{R}^{c'\times 3})$ if $\oCNN{}{}{2}{c,1}\nsupseteq C(K, \mathbb{R}^{3})$.
Consider a two-layered CNN $f\in \CNN{\sigma}{d,3}{2}{(c,n,1)}$ with $n$ intermediate channels.
This $f$ can be represented as the composition of convolutional layers $C_1\in \mathcal{L}^{3}_{c, n}$ and $C_2\in \mathcal{L}^{3}_{n, 1}$:
\begin{equation}
    f = C_2\circ\sigma\circ C_1.
\end{equation}
Let $C_1$ have a kernel matrix $(a^{i,j})_{1\leq i\leq n, 1\leq j\leq c}$ where $a^{i,j} = (a^{i,j}_{-1}, a^{i,j}_{0}, a^{i,j}_{1})\in \mathbb{R}^3$ and biases $\delta_i\in \mathbb{R}$.
Similarly, the convolutional layer $C_2$ has kernels $b^{i} = (b^{i}_{-1}, b^{i}_{0}, b^{i}_{1})\in\mathbb{R}^3$ and bias $\delta_0$.
Now, for $x\in\mathbb{R}^{c\times 3}$, the $i$-th channel $C^i_1(x)$ of $C_1(x)$ becomes:
\begin{align}
   & C^i_1(x) = \sum_{j=1}^c a^{i,j}\circledast x^j + \delta_i \boldsymbol{1}_3
    \\&= \left(\sum_{j=1}^c a^{i,j}_{0}x^j_1 + a^{i,j}_{-1}x^j_2 ,
    \sum_{j=1}^c a^{i,j}_{1}x^j_1 + a^{i,j}_{0}x^j_2 + a^{i,j}_{-1}x^j_3 ,
    \sum_{j=1}^c a^{i,j}_{1}x^j_2 + a^{i,j}_{0}x^j_3 
    \right) + \delta_i \boldsymbol{1}_3,
\end{align}
where $x^j = (x^j_1, x^j_2, x^j_3)\in\mathbb{R}^3$.
Let $f=C_2\circ \sigma \circ C_1  $ be represented as $f= (f_1, f_2, f_3)\in \mathbb{R}^3$. We obtain the following expressions:
\begin{multline}
    f_1(x)     =\sum_{i=1}^n b^i_{0} \sigma\left(\sum_{j=1}^c a^{i,j}_{0} x^j_1 + a^{i,j}_{-1} x^j_2 + \delta_i\right) \\+ b^i_{-1} \sigma\left(\sum_{j=1}^c a^{i,j}_{1} x^j_1 +a^{i,j}_{0} x^j_2 + a^{i,j}_{-1} x^j_3+ \delta_i\right)+ \delta_0,
\end{multline}
\begin{multline}
    f_2(x) = \sum_{i=1}^n b^i_{1} \sigma\left(\sum_{j=1}^c a^{i,j}_{0} x^j_1 + a^{i,j}_{-1} x^j_2 + \delta_i\right) \\+ b^i_{0} \sigma\left(\sum_{j=1}^c a^{i,j}_{1} x^j_1 +a^{i,j}_{0} x^j_2 + a^{i,j}_{-1} x^j_3+ \delta_i\right)+ b^i_{-1} \sigma\left(\sum_{j=1}^c a^{i,j}_{1} x^j_2 +a^{i,j}_{0} x^j_3 + \delta_i\right) + \delta_0,
\end{multline}
    Consider two input values, $z,w\in \mathbb{R}^{c\times 3}$ defined as  $z^j = (x^j_1, x^j_2,0)\in \mathbb{R}^3$ and $w^j = (0, x^j_1, x^j_2)\in \mathbb{R}^3$ for $j\in [1,c]$.
     $w$ is the translation of $z$ and vice versa.
    Then, the following equation holds:
    \begin{multline}
  f_1(z) - f_2(w) =
    \left(\sum_{i=1}^n b^i_{0} \sigma\left(\sum_{j=1}^c a^{i,j}_{0} x^j_1 + a^{i,j}_{-1} x^j_2 + \delta_i\right) + b^i_{-1} \sigma\left(\sum_{j=1}^c a^{i,j}_{1} x^j_1 +a^{i,j}_{0} x^j_2 +\delta_i\right)\right)
      \\ - \left( \sum_{i=1}^n b^i_{1} \sigma\left(\sum_{j=1}^c a^{i,j}_{-1} x^j_1 + \delta_i\right) + b^i_{0} \sigma\left(\sum_{j=1}^c a^{i,j}_{0} x^j_1 + a^{i,j}_{-1} x^j_2+ \delta_i\right) \right.
      \\ \left.+ b^i_{-1} \sigma\left(\sum_{j=1}^c a^{i,j}_{1} x^j_1 +a^{i,j}_{0} x^j_2 + \delta_i\right)\right)
    .
    \end{multline}
    This simplifies to:
    \begin{equation}
        f_1(z) - f_2(w) = - \sum_{i=1}^n b^i_{1} \sigma\left(\sum_{j=1}^c a^{i,j}_{-1} x^j_1 + \delta_i\right).
    \end{equation}
   Observing that $ f_1(z) - f_2(w)$ is a function of variables $x^j_1$, let $x_1: = (x^1_1, x^2_1, \dots, x^c_1)\in \mathbb{R}^c$.
   Define $h:\mathbb{R}^c\rightarrow\mathbb{R}$ as:
    \begin{equation}
        h(x_1): = f_1(z) - f_2(w).
    \end{equation}
Let the compact set $K$ encompass the open cube $(-\epsilon_0, \epsilon_0)^{c\times 3}$ and consider sufficiently small $x^j_1, x^j_2$ such that $z,w\in K$.
    Additionally, define $g\in C(K ,\mathbb{R}^3)$ as:
    \begin{equation}
        g(y) =(g_1(y), g_2(y), g_3(y)):= (y^{i_0}_2, 0, 0)
    \end{equation}
    for $y\in \mathbb{R}^{c\times 3}$ and $i_0\in [1,c]$.
    Then, this equation holds:
    \begin{equation}
        |(f_1 -g_1)(z) -(f_2-g_2)(w)| = |h(x_1)-x^{i_0}_2|.
    \end{equation}
    If $g\in \oCNN{}{}{2}{c, 1}$, there exists $f\in \CNN{}{}{2}{c,1}$ such that,
    \begin{equation}
        ||f-g||_{\infty,K}< \frac{\epsilon_0}{4}.
    \end{equation}
    This implies that $|(f_1 -g_1)(z)|<\frac{\epsilon_0}{4}$ and $|(f_2-g_2)(w)|<\frac{\epsilon_0}{4}$ for arbitrary $z,w\in(-\epsilon_0, \epsilon_0)^{c\times 3}$.
    However,
    \begin{equation}
    \begin{aligned}
        |h(x_1)-x^{i_0}_2| = |(f_1 -g_1)(z) &-(f_2-g_2)(w)|
        \\&< |(f_1 -g_1)(z)| + |(f_2-g_2)(w)| < \frac{\epsilon_0}{2},
    \end{aligned}
    \end{equation}
This equation holds for $x^{i_0}_2 \in \{ -0.9\epsilon_0, 0.9\epsilon_0\} $ and any arbitrary $x_1\in (-\epsilon_0, \epsilon_0)^c$.
This leads to a contradiction, completing the proof.
\end{proof}

\subsection{Proof of Lemma \ref{lemma:activation_convolution}} \label{appendix:proof_activation_convolution}
\begin{proof}
Utilize mathematical induction on $N$.
For $N=1$, $\CNN{\sigma*\varphi}{d, 2k+1}{1}{c,c'} = \CNN{\sigma}{d,2k+1}{1}{c,c'} = \mathcal{L}^{d,2k+1}_{c,c'}$ satisfying the induction hypothesis.
Assume that the induction hypothesis holds for $N = N_0$; i.e., $\oCNN{\sigma*\varphi}{d,2k+1}{N_0}{c,c'}\subset\oCNN{\sigma}{d,2k+1}{N_0}{c,c'}$.
Because $\varphi$ has a compact support, for a positive $\epsilon>0$, there exist $n\in \mathbb{N}$, $i\in [1,n]$, and $x_i\in \mathbb{R}$ such that:
\begin{equation}
    \left\|\sigma * \varphi(x) - \sum_{i=1}^n \sigma(x-x_i)\varphi(x_i)\right\|_{\infty, \mathbb{R}}<\epsilon.
\end{equation}
We will first prove: 
\begin{equation}
    \ossCNN{\sigma*\varphi}{d,2k+1}{N_0}{c,c'}
    \subset 
    \ossCNN{\sigma}{d,2k+1}{N_0}{c,c'}
\end{equation}
For $f\in \ssCNN{\sigma*\varphi}{d,2k+1}{N_0}{c,c'}$, there exist $n\in\mathbb{N}_0$, $i\in [1,n]$, $a_i\in \mathbb{R}$, and $f_i\in \CNN{\sigma*\varphi}{d,2k+1}{N_0}{c,c'}$ such that 
\begin{equation}
    f = \sum^{n}_{i=1}a_i \left((\sigma*\varphi) \circ f_i\right).
\end{equation}
Let $A = 1 + \sum_{i=1}^n {|a_i|} $.
Then, there exist $m\in \mathbb{N}$, $j\in [1,m]$, and $x_j\in \mathbb{R}$ satisfying:
\begin{equation}
    \left\|\sigma * \varphi(x) - \sum_{j=1}^m \sigma(x-x_j)\varphi(x_j)\right\|_{\infty, \mathbb{R}}<\frac{\epsilon}{A}.
\end{equation}
Define $g$ as:
\begin{equation}
    g:= \sum^{n}_{i=1}a_i \sum_{j=1}^m \varphi(x_j)\left( \sigma(f_i - x_j)\right).
\end{equation}
Because $f_i\in  
\CNN{\sigma*\varphi}{d,2k+1}{N_0}{c,c'}
\subset \oCNN{\sigma}{d,2k+1}{N_0}{c,c'}$ by the induction hypothesis, $g\in \ossCNN{\sigma}{d,2k+1}{N_0}{c,c'}$.
Additionally, 
\begin{align}
    \|f-g\|_{\infty,K} &= \left\|\sum^{n}_{i=1}a_i \left((\sigma*\varphi) \circ f_i - \sum_{j=1}^m 
    \varphi(x_j)\left( \sigma(f_i - x_j)\right)
    \right)\right\|_{\infty,K}
    \\ &< \sum_{i=1}^n |a_i|\left\|\left((\sigma*\varphi) \circ f_i - \sum_{j=1}^m 
    \varphi(x_j)\left( \sigma(f_i - x_j)\right)\right)\right\|_{\infty,K} <\frac{ \sum_{i=1}^n |a_i| \epsilon}{A} < \epsilon.
\end{align}

Thus, $f\in \ossCNN{\sigma}{d,2k+1}{N_0}{c,c'}$.
Because $f$ is an arbitrary element of $\ssCNN{\sigma*\varphi}{d,2k+1}{N_0}{c,c'}$,
$\ossCNN{\sigma*\varphi}{d,2k+1}{N_0}{c,c'} \subset \ossCNN{\sigma}{d,2k+1}{N_0}{c,c'}$.
Now, our aim is to demonstrate:
\begin{equation}
\oCNN{\sigma*\varphi}{d,2k+1}{N_0+1}{c,c'}
\subset \oCNN{\sigma}{d,2k+1}{N_0+1}{c,c'}.
\end{equation}
Take $f\in \CNN{\sigma*\varphi}{d,2k+1}{N_0+1}{c,c'}$.
Then, there exist $C\in \mathcal{L}^{d,2k+1}_{c'',c'}$ and $f_0\in \CNN{\sigma*\varphi}{d,2k+1}{N_0}{c,c''}$ such that:
\begin{equation}
 f= C\circ (\sigma*\varphi) \circ f_0.
\end{equation}
Because $(\sigma*\varphi) \circ f_0\in \ossCNN{\sigma*\varphi}{d,2k+1}{N_0}{c,c'} \subset \ossCNN{\sigma}{d,2k+1}{N_0}{c,c'}$, and leveraging Lemma \ref{spacelemma}, $ f= C\circ\left( (\sigma*\varphi) \circ f_0\right)\in \oCNN{\sigma}{d,2k+1}{N_0+1}{c,c'}$.
This satisfies the induction hypothesis for $N = N_0+1$, concluding the proof.
\end{proof}

\subsection{Proof of Proposition \ref{lemma:main_theorem}}\label{appendix:proof_main_theorem}
\begin{proof}
We observe that it suffices to prove the case when $c'=1$: 
$\oCNN{}{}{d-1}{c,1} = C(K, \mathbb{R}^{1\times d})$.
This follows from the function concatenation allowed by Lemma \ref{spacelemma}.
The trajectory of the proof aligns with the concept outlined by \citet{leshno1993multilayer}.
The primary goal is to approximate all polynomials, enabling the approximation of all continuous functions within a compact domain via the Stone--Weierstrass theorem \citep{de1959stone}.
A significant distinction lies in independently considering multivariate polynomials for each position within the output vector.
The convolution operation introduces complexity, which constitutes the primary challenge in solving this problem.


For an input vector $x = (x^1, x^2, \dots , x^c)\in \mathbb{R}^{c\times d}$, the translation of $x^i = (x^i_1, x^i_2, \dots, x^i_d)$ is defined as follows:
\begin{equation}
    p^i_{-j} := U_{j}x^i= (0, \dots, 0, x^i_1, x^i_2, \dots, x^i_{d-j}),
\end{equation}
\begin{equation}
    p^i_{0} := x^i = (x^i_1, x^i_2, \dots, x^i_d),
\end{equation}
and 
\begin{equation}
    p^i_{j} :=  U_{-j}x^i = (x^i_{j+1}, \dots, x^i_{d-1}, x^i_d, 0, \dots, 0 ),
\end{equation}
where $U_j$ is the translation matrix defined in Eq (\ref{eq:u_matrix}), and $x^i$ is interpreted as a column vector.
The $k$-th component $\left(p^i_{j}\right)_k$ of $p^i_j$ becomes:
\begin{equation}
    \left(p^i_{j}\right)_k = \begin{cases}
        x^i_{j+k} &\text{ if } 1\leq j+k\leq d\\
        0 &\text{ otherwise}
    \end{cases}.
\end{equation}
We initially consider a $C^{\infty}$ continuous activation function $\sigma$.

\paragraph{Case 1. $d=4$:} 
First, we aim to demonstrate that for a monomial $M_1$ composed of variables $x^i_1, x^i_2$, and $ x^i_3$, excluding $x^i_4$, the vector $(M_1, 0, 0, 0)$ belongs to $ \osCNN{\sigma}{}{2}{c,1}$.
More precisely, $M_1$ is defined as:
\begin{equation}
    M_1 = \prod_{i=1}^c \prod_{j=1,2,3} (x^{i}_{j})^{\alpha_{i,j}},
\end{equation}
where $\alpha_{i,j}\in \mathbb{N}_0$.
According to Corollary \ref{Corollary:subset}, $p^i_j = U_{-j}x^i\in \oCNN{}{}{2}{c,1}$ for $j\in [-2,2]$. 
By Corollary \ref{lemma:onehot}, $e_1 \in  \oCNN{}{}{2}{c,1}$. 
Then, applying Lemma \ref{lemma:product}, we obtain:
\begin{equation}
    e_1 \odot \left(\bigodot^c_{i=1}\bigodot_{j=0,1,2} (p^i_{j})^{\alpha_{i,j+1}} \right)
    \in  \osCNN{\sigma}{}{2}{c,1}.
\end{equation}
The first component of $\bigodot^c_{i=1}\bigodot_{j=0,1,2} (p^i_{j})^{\alpha_{i,j+1}}$ is calculated as
\begin{equation}
    \left(\bigodot^c_{i=1}\bigodot_{j=0,1,2} (p^i_{j})^{\alpha_{i,j+1}}\right)_1 = \prod^c_{i=1}\prod_{j=0,1,2} (p^i_{j})_1^{\alpha_{i,j+1}} = \prod^c_{i=1}\prod_{j=1,2,3} (x^i_{j})^{\alpha_{i,j}} = M_1.
\end{equation}
Therefore,
\begin{equation}
    e_1 \odot \left(\bigodot^c_{i=1}\bigodot_{j=0,1,2} (p^i_{j})^{\alpha_{i,j+1}} \right) = (M_1,0,0,0)\in \osCNN{\sigma}{}{2}{c,1}.
    \end{equation}

Similarly, when considering a monomial $M_2$ composed of $x^i_2, x^i_3, x^i_4$, excluding $x^i_1$, $(0, 0, 0, M_2)$ becomes an element of $\osCNN{\sigma}{}{2}{c,1}$.
In other words, for $\alpha_{i,j}\in \mathbb{N}_0$,
\begin{equation}
    \left(0,0,0,\prod_{i=1}^c \prod_{j=2,3,4} (x^{i}_{j})^{\alpha_{i,j}}\right)\in \osCNN{\sigma}{}{2}{c,1}.
\end{equation}
The proof is evident due to symmetry.

Next, we aim to prove that for a monomial $M_3$ containing at least one $x^{i}_4$ for some $i\in[1,c]$, $(0, M_3, 0,0 )$ is an element of $\osCNN{\sigma}{}{2}{c,1}$.
In other words, for $M_3$ defined as:
\begin{equation}
    M_3 = x^{i_0}_4 \prod_{i=1}^c \prod_{j=1,2,3,4} (x^{i}_{j})^{\alpha_{i,j}},
\end{equation}
$(0, M_3, 0,0 )\in \osCNN{\sigma}{}{2}{c,1}$ where $\alpha_{i,j}\in \mathbb{N}_0$, and $i_0\in [1, c]$.

By Corollary \ref{lemma:onehot}, $e_1, e_4 \in  \oCNN{}{}{2}{c,1}$.
Additionally, $\boldsymbol{1}_4 = e_1+e_2+e_3+e_4\in  \oCNN{}{}{2}{c,1}$.
Therefore, $\left(e_1+e_2+e_3+e_4\right) - e_1 - e_4 = e_2 + e_3\in \oCNN{}{}{2}{c,1}$.
Then, we have
\begin{equation}
      \left( e_2 + e_3 \right)\odot  \left(p^{i_0}_2 \odot \bigodot^c_{i=1}\bigodot_{j=-1,0,1,2}(p^{i}_{j})^{\alpha_{i,j+2}} \right) \in \osCNN{\sigma}{}{2}{c,1}.
\end{equation}
Because the third component of $p^{i_0}_2$ is zero, only the second component is nonzero.
Furthermore, the second component can be calculated as
\begin{multline}
         \left(p^{i_0}_2 \odot \bigodot^c_{i=1}\bigodot_{j=-1,0,1,2}(p^{i}_{j})^{\alpha_{i,j+2}} \right)_2 
     = (p^{i_0}_2)_2 \times \prod^c_{i=1}\prod_{j=-1,0,1,2}(p^{i}_{j})_2^{\alpha_{i,j+2}} 
     \\= x^{i_0}_4 \prod^c_{i=1}\prod_{j=1,2,3,4} (x^i_j)^{\alpha_{i,j}} = M_3.
\end{multline}
Therefore, 
\begin{equation}
     \left( e_2 + e_3 \right)\odot  \left(p^{i_0}_2 \odot \bigodot^c_{i=1}\bigodot_{j=-1,0,1,2}(p^{i}_{j})^{\alpha_{i,j+2}} \right) = (0,M_3,0,0)\in \osCNN{\sigma}{}{2}{c,1}.
\end{equation}

Similarly, applying a symmetrical argument demonstrates that for a monomial $M_4$ containing at least one $x^i_1$, $(0,0, M_4,0)$ is an element of $\osCNN{\sigma}{}{2}{c,1}$.
So far, we have established: 
\begin{itemize}
    \item If a monomial $M_1$ does not contain any $x^i_4$, then $(M_1, 0,0,0)\in \osCNN{\sigma}{}{2}{c,1}$,
    \item if a monomial $M_2$ does not contain any $x^i_1$, then $(0,0,0, M_2)\in \osCNN{\sigma}{}{2}{c,1}$,
    \item for a monomial $M_3$ containing at least one $x^i_4$, $(0,M_3, 0,0)\in \osCNN{\sigma}{}{2}{c,1}$,
    \item and for a monomial $M_4$ containing at least one $x^i_1$, $(0,0, M_4,0)\in \osCNN{\sigma}{}{2}{c,1}$.
\end{itemize}
Here, we will prove that for any arbitrary monomial $M_0$, $(M_0,0,0,0)$, $(0, M_0,0,0)$, $(0, 0, M_0,0)$, and $(0, 0,0,M_0)$ are in $ \oCNN{}{}{3}{c,1}$.
By Lemma \ref{spacelemma}, for an arbitrary convolutional layer $C\in \mathcal{L}^{3}_{1,1}$ and a function $f\in \osCNN{\sigma}{}{2}{c,1}$, $C(f)\in \oCNN{}{}{3}{c,1} $.
When a monomial $M$ contains at least one $x^i_4$ for $i\in [1,c]$, $(0,M, 0,0)\in \osCNN{\sigma}{}{2}{c,1} $. 
Additionally, considering $C(x) = U_{0}x$, $C((0,M, 0,0)) = (0,M, 0,0)\in \oCNN{}{}{3}{c,1} $, and for $C(x) = U_{-1}x$, $C((0,M, 0,0)) = (M,0, 0,0)\in \oCNN{}{}{3}{c,1} $.
If a monomial $M$ does not contain any $x^i_4$, $(M, 0,0,0)\in \osCNN{\sigma}{}{2}{c,1}$. 
Moreover, for $C(x) = U_{0}x$, $C((M,0, 0,0)) = (M,0, 0,0)\in \oCNN{}{}{3}{c,1} $, and for $C(x) = U_{1}x$, $C((M,0, 0,0)) = (0,M, 0,0)\in\oCNN{}{}{3}{c,1} $.
Thus, for any arbitrary monomial $M$, $(M, 0,0,0)$ and $(0,M,0,0)$ are elements of $\oCNN{}{}{3}{c,1} $.
Furthermore, by symmetry, $(0,0,M, 0)$ and $(0,0,0,M)$ are in $\oCNN{}{}{3}{c,1} $. 
This completes the proof for the case of $d=4$.

\paragraph{Case 2. $d\geq 5$:}

The progression of the proof closely follows Case 1, with the distinction lying in the ability to construct the projection $e_k\odot$ for all $k\in [1,d]$ when $d\geq 5$. 
By Corollary \ref{lemma:onehot}, for $t\in [1,d-3] \cup [4,d] $, $e_t$ belongs to $\oCNN{}{}{d-2}{c,1}$.
When $d >5$, $[1,d-3] \cup [4,d] = [1,d]$. 
 For $d=5$, $[1,d-3] \cup [4,d] = \{ 1,2,4,5\}$, and because $ e^5_3 = \boldsymbol{1}_5 - \sum_{t=1,2,4,5}$, $e^5_t \in \oCNN{}{}{d-2}{c,1}$ for all $t\in [1,5]$.
 Therefore, $e_t\in \oCNN{}{}{d-2}{c,1}$ for all $d\geq 5$ and $t\in [1,d]$. 

For an arbitrary monomial $M = \prod_{i=1}^c\prod_{j=1}^d (x^i_j)^{\alpha_{i,j}}$, we will demonstrate that for $t\in [2,d-1]$, the vector $Me_t$ is in $\osCNN{\sigma}{}{d-2}{c,1}$:
\begin{equation}
    Me_t = (0,\dots,0,M,0,\dots,0)\in \osCNN{\sigma}{}{d-2}{c,1}.
\end{equation}
Because $j-t\in [2-d, d-2]$ for $j\in [1,d]$ and $t\in [2,d-1]$, we conclude that $p^i_{j-t}\in \oCNN{}{}{d-2}{c,1}$.
By Lemma \ref{lemma:product}, we deduce:
\begin{equation}
     e_t \odot \left(\bigodot_{i=1}^c \bigodot_{j=1}^{d} (p^{i}_{j-t})^{\alpha_{i,j}}\right)\in \osCNN{\sigma}{}{d-2}{c,1}.
\end{equation}
This leads to:
\begin{align}
     e_t \odot \left(\bigodot_{i=1}^c \bigodot_{j=1}^{d} (p^{i}_{j-t})^{\alpha_{i,j}}\right) = \bigodot_{i=1}^c \bigodot_{j=1}^{d} (e_t\odot p^{i}_{j-t})^{\alpha_{i,j}} = \bigodot_{i=1}^c \bigodot_{j=1}^{d} (x^i_j e_t)^{\alpha_{i,j}}
     \\ =\prod_{i=1}^c \prod_{j=1}^{d} (x^i_j)^{\alpha_{i,j}} e_t = Me_t.
\end{align}
Therefore, $Me_t\in \osCNN{\sigma}{}{d-2}{c,1}$ for $t\in [2,d-1]$.

Finally, employing suitable $U_1x, U_0x, U_{-1}x\in \mathcal{L}^3_{1,1}$ for the last convolutional layer, we derive $Me_t\in \oCNN{}{}{d-1}{c,1}$ for all $t\in [1,d]$.
This completes the proof for a non-polynomial $C^{\infty}$ activation function $\sigma$.

To extend this result to general non-polynomial functions, we employ the following lemma and Lemma \ref{lemma:activation_convolution}.

\begin{lemma}
For any non-polynomial continuous function $\sigma$, there exists a compactly supported $C^{\infty}$ function $\varphi$ such that $\sigma * \varphi$ is a non-polynomial smooth function.

\end{lemma}
\begin{proof}
    It is an immediate consequence of Steps 6 and 7 in Section 6 of \cite{leshno1993multilayer}.
\end{proof}
The lemma implies  the existence of a function $\varphi$ such that $\oCNN{\sigma*\varphi}{d,2k+1}{N}{c,c'} = C(K,\mathbb{R}^{c'\times d})$.
Then, by Lemma \ref{lemma:activation_convolution}, $\oCNN{\sigma*\varphi}{d,2k+1}{N}{c,c'}
    \subset \oCNN{\sigma}{d,2k+1}{N}{c,c'}$, further leading to $\oCNN{\sigma}{d,2k+1}{N}{c,c'} = C(k,\mathbb{R}^{c'\times d})$.
    This completes the proof.
\end{proof}

\subsection{Proof of Lemma \ref{lemma: d23}}\label{appendix:proof_d23}
\begin{proof}
   \paragraph{Case 1 $d=2$:} 
For a monomial $M$ containing at least one $x^{i}_1$, $(0,M)$ belongs to $\osCNN{\sigma}{}{1}{c,1}$. 
In other words, for $i_0\in [1,c]$, $\alpha_{i,j}\in \mathbb{N}_0$, and $M = x^{i_0}_1 \prod_{i=1}^c \prod_{j=1,2} (x^{i}_{j})^{\alpha_{i,j}}$, $(0, M)\in \osCNN{\sigma}{}{1}{c,1}$. 
This relation originates from the equation:
\begin{equation}
p^{i_0}_{-1}\bigodot_{i=1}^c\bigodot_{j=1,2} (p^{i}_{j-2})^{\alpha_{i,j}} = (0, M) \in \osCNN{\sigma}{}{1}{c,1}.
\end{equation}
Then, for $C(x) = U_{-1}x$, $C((0,M))= (M,0)\in \oCNN{}{}{2}{c,1}$, and for $C(x) = U_{0}x$, $C((0,M)) = (0,M)\in \oCNN{}{}{2}{c,1}$.
Through a symmetric process, when a monomial $M$ contains at least one $x^{i}_2$, $(M,0), (0,M)\in \oCNN{}{}{2}{c,1}$.
Therefore, all nonconstant monomials are in $\oCNN{}{}{2}{c,1}$
By Corollary \ref{lemma:onehot}, $e_1$ and $e_2$ are in $\oCNN{}{}{2}{c,1}$, leading to the inclusion of all monomials.
This completes the proof for the case when $d=2$.

\paragraph{Case 2 $d=3$:} 
Consider an arbitrary monomial $M=\prod_{i=1}^{c}\prod_{j=1}^{3}(x^i_j)^{\alpha_{i,j}}$ where $\alpha_{i,j}\in \mathbb{N}_0$.
By leveraging Corollary \ref{lemma:onehot}, we establish $e_i\in \oCNN{}{}{2}{c,1}$, for $ i\in [1, 1]\cup [3, 3] = \{1, 3\}$.
Moreover, owing to $e_2 = \boldsymbol{1}_3 - e_1 - e_3$, $e_2$ is in $\oCNN{}{}{2}{c,1}$.
Because $p^i_j \in \oCNN{}{}{2}{c,1}$ for $j\in [-2,2]$, by Lemma \ref{lemma:product}, we have
\begin{equation}
    e_2 \odot \left(\bigodot_{i}^c\bigodot_{j=-1,0,1} (p^{i}_{j})^{\alpha_{i,j+2}}\right) = (0,M,0)\in \osCNN{\sigma}{}{2}{c,1}.
\end{equation}
Employing the convolutional layers $U_{-1}x, U_0x, U_{1}x\in \mathcal{L}^3_{1,1}$ as the final layer, $(M,0,0), (0,M,0), (0,0,M)\in \oCNN{}{}{3}{c,1}$. 
This completes the proof.
\end{proof}

\section{Proofs in Section \ref{sec:cnn_width}}\label{appendix:proof_cnn_width}
\subsection{Proof of Lemma \ref{lemma:activation_remove}}\label{appendix:proof_activation_remove}
\begin{proof}
Let $C_1\in \mathcal{L}^{2k+1}_{c_1, c_2}$ has kernels $w^1_{j,i}$ and biases $\delta^1_j$ for $i\in [1,c_1]$ and $j\in [1,c_2]$, and let $C_2\in \mathcal{L}^{2k+1}_{c_2, c_3}$ has kernels $w^2_{k,j}$ and biases $\delta^2_k$ for $j\in [1,c_2]$ and $k\in [1,c_3]$. 
We define $C'_1\in \mathcal{L}^{2k+1}_{c_1, c_2}$ with kernels $w'^1_{j,i}$ and biases $\delta'^1_j$ and $C'_2 \in \mathcal{L}^{2k+1}_{c_2, c_3}$ with kernels $w'^2_{k,j}$ and biases $\delta'^2_k$ as follows:
\begin{equation}
    w'^1_{j,i} := \begin{cases}
     w^1_{j,i}          \text{ \quad if } j\in I
        \\ \frac{w^1_{j,i}}{N}  \text{ \quad otherwise }
    \end{cases}, 
       \delta'^1_j := \begin{cases}
     \delta^1_j          \text{ \qquad\quad if } j\in I
         \\ \alpha + \frac{\delta^1_j}{N} \text{ \quad otherwise }
    \end{cases},
\end{equation}
and 
\begin{equation}
    w'^2_{k,j} := \begin{cases}
     w^2_{k,j}          \text{ \quad\quad\quad if } j\in I
         \\ \frac{N}{\sigma'(\alpha)} w^2_{k,j} \text{\quad  otherwise }
    \end{cases},  \quad \delta'^2_k := 
         -\frac{N\sigma(\alpha)}{\sigma'(\alpha)} + {\delta^2_k}.
\end{equation}
   Then, $f^{k}$, the $k$-th channel of $f:=C'_2\circ\sigma  \circ C'_1$, becomes
       \begin{align}
         f^k(x) &:= \sum_{j=1}^{c_2}w'^2_{k,j}\circledast \sigma\left(\sum_{i=1}^{c_1} w'^1_{j,i}\circledast x^i + \delta'^1_j \boldsymbol{1}_d\right) +\delta'^2_k \boldsymbol{1}_d
         \\& = \sum_{j\in I} w'^2_{k,j}\circledast \sigma\left(\sum_{i=1}^{c_1} w'^1_{j,i}\circledast x^i + \delta'^1_j \boldsymbol{1}_d\right) 
         \\ &\quad\quad\quad\quad +  \sum_{j\notin I} w'^2_{k,j}\circledast \sigma\left(\sum_{i=1}^{c_1} w'^1_{j,i}\circledast  x^i + \delta'^1_j \boldsymbol{1}_d\right) +\delta'^2_k \boldsymbol{1}_d
         \\& =  \sum_{j\in I} w^2_{k,j}\circledast \sigma\left(\sum_{i=1}^{c_1} w^1_{j,i}\circledast  x^i + \delta^1_j \boldsymbol{1}_d\right)
         \\&\quad\quad\quad\quad+ \sum_{j\notin I} \frac{N}{\sigma'(\alpha)} w^2_{k,j} \circledast\sigma\left(\sum_{i=1}^{c_1} \frac{w^1_{j,i}}{N}\circledast  x^i + \frac{\delta^1_j}{N} + \alpha \right) - \frac{N\sigma(\alpha)}{\sigma'(\alpha)} + {\delta^2_k \boldsymbol{1}_d}
         .
    \end{align}
The $k$-th channel of $g:=C_2 \circ \widetilde{\sigma}_I \circ C_1$, denoted as $g^k$, is
\begin{align}
    &g^k(x)= \sum_{j=1}^{c_2}w^2_{k,j}\circledast \widetilde{\sigma}_I\left(\sum_{i=1}^{c_1} w^1_{j,i}\circledast  x^i + \delta^1_j \boldsymbol{1}_d\right) +\delta^2_k \boldsymbol{1}_d
    \\ &= \left(\sum_{j\in I}+  \sum_{j\notin I}\right) \left(w^2_{k,j}\circledast \widetilde{\sigma}_I\left(\sum_{i=1}^{c_1} w^1_{j,i}\circledast  x^i + \delta^1_j \boldsymbol{1}_d\right) \right)+\delta^2_k \boldsymbol{1}_d
    \\ &= \sum_{j\in I} w^2_{k,j}\circledast {\sigma}\left(\sum_{i=1}^{c_1} w^1_{j,i}\circledast  x^i + \delta^1_j \boldsymbol{1}_d\right) +  \sum_{j\notin I} w^2_{k,j}\circledast \left(\sum_{i=1}^{c_1} w^1_{j,i}\circledast  x^i + \delta^1_j \boldsymbol{1}_d\right) +\delta^2_k \boldsymbol{1}_d.
\end{align}
Then, the difference between the two functions, $g^k - f^k$, becomes
\begin{align}
    &g^k(x) - f^k(x) 
    \\&=  \sum_{j\in I} w^2_{k,j} \circledast{\sigma}\left(\sum_{i=1}^{c_1} w^1_{j,i}\circledast  x^i + \delta^1_j \boldsymbol{1}_d\right) +  \sum_{j\notin I} w^2_{k,j} \circledast\left(\sum_{i=1}^{c_1} w^1_{j,i}\circledast  x^i + \delta^1_j \boldsymbol{1}_d \right) +\delta^2_k \boldsymbol{1}_d
    \\ &- \sum_{j\in I} \left(w^2_{k,j} \circledast\sigma\left(\sum_{i=1}^{c_1} w^1_{j,i}\circledast  x^i + \delta^1_j \boldsymbol{1}_d\right) \right)
   \\ &\quad\quad\quad  -\sum_{j\notin I} \left(\frac{N}{\sigma'(\alpha)} w^2_{k,j} \circledast\sigma\left(\sum_{i=1}^{c_1} \frac{w^1_{j,i}}{N}\circledast  x^i + \frac{\delta^1_j \boldsymbol{1}_d}{N} + \alpha \right)\right) + \frac{N\sigma(\alpha)}{\sigma'(\alpha)} - {\delta^2_k} \boldsymbol{1}_d
\\ &= \sum_{j\notin I} w^2_{k,j} \circledast\left(\sum_{i=1}^{c_1} w^1_{j,i}\circledast  (x^i) + \delta^1_j \boldsymbol{1}_d\right) +\delta^2_k \boldsymbol{1}_d
 \\ &\quad\quad\quad - \sum_{j\notin I} \left(\frac{N}{\sigma'(\alpha)} w^2_{k,j}\circledast \sigma\left(\sum_{i=1}^{c_1} \frac{w^1_{j,i}}{N}\circledast  (x^i) + \frac{\delta^1_j \boldsymbol{1}_d}{N} + \alpha \right)\right)  + \frac{N\sigma(\alpha)}{\sigma'(\alpha)} - {\delta^2_k \boldsymbol{1}_d}.
\end{align}
Let $u^j$ be defined as $u^j:=\sum_{i=1}^{c_1} w^1_{j,i}\circledast  x^i + \delta^1_j \boldsymbol{1}_d$.
Then,
\begin{align}
    g^k(x) - f^k(x)&= \sum_{j\notin I}w^2_{k,j} \circledast\left(u^j - \frac{N}{\sigma'(\alpha)} \sigma\left(\frac{u^j}{N} + \alpha \right)  + \frac{N\sigma(\alpha)}{\sigma'(\alpha)}\right)
\\ &= \sum_{j\notin I}w^2_{k,j} \circledast\frac{u^j}{\sigma'(\alpha) }\left( \sigma'(\alpha)- \frac{\sigma(\alpha + \frac{u^j}{N}) - \sigma(\alpha)}{\frac{u^j}{N}} \right)
    \\ &= \sum_{j\notin I}w^2_{k,j} \circledast\frac{u^j}{\sigma'(\alpha) }\left( \sigma'(\alpha)- \sigma'(\alpha + t_j) \right)\xrightarrow[]{\text{N}\rightarrow\infty} 0,
\end{align}
where $|t_j|<\left|\frac{u^j}{N}\right|$.
The uniform convergence occurs due to $x$ residing in the compact domain $K$, ensuring uniform boundedness of $u^j$ for all $x$. 
Consequently, $\frac{u^j}{N}$ converges uniformly to $0$ across all $x$ within this compact set.
\end{proof}

\subsection{Proof of Lemma \ref{lem:remove_activation_multi}}\label{appendix:proof_remove_activation_multi}
\begin{proof}
Utilize mathematical induction on $N$.
Lemma \ref{lemma:activation_remove} confirms the induction hypothesis is satisfied for $N=2$.
Assume the hypothesis holds for $N = N_0$.
For $N= N_0 + 1$, consider the function $f_{N_0 + 1}$ defined as:
 \begin{equation}
     f_{N_0 + 1} = C_{N_0 + 1} \circ \widetilde{\sigma}_{I_{N_0}}\circ C_{N_0}\circ \dots \circ \widetilde{\sigma}_{I_1}\circ C_1.
 \end{equation}

 Then, for $f_{N_0}: = C_{N_0}\circ \widetilde{\sigma}_{I_{N_0-1}}\circ \dots \circ C_1$:
 \begin{equation}
     f_{N_0 + 1} = C_{N_0 + 1} \circ\widetilde{\sigma}_{I_{N_0}} \circ f_{N_0}.
 \end{equation}
Define the compact set $K'\subset \mathbb{R}^{c_{N_0}\times d}$ as:
\begin{equation}
    K':= f_{N_0}(K) + B_{1}(0) = \left\{ f_{N_0}(x) + \alpha | x\in K, \|\alpha\|_{\infty} \leq 1 \right\}.
\end{equation}
Because a continuous function is uniformly continuous on a compact domain, there exists $0<\delta<1$ such that if $\|x-y\|<\delta$ for $x,y\in K'$, then
\begin{equation}
     \left\|C_{N_0+1}\circ \widetilde{\sigma}_{I_{N_0}}(x) -  C_{N_0+1}\circ \widetilde{\sigma}_{I_{N_0}}(y)\right\|_{\infty}<\frac{\epsilon}{2}.
\end{equation}
 By the induction hypothesis, there exists $g \in \CNN{\sigma}{d, 2k+1}{N_0}{(c_0, c_1,\dots, c_N)}$, such that
\begin{equation}
    \|f_{N_0}- g\|_{\infty,K}<\delta.
\end{equation}
Then, because $f_{N_0}(x)$ and $g(x)$ are in $K'$ for $x\in K$,
\begin{equation}
     \|f_{N_0+1} - C_{N_0 + 1} \circ\widetilde{\sigma}_{I_{N_0}} \circ g\|_{\infty,K} = \|C_{N_0 + 1} \circ\widetilde{\sigma}_{I_{N_0}} \circ f_{N_0} - C_{N_0 + 1} \circ\widetilde{\sigma}_{I_{N_0}} \circ g\|_{\infty,K}<\frac{\epsilon}{2}.
\end{equation}
Express $g$ as:
\begin{equation}
    g = C'_{N_0}\circ \sigma \circ \dots \circ\sigma \circ C'_1,
\end{equation}
where $C'_i\in \mathcal{L}^{2k+1}_{c_{i-1}, c_{i}}$.

By Lemma \ref{lemma:activation_remove}, there exist convolutional layers $C''_{N_0 +1}\in \mathcal{L}^{2k+1}_{c_{N_0}, c_{N_0+1}}$ and $C''_{N_0}\in \mathcal{L}^{2k+1}_{c_{N_0-1}, c_{N_0}}$ such that:
\begin{equation}
    \|C_{N_0+1}\circ \widetilde{\sigma}_{I_{N_0}} \circ C'_{N_0} - C''_{N_0+1} \circ \sigma \circ C''_{N_0}\|_{\infty,K''}<\frac{\epsilon}{2}.
\end{equation}
where the compact set $K''$ is defined as $K'':=\sigma\circ C'_{N_0-1}\circ\dots\circ\sigma\circ C'_1 (K)$.
Define $h\in \CNN{\sigma}{d, 2k+1}{N_0}{(c_0,\dots, c_N)}$ as:
\begin{equation}
    h:= C''_{N_0+1} \circ \sigma \circ C''_{N_0} \circ \sigma\circ C'_{N_0-1}\circ\dots \circ C'_1.
\end{equation}
The subsequent equation holds:
\begin{multline}
    \|C_{N_0 + 1} \circ\widetilde{\sigma}_{I_{N_0}} \circ g - h \|_{\infty,K} 
    \\ = \|C_{N_0 + 1} \circ\widetilde{\sigma}_{I_{N_0}} \circ C'_{N_0}\circ \sigma \circ \dots \circ\sigma \circ C'_1  
      - C''_{N_0+1} \circ \sigma \circ C''_{N_0} \circ \sigma\circ C'_{N_0-1}\circ\dots \circ C'_1 \|_{\infty,K}
    \\ <  \|C_{N_0 + 1} \circ\widetilde{\sigma}_{I_{N_0}} \circ C'_{N_0} -C''_{N_0+1} \circ \sigma \circ C''_{N_0} \|_{\infty,K''}<\frac{\epsilon}{2}.
\end{multline}
To sum up,
\begin{equation}
    \|f_{N_0+1}-h\|_{\infty,K}<\|f_{N_0+1} - C_{N_0 + 1} \circ\widetilde{\sigma}_{I_{N_0}} \circ g \|_{\infty,K} + \|C_{N_0 + 1} \circ\widetilde{\sigma}_{I_{N_0}} \circ g - h \|_{\infty,K}<\epsilon. 
\end{equation}
Therefore, the induction hypothesis holds for $N = N_0+1$.
This completes the proof.
\end{proof}

\subsection{Proof of Lemma \ref{lemma:polynomial}} \label{appendix:proof_polynomial}
\begin{proof}
It is straightforward to verify that $\overline{ \Delta^{p,d,2k+1}_{c,c', w}}$ remains unchanged under affine transformations applied to the activation function. 
   Specifically, consider $q$ defined as
    \begin{equation}
        q(z) = ap(cz+d)+d,
    \end{equation}
    where $a\neq 0$, $c\neq 0$, and $a,b,c,d\in\mathbb{R}$,
     This leads to:
    \begin{equation}
       \overline{\Delta^{p, d,2k+1}_{c,c',w}}= \overline{\Delta^{q, d,2k+1}_{c,c',w}}
    \end{equation}
    Therefore, without loss of generality, assume that 
    \begin{equation}
        p(z) = z + \sum_{i=2}^m a_iz^i.
    \end{equation}
    Define $p_n(z)$ as:
    \begin{equation}
        p_n(z) := s^n p\left(\frac{z}{s^n}\right) = z +  \sum_{i=2}^m \frac{a_i}{s^{n(i-1)}}z^i.
    \end{equation}
   Additionally, define the composition of $p_n$ as:
    \begin{equation}
        r_N:= p_1\circ p_2 \dots \circ p_N.
    \end{equation}
Next, we'll demonstrate that for $s>1$, $r_N$ uniformly converges to a non-polynomial entire function. This is established by applying Theorem 1.4 from \cite{kojima2012convergence}.
    \begin{theorem}[\cite{kojima2012convergence}]
    Let $f_n(z)=a_{n, 0}+a_{n, 1} z+a_{n, 2} z^2+a_{n, 3} z^3+\cdots(n=1,2,3 \ldots)$ be an entire function and define
\begin{equation}
    A_n:=\sup \left\{\left|a_{n, r}\right|^{1 /(r-1)} \mid \quad r=2,3,4, \ldots\right\} .
\end{equation}
Suppose that
\begin{equation}
    \sum_{n=1}^{\infty} A_n, \quad \sum_{n=1}^{\infty}\left|a_{n, 0}\right|, \quad \text { and } \quad \prod_{n=1}^{\infty} a_{n, 1}
\end{equation}
are convergent. 
Define
\begin{equation}
    F_N(z):= f_1\circ f_{2}\circ\dots \circ f_N(z).
\end{equation}
Then, the sequence of entire functions $\left\{F_N(z)\right\}_{N=1}^{\infty}$ uniformly converges on any compact subset of $\mathbb{C}$. 
Specifically, $\operatorname{lim}_{n\rightarrow \infty} F_n(z)$ is entire.
    \end{theorem}
    To verify the assumption of the theorem, define $\rho_n$ as
    \begin{equation}
        \rho_n := \operatorname{max}_{i=2,3,\dots, m} \left|\frac{a_i}{s^{n(i-1)}}\right|^{\frac{1}{i-1}}.
    \end{equation}
    Then, for $A:= \operatorname{max}_{i=2,3,\dots, m}|a_i|^{\frac{1}{i-1}}$, $\rho_n = \frac{A}{s^n}$.
    Moreover, we have
\begin{equation}
    \sum_{n=1}^{\infty} \rho_n = \sum_{n=1}^{\infty} \frac{A}{s^n} <\infty.
\end{equation}
This implies that $ r_N$ uniformly converges to a holomorphic function in an arbitrary compact domain.
Denote this holomorphic function as $\sigma$.
Because holomorphic functions $r_N$ converge uniformly to $\sigma$, their derivatives $r'_N$ also converge to $\sigma'$.
Therefore, $\sigma'(0) = \operatorname{limit}_{N\rightarrow\infty}r'_N(0) = 1$.
Additionally, $\sigma$ adheres to the following equation: 
\begin{equation}
    \sigma(sz) = sp\left(\sigma(z)\right).
\end{equation}
If $\sigma$ were a non-constant polynomial, the degrees of the left-hand side and the right-hand side would differ.
Therefore, $\sigma$ must be a constant if it is a polynomial.
However $\sigma'(0) =1$, which makes contradiction.
Thus, $\sigma$ is non-polynomial, and we proved that the composition of $p$ and affine transformation converges to the non-polynomial $C^{\infty}$ function $\sigma$.
We will construct a CNN that models this process.

To prove $\oCNN{\sigma}{d,2k+1}{N}{(c,w,\dots, w,c')}\subset \overline{\Delta^{p, d,2k+1}_{c,c',w}}$ for general $N$, we employ mathematical induction on $N$.
The base case $N=1$ is trivially satisfied.
Assume that the induction hypothesis holds for $N=N_0$: $\oCNN{\sigma}{d,2k+1}{N_0}{(c,w,\dots, w,c')}\subset \overline{\Delta^{p, d,2k+1}_{c,c',w}}$, and consider a function $f\in \oCNN{\sigma}{d,2k+1}{N_0+1}{(c,w,\dots, w,c')}$.
Then, $f$ can be represented as
\begin{equation}
    f = C_{N_0+1}\circ \sigma \circ \dots \circ \sigma\circ C_{0},  
\end{equation}
where $C_{0}\in \mathcal{L}^{2k+1}_{c,w}$, $C_{N_0+1}\in \mathcal{L}^{2k+1}_{w, c'}$, and $C_i\in \mathcal{L}^{2k+1}_{w, w}$ for $i\in [1,N_0]$.
Define $g\in \oCNN{\sigma}{d,2k+1}{N_0}{(c,w,\dots, w,c')}$ as: 
\begin{equation}
   g := C_{N_0}\circ \sigma \circ \dots \circ \sigma\circ C_{0}.
\end{equation}
Then,
\begin{equation}
    f = C_{N_0+1}\circ \sigma \circ g.
\end{equation}
For a positive number $\epsilon\in \mathbb{R}_{+}$, there exists $0<\delta<1$ such that if $\|x-y\|_{\infty}<\delta$, then $\|C_{N_0+1}(x)-C_{N_0+1}(y)\|_{\infty}<\frac{\epsilon}{2}$.
Additionally, there exists $M\in \mathbb{N}$ such that $\|\sigma-r_M\|_{\infty, g(K)}<\delta$.
Then, define $h$ as 
\begin{equation}
    h = C_{N_0+1}\circ\left( s * Id\right)\circ p \circ \left(s * Id\right)\circ p\circ  \cdots \circ \left(s * Id\right)\circ p\circ \left(\frac{Id}{s^M}\right)\circ g.
\end{equation}
Then,
\begin{equation}
    \|f-h\|_{\infty,K} 
    =\| C_{N_0+1}\circ \left(\sigma - r_M\right)\circ g\|_{\infty,K}< \frac{\epsilon}{2}.
\end{equation}
In addition, there exists $0<\delta'<1$ such that if $x,y\in g(K) + B_1(0)$ and $\|x-y\|_{\infty}<\delta'$, $\|C_{N_0+1}\circ r_M(x) - C_{N_0+1}\circ r_M(y)\|_{\infty}<\frac{\epsilon}{2}$.

By the induction hypothesis, there exists $h_2\in {\Delta^{p, d,2k+1}_{c,w,w}}$ such that 
\begin{equation}
    \|g-h_2\|_{\infty,K}<\delta'.
\end{equation}
We get
\begin{equation}
    \|h- C_{N_0+1}\circ r_M \circ h_2\|_{\infty,K} = \|C_{N_0+1}\circ r_M(g)- C_{N_0+1}\circ r_M (h_2)\|_{\infty,K}<\frac{\epsilon}{2}.
\end{equation}
Furthermore,
\begin{equation}
   \|f-C_{N_0+1}\circ r_M \circ h_2\|_{\infty,K} =\|f-h\|_{\infty,K}  + \|h- C_{N_0+1}\circ r_M \circ h_2\|_{\infty,K} <\epsilon.
\end{equation}
Because $h_2\in {\Delta^{p, d,2k+1}_{c,w,w}}$, and $C_{N_0+1}\circ r_M \circ h_2= C_{N_0+1}\circ \left( s * Id\right)\circ p \circ \left(s * Id\right)\circ p\circ  \cdots \circ \left(s * Id\right)\circ p\circ \left(\frac{Id}{s^M}\right)\circ h_2$, $C_{N_0+1}\circ r_M \circ h_2\in \overline{\Delta^{p, d,2k+1}_{c,w,w}}$, and this completes the proof.

\end{proof}

\section{Proofs in Section \ref{sec:multidimensional}}
\subsection{Proof of Proposition \ref{lemma:multidimesional_d3_not_universal}} \label{appendix:proof_multidimesional_d3_not_universal}
\begin{proof}
    Without loss of generality, assume that $d_1 = 3$.
Consider an arbitrary two-layered CNN $f\in \CNN{\sigma}{\boldsymbol{d}, 3\boldsymbol{1}_D}{2}{(c,n,1)}$ with $n$ intermediate channels.
Then, there exist convolutional layers $C_1\in \mathcal{L}^{\boldsymbol{d},3\boldsymbol{1}_D}_{c, n}$ and $C_2\in \mathcal{L}^{\boldsymbol{d},3\boldsymbol{1}_D}_{n, 1}$ such that
\begin{equation}
    f = C_2\circ\sigma\circ C_1.
\end{equation}
Let $C_1$ has kernels $(a^{i_1,i_2})_{1\leq i_1\leq n, 1\leq i_2\leq c}\in M_{n,c}(\mathbb{R}^{3\boldsymbol{1}_K})$ and biases $\delta_{i_1}\in \mathbb{R}$, while $C_2$ has kernels $b^{i} \in\mathbb{R}^{3\boldsymbol{1}_D}$ and bias $\delta_0$.

    Now consider two input values, $z,w\in \mathbb{R}^{c\times 3\boldsymbol{1}_D}$, defined as
    \begin{equation}
        z^i_{\boldsymbol{j}} = \begin{cases}
            x^i_1 &\text{ if } \boldsymbol{j} = (1, 1, \dots, 1)
            \\x^i_2 &\text{ if } \boldsymbol{j} = (2, 1, \dots, 1)
            \\ 0 &\text{ otherwise }
        \end{cases}, 
        \text{ and }         
        w^i_{\boldsymbol{j}} = \begin{cases}
            x^i_1 &\text{ if } \boldsymbol{j} = (2, 1, \dots, 1)
            \\x^i_2 &\text{ if } \boldsymbol{j} = (3, 1, \dots, 1)
            \\ 0 &\text{ otherwise }
        \end{cases}, 
    \end{equation}
    for $i\in [1,c]$ and $x^i_1, x^i_2\in \mathbb{R}$.
    Then, we compute $f_{(1,1,\dots, 1)}(z) - f_{(2,1,\dots, 1)}(w)$. 
    \begin{equation}
        f(z)_{(1,1,\dots, 1)} =     \sum_{i=1}^n \sum_{|\boldsymbol{j}|\leq 1}b^i_{-\boldsymbol{j}} \sigma\left( \left(C^i_1(z)\right)_{(1,1,\dots,1)+\boldsymbol{j}}\right) + \delta_0
    \end{equation}
    \begin{equation}
        f(w)_{(2,1,\dots, 1)} =     \sum_{i=1}^n \sum_{|\boldsymbol{j}|\leq 1}b^i_{-\boldsymbol{j}} \sigma\left( \left(C^i_1(w)\right)_{(2,1,\dots,1)+\boldsymbol{j}}\right) + \delta_0
    \end{equation}
    Additionally, for $\boldsymbol{j} =(j_1, j_2,\dots, j_D)$ with $j_1\geq 0$, we have
    \begin{equation}
        \left(C^i_1(z)\right)_{(1,1,\dots, 1)+\boldsymbol{j}} = \left(C^i_1(w)\right)_{(2,1,\dots, 1)+\boldsymbol{j}}=\sum_{i'=1}^c a^{i,i'}_{\boldsymbol{j}}x^{i'}_1 + a^{i,i'}_{\boldsymbol{j}-(1,0,\dots,0)}x^{i'}_2.
    \end{equation}
    Therefore, 
    \begin{equation}
        f(z)_{(1,1,\dots, 1)} - f(w)_{(2,1,\dots, 1)} = \sum_{i=1}^n\sum_{\{\boldsymbol{j}|j_1 = -1\}} - b^i_{-\boldsymbol{j}}
        \sigma\left(\left(C^i_1(w)\right)_{(2,1,\dots,1)+\boldsymbol{j}}\right).
    \end{equation}
  Because,  for $\boldsymbol{j}$ with $j_1=-1$, $\left(C^i_1(w)\right)_{(2,1,\dots,1)+\boldsymbol{j}}= \sum_{i'=1}^c a^{i,i'}_{\boldsymbol{j}}x^{i'}_{1}$ is the function of input components $x^{i'}_1$, we can observe that $f_{(1,1,\dots, 1)}(z) - f_{(2,1,\dots, 1)}(w)$ becomes the function of input components $x^{i'}_1$, which results in the same contradiction as in the proof of Proposition \ref{prop:d3_not_universal}, which completes the proof.
\end{proof}

\subsection{Proof of Corollary \ref{corollary:multidimensional_subset}}\label{appendix:proof_multidimensional_subset}
\begin{proof}
The proof for the first argument aligns with Corollary \ref{Corollary:subset}. 
  
  Now, for the second argument, we employ mathematical induction on $l$.
  For $l=1$, the mapping $x\mapsto U^{\boldsymbol{d}}_{\boldsymbol{j}}x^i\in \mathcal{L}^{3\boldsymbol{1}_D}_{c',1}$ for $i\in [1,c']$.
  Then, by Lemma \ref{spacelemma}, $U^{\boldsymbol{d}}_{\boldsymbol{j}} \circ g^i\in \oCNN{}{}{l_0+1}{c,1}$ for $g\in \oCNN{}{}{l_0}{c,1}$, $|\boldsymbol{j}|\leq 1$, and $i\in [1,c']$.
  Thus, the induction hypothesis is satisfied for $l=1$.
  Assume that the induction hypothesis holds for $l=l'$: $U^{\boldsymbol{d}}_{\boldsymbol{j}} g^i\in \oCNN{}{}{l_0+l'}{c,1}$ for $|\boldsymbol{j}|\leq l'$. 
  
   Now consider an index $|\boldsymbol{j}|\leq l'+1$. 
   Define $\boldsymbol{j}'$ as 
   \begin{equation}
       j'_k = \begin{cases}
           l' &\text{ if } j_k = l'+1\\ 
            -l' &\text{ if } j_k = -l'-1\\
            j_k  &\text{ otherwise}
            \end{cases}.
   \end{equation}
   Then, $|\boldsymbol{j} - \boldsymbol{j}'|\leq 1$, $|\boldsymbol{j}'|\leq l'$, and $\boldsymbol{j}'$ and $\boldsymbol{j} - \boldsymbol{j}'$ have the same signs throughout.
   Therefore, $U^{\boldsymbol{d}}_{\boldsymbol{j}} = \left(U^{\boldsymbol{d}}_{\boldsymbol{j} - \boldsymbol{j}'}\right)U^{\boldsymbol{d}}_{\boldsymbol{j}'} $.
   Because by the induction hypothesis indicating that $U^{\boldsymbol{d}}_{\boldsymbol{j}'}g^i\in \oCNN{}{}{l_0+l'}{c,1}$ and by Lemma \ref{spacelemma}, we can assert that $U^{\boldsymbol{d}}_{\boldsymbol{j}}g^i = U^{\boldsymbol{d}}_{\boldsymbol{j}- \boldsymbol{j}'}\circ \left(U^{\boldsymbol{d}}_{\boldsymbol{j}'}g^i\right)\in \oCNN{}{}{l_0+l'+1}{c,1}$.
   Thus, the induction hypothesis holds for $l=l'+1$, and this completes the proof.
\end{proof}

\subsection{Proof of Lemma \ref{lemma:multidimensional_onehot}}\label{appendix:proof_multidimensional_onehot}
\begin{proof}
    Utilize mathematical induction on $n$.
    For $n=1$, we know that $\boldsymbol{1}_{\boldsymbol{d}}\in \oCNN{}{\boldsymbol{d}, 3\boldsymbol{1}_D}{1}{c,1}$, and $\boldsymbol{1}_{\boldsymbol{d}} =  \bigotimes_{i =1}^D \boldsymbol{1}_{d_i}$. Because $\boldsymbol{1}_{d_i} = \sum_{i=1}^{d_i}e^{d_i}_i\in V^2_{1,i}$, the induction hypothesis holds.
    
    For the case of $n=2$, consider $ \bigotimes_{i=1}^D v_i$, where $v_i\in V^3_{2,i}=\left\{ e^{d_i}_1,e^{d_i}_{d_i},\sum_{j=2}^{d_i-1} e^{d_i}_{j}\right\}$. 
    Then, $[1,D]$ is partitioned into three subsets: $I_1$, $I_2$, and $I_3$.
    Here $v_i=e^{d_i}_1$ if $i\in I_1$, $v_i =e^{d_i}_{d_i}$ if $i\in I_2$, and $v_i = \sum_{j=2}^{d_i-1}e^{d_i}_j$ if $i\in I_3$.
    Without loss of generality, assume that $I_1 = [1, m_1]$, $I_2 = [m_1+1, m_1+m_2]$, and $I_3 = [m_1 + m_2 + 1, m_1+m_2+m_3]$.
 Then, 
 \begin{align}
     &\left(\prod_{i\in I_1}  \left(Id - U^{\boldsymbol{d}}_{e^{D}_i}\right)
     \prod_{i\in I_2}  \left(Id - U^{\boldsymbol{d}}_{-e^{D}_i}\right)
     \prod_{i\in I_3}\left( U^{\boldsymbol{d}}_{e^{D}_i} + U^{\boldsymbol{d}}_{-e^{D}_i}- Id\right) \right)\boldsymbol{1}_{\boldsymbol{d}} 
     \\ &= 
          \left(\prod_{i=1}^{m_1}  \left(Id - U^{\boldsymbol{d}}_{e^{D}_i}\right)
     \prod_{i = m_1+1}^{m_1+m_2}  \left(Id - U^{\boldsymbol{d}}_{-e^{D}_i}\right)
     \prod_{i= m_1+m_2 +2}^{m_1+m_2+m_3}\left( U^{\boldsymbol{d}}_{e^{D}_i} + U^{\boldsymbol{d}}_{-e^{D}_i}- Id\right) \right) \bigotimes_{i=1}^D \boldsymbol{1}_{d_i} 
     \\ &=  
     \bigotimes_{i=1}^{m_1}  \left(\left(Id - U^{{d_i}}_{1}\right)\boldsymbol{1}_{d_i} \right) \otimes
    \bigotimes_{i = m_1+1}^{m_1+m_2}   \left(\left(Id - U^{{d_i}}_{-1}\right)\boldsymbol{1}_{d_i} \right)\otimes
     \bigotimes_{i= m_1+m_2 +2}^{m_1+m_2+m_3}\left(\left( U^{{d_i}}_{1} + U^{{d_i}}_{-1}- Id\right) \boldsymbol{1}_{d_i} \right)
     \\ &= \left(\bigotimes_{i\in I_1}  e^{d_i}_1\right)\otimes
     \left(\bigotimes_{i\in I_2} e^{d_i}_{d_i}\right)\otimes
     \left(\bigotimes_{i\in I_3} Id-e^{d_i}_1 - e^{d_i}_{d_i}\right)
      \\&= \bigotimes_{i=1}^D v_i.
 \end{align}
 Because $\left(\prod_{i\in I_1}  \left(Id - U^{\boldsymbol{d}}_{e_i}\right)
     \prod_{i\in I_2}  \left(Id - U^{\boldsymbol{d}}_{-e_i}\right)
     \prod_{i\in I_3}\left( U^{\boldsymbol{d}}_{e_i} + U^{\boldsymbol{d}}_{-e_i}- Id\right)\right) $ is the sum of the compositions of $Id$ and $U^{\boldsymbol{d}}_{e^D_i}$ for different $i$, it belongs to $\mathcal{L}^{3\boldsymbol{1}_{D}}_{1,1}$, and $\bigotimes_{i=1}^D v_i\in \oCNN{}{}{2}{c,1}$.
     Hence, the induction hypothesis holds for $n=2$.

We assume that the induction hypothesis holds for $n=n_0$, and consider $n = n_0 +1$.
For arbitrary $v_i\in V^3_{n_0+1, i}$ and $\bigotimes_{i=1}^D v_i $, consider partitions of $[1,D]$: $I_1$ and $I_2$.
Here, if $i\in I_1$, then $v_i\in V^1_{n_0+1,i}$, and if $i\in I_2$, then $v_i\in V^2_{n_0+1,i}$.
  Let the number of elements of $I_2$ be $m$.
  Use mathematical induction on $m$ to prove that if the number of elements of $I_2$ is $m$, $\bigotimes_{i=1}^D v_i\in \oCNN{}{}{n_0+1}{c,1}$.
  For $m=0$, all vectors $v_i$ are in $V^1_{n_0+1,i}$ and thus take the form $e^{d_i}_j$ for $j\in [1, n_0]\cup[d_i-n_0+1, d_i]$.
  Let such indices $j$ be denoted as $s_i$: $v_i = e_{s_i}$.
    Let the sets of indices $i$ where $s_i = n_0$ be $J_1$ and $s_i=d_i-n_0+1$ be $J_2$.
    Define $\boldsymbol{j}$ as $\boldsymbol{j} := \sum_{i\in J_1} e_i - \sum_{i\in J_2} e_{i}$.
    Then, 
    \begin{equation}
        U^{\boldsymbol{d}}_{\boldsymbol{j}}\left(\bigotimes_{i=1}^D e_{s_i -j_i} \right)= \bigotimes_{i=1}^D e_{s_i }.
    \end{equation}
   Because $s_i - j_i \in [1, n_0-1]\cup [d_i-n_0+2, d_i]$ for all $i$, by the induction hypothesis, $\bigotimes_{i=1}^D e_{s_i -j_i}\in \oCNN{}{}{n_0}{c,1}$.
Therefore, by Corollary \ref{corollary:multidimensional_subset},
\begin{equation}
    \bigotimes_{i=1}^D v_{i } = \bigotimes_{i=1}^D e_{s_i } = U^{\boldsymbol{d}}_{\boldsymbol{j}}\left(\bigotimes_{i=1}^D e_{s_i -j_i} \right) \in \oCNN{}{}{n_0+1}{c,1}.
\end{equation}

Here, we assume that if the number of elements of $I_2$ is $m_0$, then, $\bigotimes_{i=1}^D v_i\in \oCNN{}{}{n_0+1}{c,1}$.
Additionally, consider the case that the number of elements of $I_2$ of $\bigotimes_{i=1}^D v_i$ is $m_0+1$.
Without loss of generality, assume that $I_2 = [1, m_0+1]$, and $I_1 = [m_0+2, D]$.
Then, 
\begin{equation}
    \bigotimes_{i=1}^D v_i = \left(\bigotimes_{i=1}^{m_0+1} v_i\right) \otimes\left(\bigotimes_{i=m_0+2}^{D} v_i\right) 
    =  \left(\bigotimes_{i=1}^{m_0+1}  \sum^{d_i-n_0}_{j=n_0+1} e_j\right) \otimes\left(\bigotimes_{i=m_0+2}^{D} v_i\right) .
\end{equation}
Consider $\bigotimes_{i=1}^{m_0+1} \sum^{d_i-n_0 +1 }_{j=n_0} e_j$.

\begin{equation}
\bigotimes_{i=1}^{m_0+1} \sum^{d_i-n_0 +1 }_{j=n_0} e_j =\bigotimes_{i=1}^{m_0+1}\left(\left( \sum^{d_i-n_0 }_{j=n_0 +1} e_j \right)+ e_{n_0} + e_{d_i-n_0}\right)= \sum_{\boldsymbol{k}\in \left\{1,2,3 \right\}^{m_0+1}} \bigotimes_{i=1}^{m_0+1} u_{i,k_i},
\end{equation}
where 
\begin{equation}
    u_{i,k_i} = \begin{cases}
        \sum_{j = n_0+1}^{d_i-n_0}e_j \text{ if } k_i=1
        \\ e_{n_0} \text{\quad\quad\quad\quad if } k_i=2
        \\ e_{d_i - n_0 +1}\text{ \quad\, if } k_i=3
    \end{cases}.
\end{equation}
Then,
\begin{equation}
    \bigotimes_{i=1}^{m_0+1} \sum^{d_i-n_0 +1 }_{j=n_0} e_j  = \sum_{\boldsymbol{k}\in \left\{1,2,3 \right\}^{m_0+1}} \bigotimes_{i=1}^{m_0+1} u_{i,k_i} = \bigotimes_{i=1}^{m_0+1}u_{i,1} + \sum_{\boldsymbol{k}\in \left\{1,2,3 \right\}^{m_0+1} - (1,1,\dots, 1)} \bigotimes_{i=1}^{m_0+1} u_{i,k_i}.
\end{equation}
Therefore, 
\begin{equation}
    \bigotimes_{i=1}^{m_0+1} \sum_{j = n_0+1}^{d_i-n_0}e_j =  \bigotimes_{i=1}^{m_0+1}u_{i,1} = \bigotimes_{i=1}^{m_0+1} \sum^{d_i-n_0 +1 }_{j=n_0} e_j  - \sum_{\boldsymbol{k}\in \left\{1,2,3 \right\}^{m_0+1} - (1,1,\dots, 1)} \bigotimes_{i=1}^{m_0+1} u_{i,k_i}.
\end{equation}
By multiplying $\left(\bigotimes_{i=m_0+2}^{D} v_i\right)$ to both sides, we get
\begin{multline}\label{eq:temp_204}
    \left(\bigotimes_{i=1}^{m_0+1} \sum_{j = n_0+1}^{d_i-n_0}e_j\right) \otimes \left(\bigotimes_{i=m_0+2}^{D} v_i\right) 
    =  \bigotimes_{i=1}^{m_0+1} \sum^{d_i-n_0 +1 }_{j=n_0} e_j  \otimes \left(\bigotimes_{i=m_0+2}^{D} v_i\right) 
 \\ - \sum_{\boldsymbol{k}\in \left\{1,2,3 \right\}^{m_0+1} - (1,1,\dots, 1)} \bigotimes_{i=1}^{m_0+1} u_{i,k_i} \otimes \left(\bigotimes_{i=m_0+2}^{D} v_i\right).
\end{multline}
As there are at most $m_0$ elements of $V^2_{n_0+1,i}$ present in the product $\bigotimes_{i=1}^{m_0+1} u_{i,k_i}$ for $\boldsymbol{k}\in \left\{1,2,3 \right\}^{m_0+1} - (1,1,\dots, 1)$, $\bigotimes_{i=1}^{m_0+1} u_{i,k_i} \otimes \left(\bigotimes_{i=m_0+2}^{D} v_i\right) \in \oCNN{}{}{n_0+1}{c,1}$ due to the induction hypothesis considering the number of elements of $I_2$.
    Furthermore, let $\boldsymbol{l}=(l_1, \dots, l_{D})$ be defined as 
    \begin{equation}
        l_i :=  \begin{cases}
            1 &\text{ if } i\geq m_0+2 \text{ and } v_i = e_{n_0}
            \\ -1 &\text{ if } i\geq m_0+2 \text{ and }  v_i = e_{d_i- n_0+1}
            \\0 &\text{ otherwise}
        \end{cases},
    \text{ and }
    w_i:= \begin{cases}
            e_{n_0 -1} &\text{ if } v_i = e_{n_0}
            \\ e_{d_i- n_0} &\text{ if } v_i = e_{d_i- n_0+1}
            \\v_i &\text{ otherwise}
        \end{cases}.
    \end{equation}
Then, 
    \begin{equation}
        \bigotimes_{i=1}^{m_0+1} \sum^{d_i-n_0 +1 }_{j=n_0} e_j  \otimes \left(\bigotimes_{i=m_0+2}^{D} v_i\right)
        = U^{\boldsymbol{d}}_{\boldsymbol{l}} \left(\bigotimes_{i=1}^{m_0+1} \sum^{d_i-n_0 +1 }_{j=n_0} e_j  \otimes \left(\bigotimes_{i=m_0+2}^{D} w_i\right)\right).
    \end{equation}
    Because $\bigotimes_{i=1}^{m_0+1} \sum^{d_i-n_0 +1 }_{j=n_0} e_j  \otimes \left(\bigotimes_{i=m_0+2}^{D} w_i\right)\in  \oCNN{}{}{n_0}{c,1}$, $ \bigotimes_{i=1}^{m_0+1} \sum^{d_i-n_0 +1 }_{j=n_0} e_j  \otimes \left(\bigotimes_{i=m_0+2}^{D} v_i\right)$ is in $\oCNN{}{}{n_0+1}{c,1}$ by Corollary \ref{corollary:multidimensional_subset}.
    As all terms on the right-hand side of Eq (\ref{eq:temp_204}) are in $\oCNN{}{}{n_0+1}{c,1}$, the right-hand side $\left(\bigotimes_{i=1}^{m_0+1} \sum_{j = n_0+1}^{d_i-n_0}e_j\right) \otimes \left(\bigotimes_{i=m_0+2}^{D} v_i\right) $ is in $\oCNN{}{}{n_0+1}{c,1}$.
    Thus, the induction hypothesis on the number of elements of $I_2$ holds, and all $\bigotimes_{i=1}^D v_i $ are in $\oCNN{}{}{n_0+1}{c,1}$.
    Again, the induction hypothesis on $n$ holds, and it completes the proof.
  \end{proof}
  
\subsection{Proof of Proposition \ref{lemma:multidimensional_main_theorem}}\label{appendix:multidimensional_main_theorem}
\begin{proof}
Similar to Proposition \ref{lemma:main_theorem}, we only need to prove the case where the activation function $\sigma$ is $C^{\infty}$, and extending this to the general activation function follows the proof of Proposition \ref{lemma:main_theorem}.
    For an input vector $x = (x^1, x^2, \dots , x^c)\in \mathbb{R}^{c\times \boldsymbol{d}}$, the translation of $x^i\in\mathbb{R}^{\boldsymbol{d}}$ is defined as follows:
\begin{equation}
    p^i_{\boldsymbol{j}} := U^{\boldsymbol{d}}_{-\boldsymbol{j}}(x^i).
\end{equation}
\paragraph{Case 1. $d=4$:} 

We will construct a tensor that possesses only one non-zero component, specifically a monomial.
Consider an arbitrary monomial $M$ defined as:
\begin{equation}
    M: = \prod_{i=1}^c \prod_{\boldsymbol{j}\in \prod_{k=1}^{D}[1, d_k]} \left(x^i_{\boldsymbol{j}}\right)^{\alpha_{i,\boldsymbol{j}}}.
\end{equation}
for $\alpha_{i,\boldsymbol{j}}\in \mathbb{N}_0$. 
Now, define the set of indices $K$ as follows:
\begin{equation}
    K:= \left\{k\in [1,D]\left| \exists i\in [1,c],\boldsymbol{j}\in \prod_{k=1}^{D}[1, d_k] \text{ such that } j_k = 4 \text{ and } \alpha_{i,\boldsymbol{j}}\neq 0 \right.\right\}.
\end{equation}
Define the index ${\boldsymbol{j}'}=(j'_{1}, \dots, {j}'_D)$ as:
\begin{equation}
    j'_k = \begin{cases}
           2 \text{ if } k\in K
        \\ 1 \text{ otherwise}
    \end{cases}.
\end{equation}
Then, by Corollary \ref{corollary:multidimensional_subset}, for arbitrary $\boldsymbol{j}\in \prod_{k=1}^{D}[1, d_k]$ and $\alpha_{i, \boldsymbol{j}}\neq 0$, we have 
\begin{equation}
 p^i_{\boldsymbol{j}-\boldsymbol{j}'}\in \oCNN{}{}{2}{c,1}.
\end{equation}
This result holds because when the $k$-th component $j_k$ of $\boldsymbol{j}$ equals four, and $\alpha_{i, \boldsymbol{j}}\neq 0$, it implies $k\in K$, and subsequently $j'_k = 2$.
Therefore $|\boldsymbol{j} - \boldsymbol{j}'|\leq 2$ for all $\boldsymbol{j}$ with nonzero $\alpha_{i, \boldsymbol{j}}$ for some $i$.


By Lemma \ref{lemma:multidimensional_onehot}, 
\begin{equation}
    v := \bigotimes_{k=1}^D v_k\in \oCNN{\sigma}{\boldsymbol{d},3\boldsymbol{1}_D}{2}{c,1}, \text{ where }    
    v_k := \begin{cases}
       e_2 + e_3 &\text{ if } k\in K 
       \\ e_1  &\text{ otherwise}
    \end{cases}.
\end{equation}

By Lemma \ref{lemma:product}, the componentwise product of elements in $\oCNN{}{}{2}{c,1}$ is in $\osCNN{\sigma}{}{2}{c,1}$. Therefore,
\begin{equation}
    w:=v \odot \left(\bigodot_{i=1}^c \bigodot_{\boldsymbol{j}\in \prod_{k=1}^{D}[1, d_k]}  \left(p^i_{\boldsymbol{j}-\boldsymbol{j}'}\right)^{\alpha_{i,\boldsymbol{j}}}\right) \in \osCNN{\sigma}{}{2}{c,1}.
\end{equation}
Then, we aim to demonstrate that $w$ equals $Me_{\boldsymbol{j}'}$.

First, evaluate the $\boldsymbol{j}'$-th component of $w$:
\begin{align}
       & w_{\boldsymbol{j}'} =\left(v \odot \left(\bigodot_{i=1}^c \bigodot_{\boldsymbol{j}\in \prod_{k=1}^{D}[1, d_k]}  \left(p^i_{\boldsymbol{j}-\boldsymbol{j}'}\right)^{\alpha_{i,\boldsymbol{j}}}\right)\right)_{\boldsymbol{j}'}
       = v_{\boldsymbol{j}'}\times \prod_{i=1}^c \prod_{\boldsymbol{j}\in \prod_{k=1}^{D}[1, d_k]}  \left(p^i_{\boldsymbol{j}-\boldsymbol{j}'}\right)_{\boldsymbol{j}'}^{\alpha_{i,\boldsymbol{j}}}
    \\ &= 1\times  \prod_{i=1}^c \prod_{\boldsymbol{j}\in \prod_{k=1}^{D}[1, d_k]} \left( U^{\boldsymbol{d}}_{\boldsymbol{j}'- \boldsymbol{j}}\right)_{\boldsymbol{j}'}^{\alpha_{i,\boldsymbol{j}}}=  \prod_{i=1}^c \prod_{\boldsymbol{j}\in \prod_{k=1}^{D}[1, d_k]} \left( x^i_{\boldsymbol{j}}\right)^{\alpha_{i,\boldsymbol{j}}} = M.
\end{align}
Here, $v_{\boldsymbol{j}'}$ becomes one because 
\begin{equation}
    v_{\boldsymbol{j}'} =  \left(\bigotimes_{k=1}^D v_k\right)_{\boldsymbol{j}'} =\prod_{k=1}^D{\left(v_k\right)_{j'_k}} = 1.
\end{equation}
$\left(v_k\right)_{j'_k}$ is one for all $k\in [1,D]$: if $k\in K$, then $v_k = e_2+e_3$ leading to $j'_k=2$, and if $k\notin K$, then, $v_k = e_1$, and $j'_k=1$.

Next, to prove that all components other than the $\boldsymbol{j}'$-th are zero.
Because $v$ is multiplied by the constant tensor $v$, we only consider indices $\boldsymbol{j}''$ for which $v_{j''}=1$. 
For such indeces, the $k$-th component of $j''$ becomes $j''_k=2,3$ for $k\in K$. 
If $\boldsymbol{j}''\neq \boldsymbol{j}'$, there exist some indices $k_0\in K$ such that $j''_{k_0} = 3$.
It sufficies to prove that $\left(p^i_{\boldsymbol{j}-\boldsymbol{j}'}\right)_{\boldsymbol{j}''}=0$ for some $i\in [1,c]$ and $\boldsymbol{j}\in \prod_{k=1}^{D}[1, d_k]$ with nonzero $\alpha_{i, \boldsymbol{j}}$.
Because ${k_0}\in K$, there exists some $i,\boldsymbol{j}$ with $j_{k_0} = 4$ and $\alpha_{i,\boldsymbol{j}}\neq 0$.
For such $\boldsymbol{j}$, $\left(p^i_{\boldsymbol{j}-\boldsymbol{j}'}\right)_{\boldsymbol{j}''} = \left(U^{\boldsymbol{d}}_{\boldsymbol{j}' - \boldsymbol{j}}(x^i)\right)_{\boldsymbol{j}''}$.
Because the $k_0$-th component of $\boldsymbol{j}''$ is three and that of $\boldsymbol{j}' - \boldsymbol{j}$ is $2-4 = -2$, the $k_0$-th component of $\boldsymbol{j}'' - \left( \boldsymbol{j}' - \boldsymbol{j} \right)$, $j''_k-j'_k+j_k= 5 >4$, which makes $\left(p^i_{\boldsymbol{j}-\boldsymbol{j}'}\right)_{\boldsymbol{j}''}$ zero.
Therefore, $w$ is zero except for the $\boldsymbol{j}'$-th component, and $Me^{\boldsymbol{d}}_{\boldsymbol{j}'}\in \osCNN{\sigma}{}{2}{c,1}$.

For an index $\boldsymbol{j}\in \prod_{i=1}^D [1,2]$, we aim to demonstrate that $Me_{\boldsymbol{j}'}\in \oCNN{\sigma}{\boldsymbol{d,3\boldsymbol{1}}}{3}{c,1}$. 
Define the convolutional layer $U^{\boldsymbol{d}}_{\boldsymbol{j}-\boldsymbol{j}'}\in \mathcal{L}^{\boldsymbol{d},3\boldsymbol{1}_D}_{1,1}$.
Then, we have:
\begin{equation}
    U^{\boldsymbol{d}}_{\boldsymbol{j}-\boldsymbol{j}'}\left( Me_{\boldsymbol{j}'}\right) = Me^{\boldsymbol{d}}_{\boldsymbol{j}}\in \oCNN{\sigma}{\boldsymbol{d},3\boldsymbol{1}}{3}{c,1},
\end{equation}
for $\boldsymbol{j}\in \prod_{i=1}^D [1,2]$.

Additionally, due to the spatial symmetry of CNNs, $Me^{\boldsymbol{d}}_{\boldsymbol{j}}\in \oCNN{\sigma}{d,3\boldsymbol{1}}{3}{c,1}$ holds for $\boldsymbol{j} \in \prod_{i=1}^D I_i$, where $I_i \in \{[1,2], [d_i-1, d_i]\}$.
Therefore, $Me^{\boldsymbol{d}}_{\boldsymbol{j}}\in \oCNN{\sigma}{d,3\boldsymbol{1}}{3}{c,1}$ for all $\boldsymbol{j}\in \prod_{i=1}^D [1,d_i]$.
This completes the proof for the case of $d=4$.

\paragraph{Case 2. $d\geq 5$:}
For an arbitrary monomial $M = \prod_{i=1}^c\prod_{\boldsymbol{j}\in \prod_{k=1}^D [1,d_k]} (x^i_{\boldsymbol{j}})^{\alpha_{i,\boldsymbol{j}}}$, we will show that the tensor $Me^{\boldsymbol{d}}_{\boldsymbol{t}}$ is in $\osCNN{\sigma}{}{d-2}{c,1}$ for $\boldsymbol{t}\in \prod_{k=1}^D \left([d_k - (d-2), d-1]\bigcap [1, d_k]\right)$.

For $d\geq 5$, $V^3_{d-2,i}$ in Lemma \ref{lemma:multidimensional_onehot} includes $\{e_k|k\in [1,d_i]\}$. 
Therefore, $e^{\boldsymbol{d}}_{\boldsymbol{j}}\in \oCNN{}{}{d-2}{c,1}$ for $\boldsymbol{j}\in \prod_{k=1}^D [1,d_k]$.
By Corollary \ref{corollary:multidimensional_subset}, $p^i_{\boldsymbol{j}}=U^{\boldsymbol{d}}_{\boldsymbol{-j}}x^i \in \oCNN{}{}{d-2}{c,1}$ for $|\boldsymbol{j}| \leq d-2 $.
When $\boldsymbol{t}\in \prod_{k=1}^D \left([d_k - (d-2), d-1]\bigcap [1, d_k]\right)$ and $\boldsymbol{j}\in \prod_{k=1}^D [1,d_k]$, then $\boldsymbol{j}-\boldsymbol{t}\in [-d+2, d-2]^D$.
Thus, by applying Lemma \ref{lemma:product}, we have:
\begin{equation}
     e_{\boldsymbol{t}} \odot \left(\prod_{i=1}^c \prod_{\boldsymbol{j}\in \prod_{k=1}^D d_k} (p^{i}_{\boldsymbol{j}-\boldsymbol{t}})^{\alpha_{i,j}}\right)\in \osCNN{\sigma}{}{d-2}{c,1}.
\end{equation}
 for $\boldsymbol{t}\in \prod_{k=1}^D \left([d_k - (d-2), d-1]\bigcap [1, d_k]\right)$.
 We obtain:
 \begin{align}
     &\left(e_{\boldsymbol{t}} \odot \left(\bigodot_{i=1}^c \bigodot_{\boldsymbol{j}\in \prod_{k=1}^D d_k} (p^{i}_{\boldsymbol{j}-\boldsymbol{t}})^{\alpha_{i,\boldsymbol{j}}}\right)\right)_{\boldsymbol{t}}
      = \prod_{i=1}^c \prod_{\boldsymbol{j}\in \prod_{k=1}^D d_k} (p^{i}_{\boldsymbol{j}-\boldsymbol{t}})_{\boldsymbol{t}}^{\alpha_{i,\boldsymbol{j}}}
   \\  & = \prod_{i=1}^c \prod_{\boldsymbol{j}\in \prod_{k=1}^D [1,d_k]} \left(U^{\boldsymbol{d}}_{\boldsymbol{t} -\boldsymbol{j}}(x^i)\right)_{\boldsymbol{t}}^{\alpha_{i,\boldsymbol{j}}} 
   =  \prod_{i=1}^c \prod_{\boldsymbol{j}\in \prod_{k=1}^D d_k} \left(x^i_{\boldsymbol{j}}\right)^{\alpha_{i,\boldsymbol{j}}} = M.
 \end{align}

Therefore, $Me_{\boldsymbol{t}}\in \osCNN{\sigma}{}{d-2}{c,1}$ for $\boldsymbol{t}\in \prod_{k=1}^D \left([d_k - (d-2), d-1]\bigcap [1, d_k]\right)$.
Finally, by using the appropriate $U^{\boldsymbol{d}}_{\boldsymbol{j}}\in \mathcal{L}^{\boldsymbol{d},3\boldsymbol{1}_D}_{1,1}$ for the last convolutional layer, we obtain 
\begin{equation}
U^{\boldsymbol{d}}_{\boldsymbol{j}}\left(Me_{\boldsymbol{t}}\right) = Me_{\boldsymbol{t}+\boldsymbol{j}} \in \oCNN{}{}{d-1}{c,1}.
\end{equation}
for $\boldsymbol{t}\in \prod_{k=1}^D \left([d_k - (d-2), d-1]\bigcap [1, d_k]\right)$ and $|\boldsymbol{j}|\leq 1$.
Therefore, $Me_{\boldsymbol{t}}\in \oCNN{}{}{d-1}{c,1}$ for $\boldsymbol{t}\in \prod_{k=1}^D [1,d_k]$. This completes the proof for $d\geq 5$.
\end{proof}

\subsection{Proof of Lemma \ref{lemma:multidimensional_d23}}\label{appendix:proof_multidimensional_d23}
\begin{proof}
    For the input vector $x = (x^1, x^2, \dots , x^c)\in \mathbb{R}^{c\times \boldsymbol{d}}$, define the translation of $x^i\in\mathbb{R}^{\boldsymbol{d}}$ as follows:
\begin{equation}
    p^i_{\boldsymbol{j}} := U^{\boldsymbol{d}}_{-\boldsymbol{j}}(x^i).
\end{equation}

   \paragraph{Case 1 $d=2$:} 
Let a monomial $M$ be defined as:
\begin{equation}
    M:=   \prod_{i=1}^c\prod_{\boldsymbol{j}\in \prod_{k=1}^D[1,d_k]} (x^{i}_{\boldsymbol{j}})^{\alpha_{i,\boldsymbol{j}}}.
\end{equation}
If $M$ is not constant, there exists at least one $x^i_{\boldsymbol{j}}$, for which the degree $\alpha_{i,\boldsymbol{j}}$ is not zero.
Denote such $i$ and $\boldsymbol{j}$ as $i'$ and $\boldsymbol{j}'$.
Then, define $\boldsymbol{t}$ as $\boldsymbol{t}:=\boldsymbol{d} - \boldsymbol{j}'+1$. 
Because $1 \leq \boldsymbol{t}= \boldsymbol{d} - \boldsymbol{j}'+1\leq \boldsymbol{d}\leq 2$, the inequality $-1\leq \boldsymbol{j}- \boldsymbol{t} \leq 1$ holds for $\boldsymbol{j}\in \prod_{k=1}^D [1,d_k]$. 
Thus,
\begin{equation}
    \bigodot_{i=1}^c \bigodot_{\boldsymbol{j}\in \prod_{k=1}^D d_k} (p^{i}_{\boldsymbol{j}-\boldsymbol{t}})^{\alpha_{i,\boldsymbol{j}}}\in \osCNN{\sigma}{}{1}{c,1}.
\end{equation}
The $\boldsymbol{t}$-th component is calculated as 
\begin{multline}
     \left(\bigodot_{i=1}^c \bigodot_{\boldsymbol{j}\in \prod_{k=1}^D d_k} (p^{i}_{\boldsymbol{j}-\boldsymbol{t}})^{\alpha_{i,\boldsymbol{j}}}\right)_{\boldsymbol{t}}
     = \prod_{i=1}^c \prod_{\boldsymbol{j}\in \prod_{k=1}^D d_k} (p^{i}_{\boldsymbol{j}-\boldsymbol{t}})_{\boldsymbol{t}}^{\alpha_{i,\boldsymbol{j}}}
     = \prod_{i=1}^c \prod_{\boldsymbol{j}\in \prod_{k=1}^D d_k} (U^{\boldsymbol{d}}_{\boldsymbol{t} - \boldsymbol{j}}x^i)_{\boldsymbol{t}}^{\alpha_{i,\boldsymbol{j}}}
     \\ = \prod_{i=1}^c \prod_{\boldsymbol{j}\in \prod_{k=1}^D d_k} (x^i_{\boldsymbol{j}})^{\alpha_{i,\boldsymbol{j}}} = M.
\end{multline}
Furthermore, for $\boldsymbol{t}'\neq \boldsymbol{t}$, there exists an index $k$ such that $t_k \neq t'_k$.
This implies that $d_k=2$, and because $t_k = 3 - j'_k$, there are two cases, $t'_k = j'_k=2 $ and $t_k=1$ or $t'_k = j'_k=1 $ and $t_k=2$.

Then, $ \left(p^{i'}_{\boldsymbol{j'}-\boldsymbol{t}}\right)_{\boldsymbol{t}'}$ becomes zero because $t'_k + j'_k - t_k$ is $0$ or $3$.
Thus, we have:
\begin{equation}
\left(\bigodot_{i=1}^c \bigodot_{\boldsymbol{j}\in \prod_{k=1}^D d_k} (p^{i}_{\boldsymbol{j}-\boldsymbol{t}})^{\alpha_{i,\boldsymbol{j}}}\right)_{\boldsymbol{t}'}
= 
\prod_{i=1}^c \prod_{\boldsymbol{j}\in \prod_{k=1}^D d_k} (p^{i}_{\boldsymbol{j}-\boldsymbol{t}})_{\boldsymbol{t}'}^{\alpha_{i,\boldsymbol{j}}} = 0.
\end{equation}
Therefore, we can infer:
\begin{equation}
    \bigodot_{i=1}^c \bigodot_{\boldsymbol{j}\in \prod_{k=1}^D d_k} (p^{i}_{\boldsymbol{j}-\boldsymbol{t}})^{\alpha_{i,\boldsymbol{j}}} = Me_{\boldsymbol{t}}\in \osCNN{\sigma}{}{1}{c,1}.
\end{equation}
Finally, using the convolutional layer $U^{\boldsymbol{d}}_{\boldsymbol{j}- \boldsymbol{t}}\in \mathcal{L}^{3\boldsymbol{1}_D}_{1,1}$ as the last layer, $U^{\boldsymbol{d}}_{\boldsymbol{j}- \boldsymbol{t}}(M e^{\boldsymbol{d}}_{\boldsymbol{t}}) = M e^{\boldsymbol{d}}_{\boldsymbol{j}}\in \oCNN{}{}{2}{c,1}$ for $\boldsymbol{j}\in \prod_{k=1}^{D} [1,d_k]$. 
We have proved that all non-constant monomials are in $\oCNN{}{}{2}{c,1}$.
Additionally, by Lemma \ref{lemma:multidimensional_onehot}, all constant functions are also in $\oCNN{}{}{2}{c,1}$.
 This completes the proof for the case of $d=2$.
 
\paragraph{Case 2 $d=3$:} 
By Corollary \ref{lemma:multidimensional_onehot}, $e^{\boldsymbol{d}}_{\boldsymbol{t}}\in \oCNN{}{}{2}{c,1}$ for $\boldsymbol{t}\in \prod_{k=1}^D[1,d_k]$.
Furthermore, $p^i_{\boldsymbol{j}}\in \oCNN{}{}{2}{c,1}$ for $|\boldsymbol{j}|\leq 2$.
Define $\boldsymbol{t}'$ as $\boldsymbol{t}':=\lfloor(\boldsymbol{d}+1)/2\rfloor$, where $\lfloor \cdot \rfloor$ represents the floor function.  
Then, because $1\leq \boldsymbol{d}\leq 3$, it follows that $1\leq \boldsymbol{t}':=\lfloor(\boldsymbol{d}+1)/2\rfloor \leq \boldsymbol{d}$, thus $e^{\boldsymbol{d}}_{\boldsymbol{t}'}\in \oCNN{}{}{2}{c,1}$.

For an arbitrary monomial $M=\prod_{i=1}^{c}\prod_{\boldsymbol{j}\in \prod_{k=1}^D [1,d_k]}(x^i_{\boldsymbol{j}})^{\alpha_{i,\boldsymbol{j}}}$, by Lemma \ref{lemma:product}, we have:
\begin{equation}
    e^{\boldsymbol{d}}_{\boldsymbol{t}'} \odot \left(\bigodot_{i=1}^c\bigodot_{\boldsymbol{j}\in \prod_{k=1}^D[1,d_k]} (p^{i}_{\boldsymbol{j}-\boldsymbol{t}'})^{\alpha_{i,\boldsymbol{j}}}\right) 
   \in \osCNN{\sigma}{}{2}{c,1},
\end{equation}
and
\begin{multline}
       \left(\bigodot_{i=1}^c\bigodot_{\boldsymbol{j}\in \prod_{k=1}^D[1,d_k]} (p^{i}_{\boldsymbol{j}-\boldsymbol{t}'})^{\alpha_{i,\boldsymbol{j}}}\right)_{\boldsymbol{t}'}  
   =    \prod_{i=1}^c\prod_{\boldsymbol{j}\in \prod_{k=1}^D[1,d_k]} (U^{\boldsymbol{d}}_{\boldsymbol{t}'-\boldsymbol{j}}x^i)_{\boldsymbol{t}'} ^{\alpha_{i,\boldsymbol{j}}}
   \\=  \prod_{i=1}^c\prod_{\boldsymbol{j}\in \prod_{k=1}^D[1,d_k]} (x^i_{\boldsymbol{j}})^{\alpha_{i,\boldsymbol{j}}}
   =M.
\end{multline}
Therefore, we have $M e^{\boldsymbol{d}}_{\boldsymbol{t}'}\in \osCNN{\sigma}{}{2}{c,1}$.
Finally, using the convolutional layer $U^{\boldsymbol{d}}_{\boldsymbol{j}- \boldsymbol{t}'}\in \mathcal{L}^{3\boldsymbol{1}_D}_{1,1}$ as the last layer, $U^{\boldsymbol{d}}_{\boldsymbol{j}- \boldsymbol{t}'}(M e^{\boldsymbol{d}}_{\boldsymbol{t}'}) = M e^{\boldsymbol{d}}_{\boldsymbol{j}}\in \oCNN{}{}{3}{c,1}$ for $\boldsymbol{j}\in \prod_{k=1}^{D} [1,d_k]$. 
And this completes the proof for the case of $d=3$.
\end{proof}

\subsection{Proof of Lemma \ref{lemma:multidimensional_subspace}}\label{appendix:proof_multidimensional_subspace}
\begin{proof}
We aim to prove that for arbitrary $1\leq \boldsymbol{n},\boldsymbol{m} \leq \boldsymbol{d}$, $E^{\boldsymbol{d}}_{\boldsymbol{n},\boldsymbol{m}}\in S^{\boldsymbol{d}}_{d}$.
     For $\boldsymbol{n} = (n_1, \dots, n_D)$ and $\boldsymbol{m} = (m_1, \dots, m_D)$, $E^{\boldsymbol{d}}_{\boldsymbol{n}, \boldsymbol{m}}$ can be represented as:
     \begin{equation}
         E^{\boldsymbol{d}}_{\boldsymbol{n}, \boldsymbol{m}} 
         = \bigotimes_{k=1}^D E^{d_k}_{n_k,m_k}.
     \end{equation}
     By Lemma \ref{lemma:linear}, each $E^{d_k}_{n_k,m_k}$ is in $S^{d_k}_{d_k}\subset S^{d_k}_{d}$.
     It implies that there exist $N_k\in \mathbb{N}$ and $T_{i,j}\in \mathcal{T}^{d_k}_1$ such that $\sum_{i=1}^{N_k}\prod_{j=1}^{d} T_{i,j} = E^{d_k}_{n_k,m_k}$, where $T_{i,j}$ and $E^{d_k}_{n_k,m_k}$ are interpreted as linear transformations from $\mathbb{R}^{d_k}$ to $\mathbb{R}^{d_k}$.
     Each $T_{i,j}$ can be expressed as a linear combination of $U^{d_k}_{-1}, U^{d_k}_{0}$, and $U^{d_k}_{1}$:
     \begin{equation}
         T_{i,j} = \beta_{i,j,-1} U^{d_k}_{-1} + \beta_{i,j,0} U^{d_k}_{0} + \beta_{i,j,1} U^{d_k}_{1},
     \end{equation}
     for $\beta_{i,j,-1}, \beta_{i,j,0}, \beta_{i,j,1}\in \mathbb{R}$.
     This leads to:
     \begin{multline}
        E^{d_k}_{n_k,m_k} = \sum_{i=1}^{N_k}\prod_{j=1}^{d} T_{i,j} 
        =\sum_{i=1}^{N_k}\prod_{j=1}^{d} \left(\beta_{i,j,-1} U^{d_k}_{-1} + \beta_{i,j,0} U^{d_k}_{0} + \beta_{i,j,1} U^{d_k}_{1}\right)
         \\ = \sum_{i=1}^{N_k} \sum_{(l_1,\dots,l_d)\in \{-1,0,1\}^d} \prod_{j=1}^d  \beta_{i,j,l_j} U^{d_k}_{l_j},
     \end{multline}
     where $\prod$ is the composition of functions.
We calculate $E^{\boldsymbol{d}}_{\boldsymbol{n},\boldsymbol{m}}$ as: 
\begin{align}
    &E^{\boldsymbol{d}}_{\boldsymbol{n},\boldsymbol{m}}  = \bigotimes_{k=1}^D E^{d_k}_{n_k,m_k}
    = \bigotimes_{k=1}^D\left( \sum_{i=1}^{N_k} \sum_{(l_1,\dots,l_d)\in \{-1,0,1\}^d} \prod_{j=1}^d  \beta_{i,j,l_j} U^{d_k}_{l_j}\right)
    \\ &=  \sum_{i_1, \dots, i_D=1}^{N_1, \dots, N_D}\sum_{\substack{(l_{1,1},\dots, l_{1,d})\\ \dots\\(l_{D,1},\dots, l_{D,d})}\in \{-1,0,1\}^d}  
    \bigotimes_{k=1}^D \left(\prod_{j=1}^d \beta_{i,j,l_{k,j}}U^{d_k}_{l_{k,j}}\right) 
    \\&=  \sum_{i_1, \dots, i_D=1}^{N_1, \dots, N_D}\sum_{\substack{(l_{1,1},\dots, l_{1,d})\\ \dots\\(l_{D,1},\dots, l_{D,d})}\in \{-1,0,1\}^d}  
 \left(\prod_{j=1}^d\prod_{k=1}^D \beta_{i,j,l_{k,j}}\right) \prod_{j=1}^d \bigotimes_{k=1}^D U^{d_k}_{l_{k,j}}.
\end{align}
     Because $\bigotimes_{k=1}^D U^{d_k}_{l_{k,j}} = U^{\boldsymbol{d}}_{\sum_{k=1}^D l_{k,j} e_{k} }\in \mathcal{L}^{\boldsymbol{d}, 3\boldsymbol{1}_D}_{1,1}$, it follows that $E^{\boldsymbol{d}}_{\boldsymbol{n},\boldsymbol{m}}\in S^{\boldsymbol{d}}_d$.
     This completes the proof.
\end{proof}



\bibliographystyle{elsarticle-harv} 
 \bibliography{reference}





\end{document}